\documentclass[journal]{IEEEtran}

\usepackage{multirow}
\usepackage{ifpdf}
\usepackage{epsfig}
\usepackage{subcaption}
\usepackage[table]{xcolor}
\usepackage[utf8]{inputenc}
\usepackage{cite}
\usepackage[numbers]{natbib}
\usepackage{graphicx}

\usepackage{color}
\definecolor{veg}{RGB}{0, 251, 20}
\definecolor{sap}{RGB}{226, 175, 165}
\definecolor{che}{RGB}{222, 150, 200}
\definecolor{sul}{RGB}{209, 109, 106}
\definecolor{sha}{RGB}{212, 212, 212}
\definecolor{psha}{RGB}{178, 167, 211}
\definecolor{maf}{RGB}{212, 231, 205}
\definecolor{mafb}{RGB}{50, 205, 50}
\definecolor{fela}{RGB}{175, 238, 238}
\definecolor{felb}{RGB}{244, 199, 131}
\definecolor{fel}{RGB}{252, 225, 198}

\usepackage{soul}

\ifCLASSINFOpdf
\else
\fi
\usepackage[cmex10]{amsmath}
\usepackage{algorithmic}
\usepackage{array}
\usepackage{stfloats}
\usepackage{url}
\usepackage{hyperref}
\hypersetup{
    colorlinks=true,
    linkcolor=blue,
    urlcolor=green,
    citecolor=blue,
    }
    
\begin{document}
\bstctlcite{IEEEexample:BSTcontrol}

%\title{Tinto: A Multi-sensor Benchmark Dataset for 3D Hyperspectral Point Cloud Segmentation in the Geosciences}
\title{Tinto: Multisensor Benchmark for 3D Hyperspectral Point Cloud Segmentation in the Geosciences}
%\title{Tinto: Multi-sensor Dataset for Geoscience Point Clouds}
%
% Tinto: A Multi-sensor Benchmark Dataset for Hyperspectral Point Cloud Segmentation in the Geosciences

%\author{Michael~Shell,~\IEEEmembership{Member,~IEEE,}
%        John~Doe,~\IEEEmembership{Fellow,~OSA,}
%        and~Jane~Doe,~\IEEEmembership{Life~Fellow,~IEEE}% <-this % stops a space
%\thanks{M. Shell is with the Department
%of Electrical and Computer Engineering, Georgia Institute of Technology, Atlanta,
%\thanks{Manuscript received April 19, 2005; revised September 17, 2014.}}

\author{Ahmed~J.~Afifi, Samuel~T.~Thiele, Aldino~Rizaldy, Sandra~Lorenz, Pedram~Ghamisi, Raimon~Tolosana-Delgado, Moritz~Kirsch, Richard~Gloaguen, Michael~Heizmann % <-this % stops a space
\thanks{Ahmed J. Afifi, Samuel T. Thiele, Aldino Rizaldy, Sandra Lorenz, Pedram Ghamisi, Raimon Tolosana-Delgado, Moritz Kirsch, and Richard Gloaguen are with the Helmholtz-Zentrum Dresden-Rossendorf (HZDR), Helmholtz Institute Freiberg for Resource Technology (HIF), 09599 Freiberg, Germany.}%
\thanks{Ahmed J. Afifi $\&$ Michael Heizmann are with the Institute of Industrial Information Technology (IIIT), Karlsruhe Institute of Technology (KIT), 76187 Karlsruhe, Germany.}%
\thanks{Pedram Ghamisi is also with the Institute of Advanced Research in Artificial Intelligence (IARAI), 1030 Vienna, Austria.}%
%\thanks{Raimon Tolosana-Delgado is with the Modelling and Valuation Department, Helmholtz Institute Freiberg for Resource Technology (HIF), 09599 Freiberg, Germany.}%
%\thanks{M. Shell is with the Department
%of Electrical and Computer Engineering, Georgia Institute of Technology, Atlanta,
%\thanks{Manuscript received April 19, 2005; revised September 17, 2014.}
}

\maketitle

\begin{abstract}
The increasing use of deep learning techniques has reduced interpretation time and, ideally, reduced interpreter bias by automatically deriving geological maps from digital outcrop models. However, accurate validation of these automated mapping approaches is a significant challenge due to the subjective nature of geological mapping and the difficulty in collecting quantitative validation data. Additionally, many state-of-the-art deep learning methods are limited to 2D image data, which is insufficient for 3D digital outcrops, such as hyperclouds. To address these challenges, we present Tinto, a multi-sensor benchmark digital outcrop dataset designed to facilitate the development and validation of deep learning approaches for geological mapping, especially for non-structured 3D data like point clouds. Tinto comprises two complementary sets: 1) a real digital outcrop model from Corta Atalaya (Spain), with spectral attributes and ground-truth data, and 2) a synthetic twin that uses latent features in the original datasets to reconstruct realistic spectral data (including sensor noise and processing artifacts) from the ground-truth. The point cloud is dense and contains 3,242,964 labeled points. We used these datasets to explore the abilities of different deep learning approaches for automated geological mapping. By making Tinto publicly available, we hope to foster the development and adaptation of new deep learning tools for 3D applications in Earth sciences. The dataset can be accessed through this link: \url{https://doi.org/10.14278/rodare.2256}.
\end{abstract}


% \fi
\IEEEpeerreviewmaketitle

\section{Introduction} \label{introduction}
\IEEEPARstart{T}{he} need for annotated datasets to train and assess deep learning models has become essential in numerous advanced fields of research, including remote sensing and Earth observation \cite{9884725}. Although there are several (ongoing) attempts to promote benchmarking and open science in the remote sensing field by developing exhaustive lists of available datasets \cite{9884725}\footnote{\url{https://github.com/satellite-image-deep-learning/datasets}}, evaluation servers (e.g., DASE\footnote{\url{http://dase.grss-ieee.org/}}), cloud services (e.g., Amazon Web Services\footnote{\url{https://registry.opendata.aws/}}, Microsoft's Planetary Computer\footnote{\url{https://planetarycomputer.microsoft.com/catalog}}, Radiant Earth's MLHub\footnote{\url{https://www.mlhub.earth/}}), and benchmark datasets (e.g., \cite{9921261,8707433,chang2015shapenet,sumbul2019bigearthnet}), novel applications within the fields of geomorphology, and geology remain sparse and unclear. Also, generally applicable reference data in remote sensing and geosciences for evaluating machine learning approaches are still not available in sufficient quantity and quality \cite{9527092,lorenz2022three}.% perhaps restructure sentence.

Hyperspectral remote sensing has emerged as a powerful tool for detecting subtle spectral differences in mineralogical composition. The ability to apply this technology from a range of platforms, including satellites, airplanes, autonomous vehicles, and tripods, supports geological mapping at various scales. Integrating surface spectral data with topographic data enables the creation of hyperclouds, which are geometrically and radiometrically accurate point cloud representations of the target. Depending on the analysed range of the electromagnetic spectrum, a variety of minerals can be detected and mapped in their original 3D context. The visible and near-infrared (VNIR) and shortwave infrared range (SWIR) are useful for detecting spectral features of alteration minerals, such as oxides and hydroxides. In contrast, the long-wave infrared region (LWIR) allows for the detection of many rock-forming minerals, such as quartz and feldspars. Integrating both information sources enables comprehensive lithological mapping and has numerous geological applications, including greenfield exploration of critical raw materials, geological mapping of open-pit and underground mines, compositional mapping of rock wastes and stockpiles, and soil contamination mapping in post-mining landscapes.

Despite the vast potential of deep learning for hyperspectral mapping \cite{8697135}, it has found limited applicability for geoscientific applications, primarily due to the challenges associated with validation and the benchmarking of adapted algorithms \cite{9527092}. Ground truth is difficult to establish for large-scale datasets, due to the limited accessibility of geological outcrops, their inherently complex and heterogeneous mineralogical composition as well as the traditionally rather subjective definition of lithological domains by geological experts based on mostly visual criteria. While simulated datasets have been used in the past, they do not yet capture the complexity and variability of real-world geological environments. The rather novel concept of 3D digital spectral outcrops (e.g., hyperclouds) is causing further complications due to the unstructured and complex nature of spectral point cloud data, which is incompatible with most state-of-the-art algorithms. Current approaches partially solve these issues by simplifying the dataset either spatially (working in the 2D image space \cite{ibrahim2021exploiting}) or spectrally (segmentation based on selected features \cite{chen2019hyperspectral}) before applying segmentation in 2D or 3D, respectively. For both ways, the potential of the dataset is not fully exploited and important information might be lost in the process. Geological scenes are preferred over highly structured scenes for benchmarking deep learning models in geosciences due to their complexity, variability, and real-world relevance. Geological scenes exhibit various features like rock formations, mineralogy, and topography, creating a realistic and challenging testing environment for deep learning algorithms. Class boundaries are diffused in the geological scenes and the classes are highly mixed. Benchmarking geological scenes enhance the models' generalization capability and learning new patterns and features from the training data, allowing them to adapt to new unseen geological scenarios and making them more robust and reliable in real-world applications. Additionally, it enables researchers to understand deep learning models' performance in capturing geological information and aligns with the needs of geoscientific applications. Overall, benchmarking on geological scenes provides a robust evaluation environment and facilitates tailored deep learning solutions in geosciences. Lately, few 3D hyperspectral datasets with geoscientific application context have been openly published (e.g., \cite{laukamp2021rocklea, lorenz2022three}). Nevertheless, none of them offers an adequate amount of ground truth data to qualify as a benchmark dataset.

In this contribution, we present a large and geologically complex but well-understood real-world benchmark dataset, and a synthetic (reconstructed) equivalent, designed for testing and comparing deep learning methods for hyperspectral geological mapping. The real-world dataset covers Corta Atalaya, an abandoned open pit mine within the Minas de Rio Tinto copper mining district in Andalusia, Spain. The hyperspectral data have been acquired using plane, drone, and tripod-based acquisition and cover the VNIR, SWIR, and LWIR range of the electromagnetic spectrum. Lithology class labels have been defined for the whole dataset based on a combination of detailed laboratory analysis and derived supervised classification \cite{thiele2021multi} and were adjusted based on an expert interpretation of the geology.
However, due to the complex nature of geological datasets, this labeling cannot be treated with $100\%$ confidence. To address this shortcoming, we used the real-world benchmark dataset to derive a realistic synthetic dataset in which class labels (and associated spectral endmembers and abundances) are known with certainty. This approach allows us to develop data for which the class and abundance properties are known with confidence while retaining the spatial statistical properties and complexity of a real dataset.

To facilitate established and emerging deep learning approaches, we present these datasets both in 2D raster form (as is conventional for remote sensing applications) and 3D point cloud form (for emerging approaches that are beginning to move beyond the topological limitations imposed by 2D rasters). Challenges and limitations associated with each data representation, and a selection of tools available for working with them, are discussed.

The rest of the paper is structured as follows. Section \ref{background} reviews some related work and the available datasets for 3D point cloud processing. Section \ref{dataset} describes the proposed Tinto benchmark dataset and how it is collected, labeled, and synthesised in detail. In Section \ref{evaluation}, we discuss the baseline deep learning models that are used to evaluate the Tinto dataset, the experimental setup, and the experimental outcomes. Finally, the conclusion and remarks are drawn in Section \ref{conclusion}.

%\textcolor{red}{\textbf{Richard}}, \textcolor{red}{\textbf{Pedram}}, \textcolor{red}{\textbf{Sandra}}, \textcolor{red}{\textbf{Sam}}

\section{Background and Related Work} \label{background}
%\textcolor{red}{\textbf{Ahmed, Pedram}}

%Many 3D point cloud segmentation methods were proposed, and they used the publicly available datasets to evaluate their performance. The proposed methods vary between the traditional, machine learning, and deep learning methods. Following, we review different point cloud segmentation methods. Also, we present various 3D datasets that are used for the point cloud segmentation task.
A plethora of sophisticated methodologies has been advanced to segment 3D point clouds, often utilizing publicly available datasets to evaluate their efficacy. These methodologies encompass a broad spectrum of techniques, from traditional methods to cutting-edge machine learning and deep learning approaches \cite{grilli2017review}. In the following subsections, we provide a very brief overview of the mature field of some diverse point cloud segmentation approaches by highlighting a few relevant examples and presenting a comprehensive collection of 3D datasets commonly employed for this specific task.
\subsection{Point Cloud Segmentation}
Traditional point cloud segmentation methods rely on strict hand-crafted geometric constraints and rules. The main goal of the segmentation process is to group 3D points into non-overlapping regions. The generated regions have common semantic meanings and geometric structures \cite{xie2020linking}. With the introduction of machine learning and deep learning models in solving 2D tasks and the availability of large-scale labeled datasets, many researchers proposed machine/deep learning models to segment point clouds from the object and scene levels. Generally, deep learning models achieved remarkable performance compared to traditional and machine learning methods. Following, we discuss different point cloud segmentation methods.
\subsubsection{Edge-based}
Edge-based methods try to detect points close to the edge by calculating the rapid changes in the intensity (the feature associated with the points), normals, or the gradient. This will create boundaries between two different regions. Then, the points are grouped inside the same region where changes are small. These methods perform segmentation quickly but struggle to achieve accurate results when dealing with point clouds from large areas due to issues like noise and uneven point distribution \cite{rabbani2006segmentation, castillo2013point}.

\subsubsection{Region growing} 
Region growing-based methods involve the random selection of seed points and the measure of geometrical or feature similarity between the seeds and neighboring points. Points with similar features are merged to create one region. This process is performed iteratively until all points are merged into similar regions. These methods were firstly applied to 2.5D LiDAR data and they were widely applied for the segmentation of building structures. Similar points that belong to the same region can be selected by comparing their features or calculating the Euclidean similarity. The segmented points are selected for example by fitting a plane to a number of points in a given volume and then points with the minimum distance to that plane are merged \cite{besl1988segmentation, vo2015octree, nurunnabi2012robust}.

\subsubsection{Shallow supervised Machine Learning}
Shallow supervised machine learning refers to non-deep algorithms that use labeled data to train a model. These methods allow classifying points in a cloud based on predefined features such as maximum likelihood based on support vector machine (SVM) \cite{zhang2013svm}, random forests (RF) \cite{chehata2009airborne}, and Bayesian discriminant classifiers \cite{srivastava2007bayesian}. Other groups of methods depend on statistical contextual models such as Conditional Random Fields (CRF) \cite{lim20093d} and Markov Random Fields (MRF) \cite{lu2012simplified}. These methods focus on the statistics and the relational information of the points over different scales. Machine learning models applied for point cloud segmentation perform a neighborhood point selection, then feature extraction from the grouped points, feature selection to reduce the feature dimensionality and then segment the points semantically.

\subsubsection{Deep Learning}
Deep learning has become the most influential and hottest technique in different research fields such as computer vision, medical imaging, autonomous driving, and robotics. Deep learning is a special branch of machine learning where the models are deeper, more complex and the extracted features generally have higher dimensions than the ones extracted from traditional machine learning methods. Applicable methods for applying deep learning on 3D data depend on how the data is represented. With multi-view data, a normal 2D Convolutional Neural Network (CNN) can be easily applied, such as the MVCNN model \cite{su2015multi}. Voxel-based data can be used with 3D CNN, where the normal 2D CNN can be easily extended to 3D \cite{maturana2015voxnet}. The drawback of using voxel-based representation is the memory and the computation cost to train a model. To overcome the voxel-based and multi-view methods, models that can be applied directly on point cloud data were recently proposed as a promising solution. Models applied directly on point clouds such as the pioneer model PointNet \cite{qi2017pointnet} were followed by the improved version PointNet$++$ \cite{qi2017pointnet2} and the Dynamic Graph CNN (DGCNN) \cite{wang2019dynamic}. Other approaches of point cloud segmentation in the field of remote sensing and 3D laser scanning can be found in  \cite{lin2022weakly, huang2021granet, lin2021local}.

The research on point cloud segmentation using deep learning is a hot research topic. Different models are proposed with either sophisticated layers \cite{wu2019pointconv, li2018pointcnn, cai2019spatial} to deal with the point cloud as an unordered set or with a simple architecture using MLP as a backbone of the model \cite{ma2022rethinking} to achieve improved performance with less computation and memory cost.

\subsection{Available 3D Datasets}
%\subsubsection{3D Datasets} \hfill \break
To our knowledge, no 3D hyperspectral benchmark datasets have been published for a geoscientific application context yet. Available benchmark datasets for point cloud segmentation deal with indoor scenes, such as Stanford Large-Scale 3D Indoor Spaces (S3DIS) \cite{xu2020grid} and Semantic3D.Net \cite{hackel2017semantic3d}, or urban scenes, such as Sydney Urban Objects Dataset \cite{quadros2012occlusion}, Toronto-3D \cite{tan2020toronto3d}, and SemanticKITTI \cite{behley2019iccv}, for semantic segmentation. Other datasets focus on the objects, for instance segmentation or object parts segmentation, such as ShapeNet \cite{chang2015shapenet} and ModelNet40-C \cite{sun2022benchmarking}. In most cases, point cloud attributes are limited to the 3D coordinates (X, Y, Z), intensity or RGB color values. The Maarmorilik Dataset \cite{lorenz2022three} is an open-source 3D hyperspectral dataset capturing the complex geology of the Black Angel Mountain in Maarmorilik, West Greenland, alongside a detailed and interactive tutorial documenting relevant processing workflows for hypercloud data. It includes RGB and VNIR-SWIR hyperspectral data but does not provide ground truth and thus cannot be defined as a benchmark.

In \cite{lopez, brell20193d}, similar work has been conducted by generating hyperspectral point clouds, but the datasets are not available to the public. The MUUFL Gulfport dataset \cite{muufl} and GRSS18 dataset \cite{8727489}, on the other hand, provide the ground truth and are publicly open. They consist of LiDAR and hyperspectral data. However, the ground truth is in a 2D format rather than in a 3D point format as offered by our dataset. Some work \cite{isprs-2019, weinmann2017geospatial, isprs-2018, Chen19} have also been done investigating the utilization of hyperspectral data with LiDAR for classification task but neglecting the use of deep learning. Moreover, these works \cite{isprs-2019, weinmann2017geospatial, isprs-2018} performed the classification in a 2D style. The work in \cite{weidner2021classifying} investigated the use of smartphone and UAV photogrammetry to assess rock slope hazards in mountainous regions. Different datasets were created according to lighting conditions, slope morphologies, and seasons. They were manually labeled and were classified using Random Forest into geologically relevant categories. The research compares 12 different point cloud feature sets, finding that feature sets focused on geometry, slope, and texture perform significantly better than those incorporating absolute color features, which are sensitive to lighting changes and struggle to distinguish between geological materials. More recent works \cite{rs14092113, decker2023hyperspectral, mitschke2022hyperspectral} use deep learning models for hyperspectral point cloud segmentation and proved that hyperspectral data improved the performance of the models. These results motivated us to employ various deep learning models in our work.

To the best of the authors' knowledge, the Tinto dataset will be the first-ever dataset that provides the following features: \textbf{(1)} a 3D point cloud of a real outcrop, \textbf{(2)} the corresponding ground truth, \textbf{(3)} the same scene captured using different sensors (RGB, VNIR, SWIR, and LWIR), \textbf{(4)} hyperspectral information attached to each point in the point cloud, \textbf{(5)} two types of corresponding synthetic data (clean and noisy data), and \textbf{(6)} 2D views of the scene from three different directions.

\section{Tinto Dataset}
\label{dataset}

\subsection{Data Acquisition and Correction}

Several steps of the acquisition and processing of the Corta Atalaya hyperclouds have been previously described by \cite{thiele2021multi, kirsch2021characterisation}. To summarise briefly, a tripod-mounted hyperspectral Specim AisaFenix camera was used to capture oblique VNIR and SWIR imagery from three locations on the edge of the Corta Atalaya open-pit. Each of these rasters was then back-projected onto a dense 3D point cloud derived from 488 RGB photographs (captured using a Nikon D850 DSLR camera and Nikkor 85 mm f/1.8G lens) using the structure from motion multi-view stereo method implemented in Agisoft Metashape Professional v1.6. Atmospheric effects, which result largely from (i) the spectral signature of sunlight, (ii) interactions between this light and the atmosphere, and (iii) uneven illumination across the complex surface of the Corta Atalaya mine, were corrected using the method described by \cite{thiele2021novel} during the back-projection step.

The long-wave infrared hyperspectral data was collected in August 2020 during a larger hyperspectral airborne mapping campaign \cite{kirsch2021characterisation}. A Hyper-Cam FTIR hyperspectral camera from Telops was deployed, covering the electromagnetic spectrum between 7.7 and 11.8 micrometers. The collected raw data were processed using a standard workflow: The individual data cubes were orthorectified using a 2.5 m resolution Lidar-based terrain model, acquired within the same campaign, and subsequently stitched to a mosaic (average ground sampling distance of 1.2 m) using Telops' Reveal Airborne geolocation tool (version 2). Initial radiometric correction to at-sensor radiance was performed using the Telops Reveal Calibrate Software Version 5.2.8. Atmospheric correction was done by an In-Scene Atmospheric Compensation algorithm, while the separation of temperature and emissivity was performed based on emissivity normalization from the radiance data \cite{kirsch2021characterisation}. The resulting calibrated emissivity mosaic was then sampled onto the same dense 3D point cloud that was used with the VNIR-SWIR data to create an LWIR hypercloud to be included in this benchmark dataset. Figure \ref{ds-tree} visualizes an overview of the Tinto benchmark dataset with the various datasets it contains \cite{tinto-dataset}. The 3D visualization of the Tinto point clouds on Potree \cite{schutz2016potree} can be accessed through this link: \url{https://www.hzdr.de/FWG/FWGE/Hyperclouds/Tinto.html}

\begin{figure}[htpb]
    \centering
    \includegraphics[width=\columnwidth]{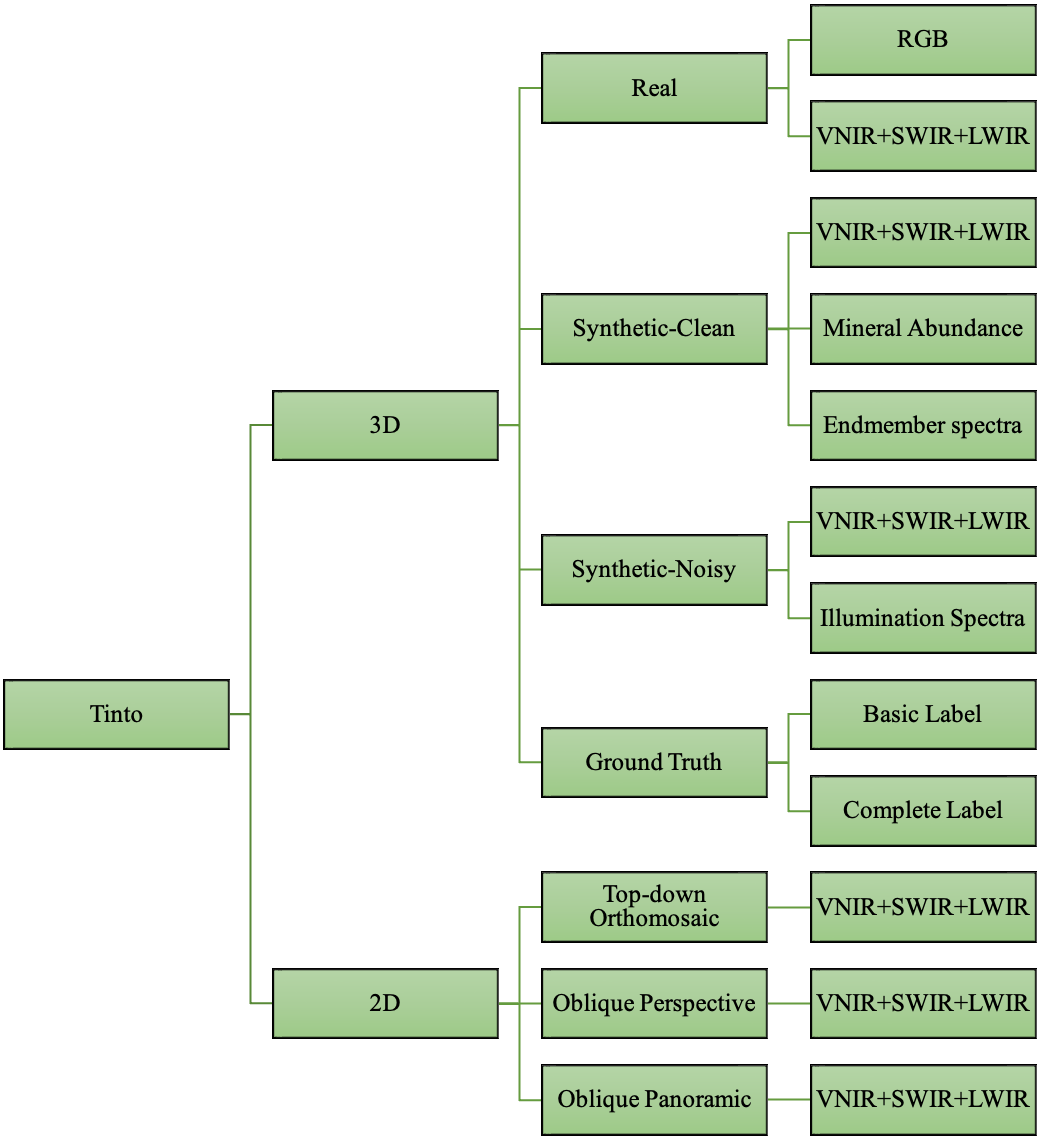}
    \caption{An overview of the Tinto benchmark structure and the various datasets it contains.}
    \label{ds-tree}
\end{figure}

\subsection{Data labelling and synthetic twin}

\begin{figure*}[!htpb]
\centering
\begin{subfigure}{0.98\textwidth}
    \centering
    \includegraphics[width=0.98\linewidth]{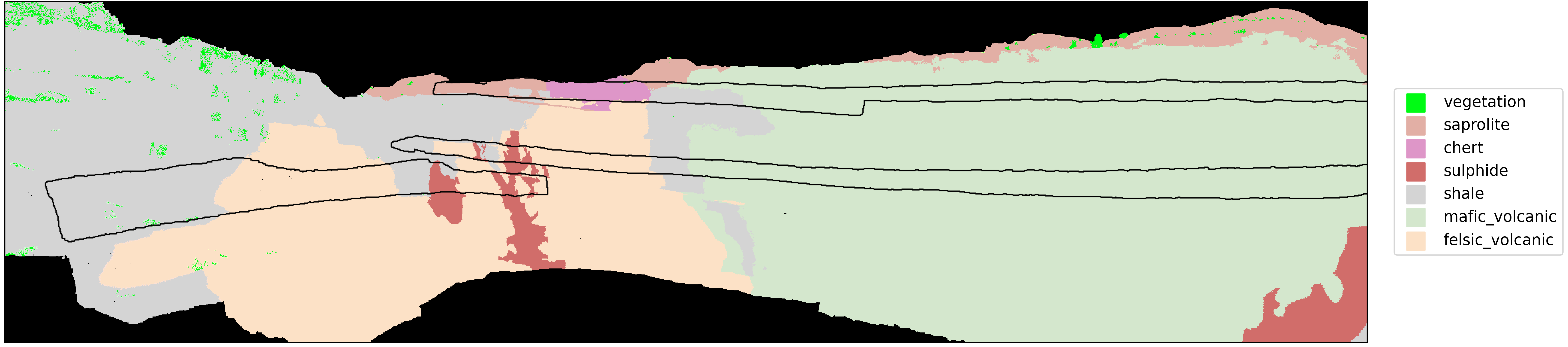} %
    \caption{Basic (6-lithology) segmentation.}
\end{subfigure} \\

\begin{subfigure}{0.98\textwidth}
    \centering
    \includegraphics[width=0.98\linewidth]{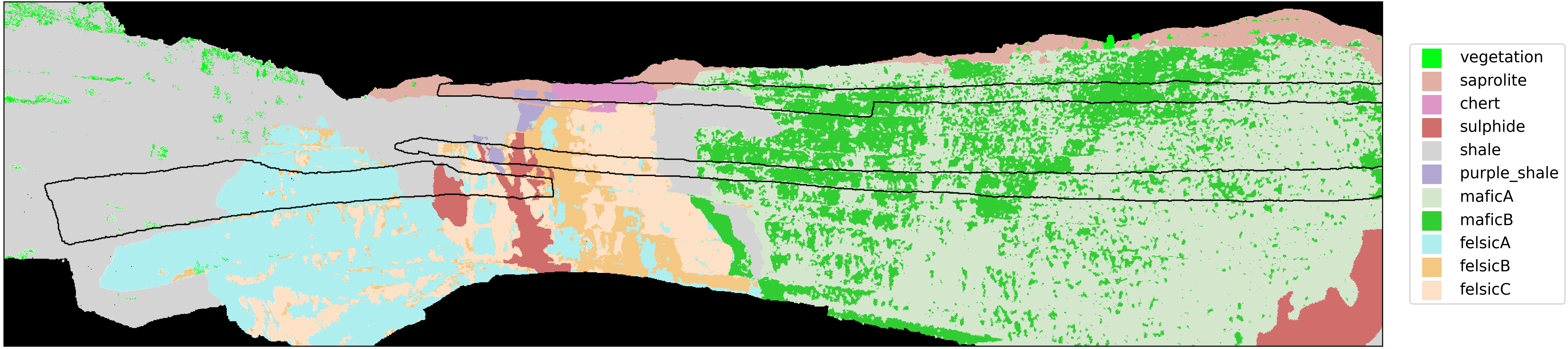} %
    \caption{Complete (10-lithology) segmentation.}
\end{subfigure}%
\caption{Simplified (top) and complete (bottom) ground truth labels provided for this benchmark dataset. The suggested training subset is outlined in black and follows traverses that match roughly with how a field geologist would collect data.}
\label{split}
\end{figure*}

Geological maps, i.e., classifications that show the spatial distribution of rock types, are generally subjective interpretations of the map author due to (i) cover by vegetation or soil, (ii) ambiguous rock type definitions, and (iii) the thematic or purpose the author has in mind when creating the map. For example, a map intended to constrain geotechnical aspects of a mine would often differ significantly from one made to quantify the composition and distribution of ore, due to subjective choices made when defining and classifying different rock units. While this ambiguity is an important justification for automated methods offered by e.g., machine learning approaches, which can improve objectivity while simultaneously allowing for data to be reprocessed for various purposes, it presents significant challenges when developing meaningful and reliable benchmarks. In this contribution, we have mitigated these challenges using two radically different approaches: (1) deriving a manually vetted but largely data-driven classification for ground-truthing purposes, and (2) back-calculating a realistic dataset (synthetic twin) from this classification result to derive a hyperspectral dataset for which the original rock composition is known for each pixel. These two approaches are described in the following sections.

\subsubsection{Geological ground-truth}

In the first approach, which aims at a meaningful ground-truth classification for the real hyperspectral data, we have integrated and synthesised hyperspectral information, sample mineralogy, field mapping and published geological understanding of Corta Atalaya. The spectral classification results of \cite{thiele2021multi} were used as a base for the ground truth and manually corrected where field data, ground sampling and expert interpretation of the high-resolution photogrammetric model showed clear mislabelling. These results (Figure \ref{split}) were subsequently checked by mine geologists on-site, resulting in a labelled data set that we consider to be as accurate as practically possible for geological applications. 

Several classes in this classification are spectrally and geologically related (e.g., classes defined by the presence of different but related alteration minerals; \cite{thiele2021multi}). Lumping these together, we derive a simplified classification containing 6 rock types. While we encourage people to use the full label set, this simplified version could be useful for evaluating approaches that perform poorly with a large number of classes (e.g., unsupervised methods).

We have also defined a suggested training subset (Figure \ref{split}) to ensure consistent results between studies. This has been selected such that (1) it covers all classes in the dataset, and (2) matches with what could be realistically achieved in practice, with training data distributed along three bench-traverses that are typical for geological mapping in open-pit environments. Note that this geometry results in a highly imbalanced training set, a common challenge for hyperspectral classification problems. Table \ref{basic-label} and Table \ref{complete-label} present the number of points per class in the basic and complete ground truth, respectively.

\begin{table}[!htpb]
    \caption{Number of points per class in the training/testing split for the basic-label ground truth. Vegetation class is excluded during training and testing.}
    \label{basic-label}
    \begin{center}
    \begin{tabular}{|l|c|c|c|}
      \hline
      \textbf{Class} & \textbf{Training} & \textbf{Testing} & \textbf{Color} \\
      \hline
      Vegetation$^\star$ & \multicolumn{2}{c|}{55,179} & \cellcolor{veg} \\
      \hline
      Saprolite & 28,231 & 309,540 & \cellcolor{sap} \\
      \hline
      Chert & 22,007 & 27,711 & \cellcolor{che} \\
      \hline
      Sulphide & 13,095 & 128,264 & \cellcolor{sul} \\
      \hline
      Shale & 49,518 & 987,790 & \cellcolor{sha} \\
      \hline
      Mafic\textunderscore volcanic & 143,861 & 1,037,782 & \cellcolor{maf} \\
      \hline
      Felsic\textunderscore volcanic & 41,256 & 398,730 & \cellcolor{fel} \\
      \hline
    \end{tabular}
    \end{center}
\end{table}

\begin{table}[!htpb]
    \caption{Number of points per class in the training/testing split for the complete-label ground truth. Vegetation class is excluded during training and testing.}
    \label{complete-label}
    \begin{center}
    \begin{tabular}{|l|c|c|c|}
      \hline
      \textbf{Class} & \textbf{Training} & \textbf{Testing} & \textbf{Color} \\
      \hline
      Vegetation$^\star$ & \multicolumn{2}{c|}{55,179} & \cellcolor{veg} \\
      \hline
      Saprolite & 28,229 & 309,263 & \cellcolor{sap} \\
      \hline
      Chert & 22,046 & 28,953 & \cellcolor{che} \\
      \hline
      Sulphide & 13,066 & 129,085 & \cellcolor{sul} \\
      \hline
      Shale & 44,321 & 946,634 & \cellcolor{sha} \\
      \hline
      Purple\textunderscore shale & 5,138 & 40,498 & \cellcolor{psha} \\
      \hline
      MaficA & 81,092 & 742,868 & \cellcolor{maf} \\
      \hline
      MaficB & 62,745 & 293,460 & \cellcolor{mafb} \\
      \hline
      FelsicA & 20,790 & 166,804 & \cellcolor{fela} \\
      \hline
      FelsicB & 6,783 & 85,818 & \cellcolor{felb} \\
      \hline
      FelsicC & 13,758 & 146,434 & \cellcolor{fel} \\
      \hline
    \end{tabular}
    \end{center}
\end{table}

\subsubsection{Synthetic twin}

Potential issues associated with remaining biases or inconsistencies in the ground-truth labels have been addressed by generating an entirely synthetic suite of spectral data by forward modelling. These share the same labels as the real dataset, as well as several latent variables and spatial relationships, but are derived using a spectral mixing model and a spatial distribution of mineral abundances simulated using spectral proxies and sample measurements for each class from \cite{thiele2021multi}. We suggest that these synthetic spectra are suited for comparing learning approaches, as the ground truth is known with certainty, while the real spectra can be used to evaluate performance on realistic data. The procedure followed to generate this synthetic twin is outlined below.

Firstly, three latent features known to correlate with specific mineral abundances (spectral proxies) were extracted from the real dataset using established minimum wavelength mapping and band-ratio techniques \cite{thiele2021multi}. These were normalised to have a mean of zero and standard deviation of one and assembled into a vector $\textbf{L}$ containing the latent feature at every point, ensuring that spatial associations present in the real dataset (and potentially informative for deep machine learning methods) are preserved in the synthetic one.

Next, mineral abundances from x-ray diffraction measurements on the ground-truth samples \cite{thiele2021multi} were used to define a mean composition for each class. To ensure the synthetic abundances sum to one, %(closure), 
the so called additive log ratio transformation (ALR) \cite{Ait1986} was used.
As reference phase, an abundant phase was chosen, generally quartz. Sulphide was used for the massive sulphide class. Hence, the ALR transformed abundance $\boldsymbol{\alpha}$ of the remaining phases was computed for each point $x$ by

\begin{equation}
\label{eq1}
%\boldsymbol{\alpha}_{i,j} = \log \left(\frac{\widehat{\textbf{A}}_{i,j}} {\widehat{\textbf{A}}_{0,j}}\right) + \sigma  \textbf{M}_i^{T} \cdot \mathbf{ \Lambda }_{j},
\alpha_{i,j}(x) = \log \left(\frac{\widehat{{A}}_{i,j}} {\widehat{{A}}_{0,j}}\right) + \sigma  \textbf{M}_i^{T} \cdot \mathbf{\Lambda}_j(x), \quad i=1, 2, \ldots, n
\end{equation}

%where $\widehat{\textbf{A}}_{i,j}$ denotes the average abundance of mineral $i$ in class $j$, $\widehat{\textbf{A}}_{0,j}$ denotes the abundance of the reference mineral for class $j$, $\mathbf{\Lambda}_j$ contains the values of the three latent variables described previously for class $j$, and $\textbf{M}$ contains a manually defined mapping matrix that determines the %correlation between mineral $i$'s abundance and each latent variable.
\noindent where $\widehat{{A}}_{i,j}$ denotes the average abundance of mineral $i\in \{0,1,\ldots, n\}$ in class $j$---$\widehat{{A}}_{0,j}$ being then the abundance of the reference mineral for class $j$---, the vector $\mathbf{\Lambda}_j(x)$ contains the values of the three latent variables described previously for class $j$ at location $x$, and $\textbf{M}_i$ contains a manually defined mapping vector that determines the %correlation between mineral $i$'s abundance and each latent variable.
contribution of each latent variable to the log-abundance of the $i$-th mineral. Finally,
$\sigma$ scales the log standard deviation of the mineral abundances within each class, and was kept at a constant value of 0.3 after some experimentation. 

\begin{figure*}[thpb]
    \centering
    \includegraphics[width=\textwidth, height=6cm]{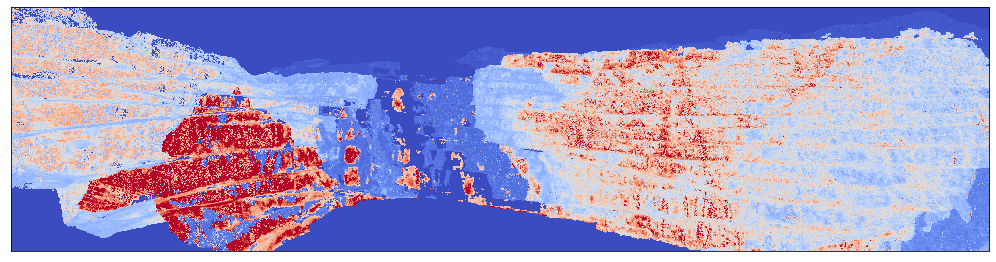}
    \caption{Example of a simulated mineral abundance map (chlorite in this case). These were used to derive synthetic reflectance spectra with realistic spatial variations.}
    \label{simulated_abun}
\end{figure*}

A vector of closed abundances $\textbf{A}(x)$ was then calculated for each point $x$ by inverting the additive log transform \cite{Ait1986}

\begin{equation}
\label{eq2}
    \textbf{A}_j(x) = \frac{ \exp\begin{bmatrix}
           0 &
           \alpha_{1,j}(x) &
           \hdots &
           \alpha_{n,j}(x)
         \end{bmatrix}
         }{ 
         \| \exp\begin{bmatrix}
           0 &
           \alpha_{1,j}(x) &
           \hdots &
           \alpha_{n,j}(x)
         \end{bmatrix} \|_{1}},
 % \textbf{A} \propto \exp( \textbf{l} )
\end{equation}

%(\textcolor{red}{Raimon says: self-redundant equation? sum over no indices? seems connected to criticisms in Eq. \ref{eq1}})- SOLVED
\noindent with $\exp(\cdot)$ the component-wise exponential function, resulting in a set of realistic mineral abundance maps (Figure \ref{simulated_abun}). Following the real geology exposed in Corta Atalaya, a degree of endmember variability was then introduced by splitting the abundance of three mineral groups (muscovite, chlorite and clay) into compositional endmembers, based on the position of the 2200, 2250 and 2160 nm absorption features respectively. This extended the number of phases in $\mathbf{A}$ from seven to ten, a realistic degree of complexity for geological outcrops. A pure endmember spectral library $\mathbf{E}$ assembled using spectra from the USGS \cite{kokaly2017usgs} was then used to derive a synthetic reflectance spectra $\mathbf{S}$ for each data point, assuming linear mixing,

\begin{equation}
\label{eq3}
    \mathbf{S} = \mathbf{E} \cdot \mathbf{A}.
\end{equation}

These synthetic reflectance spectra, and the mineral abundances used to derive them, are also included in the benchmark dataset, and could be used for testing e.g., endmember identification and unmixing methods.

\subsubsection{Degraded twin}

In reality, sensor noise and other unwanted effects (e.g., atmospheric and topographic distortions, coating, vegetation) mean that no dataset will contain perfect reflectance spectra. Hence, as a final step, the synthetic reflectance spectra were degraded to simulate realistic measurement, preprocessing and data-correction procedures. First, the reflectance spectra were converted to at-target radiance estimates using the two-light-source atmospheric model described by \cite{thiele2021novel} and the Oren-Nayar BRDF \cite{oren1994generalization}. Simulating the real acquisition procedure, these radiance data were projected onto 2D rasters using three different camera poses, and path-radiance added to the corresponding spectra proportional to the target-sensor distance, resulting in three at-sensor radiance rasters. For the LWIR dataset, light emitted by the target (and by air between the target and the sensor) was also calculated, noting that $emissivity = 1 - reflectance$ following Kirchhoff's law, and added to the at-sensor radiance. 

Each raster was then transformed according to the inverse of the sensor-specific lens calibration and converted to digital numbers by dividing by the lab-determined spectral calibration values. Sensor noise was added using dark-current data acquired during the acquisition of the real hyperspectral data, resulting in a set of three simulated raw rasters with realistic noise. 

A degraded synthetic reflectance dataset (Figure \ref{ds-tree}) was then derived by correcting the simulated raw data using the same routine as was applied to the real data (cf., Section IIIA). 

\subsection{Accompanying 2D data}

\begin{figure*}[thp]
\centering
\begin{subfigure}{0.49\textwidth}
    \centering
    \includegraphics[width=0.98\linewidth]{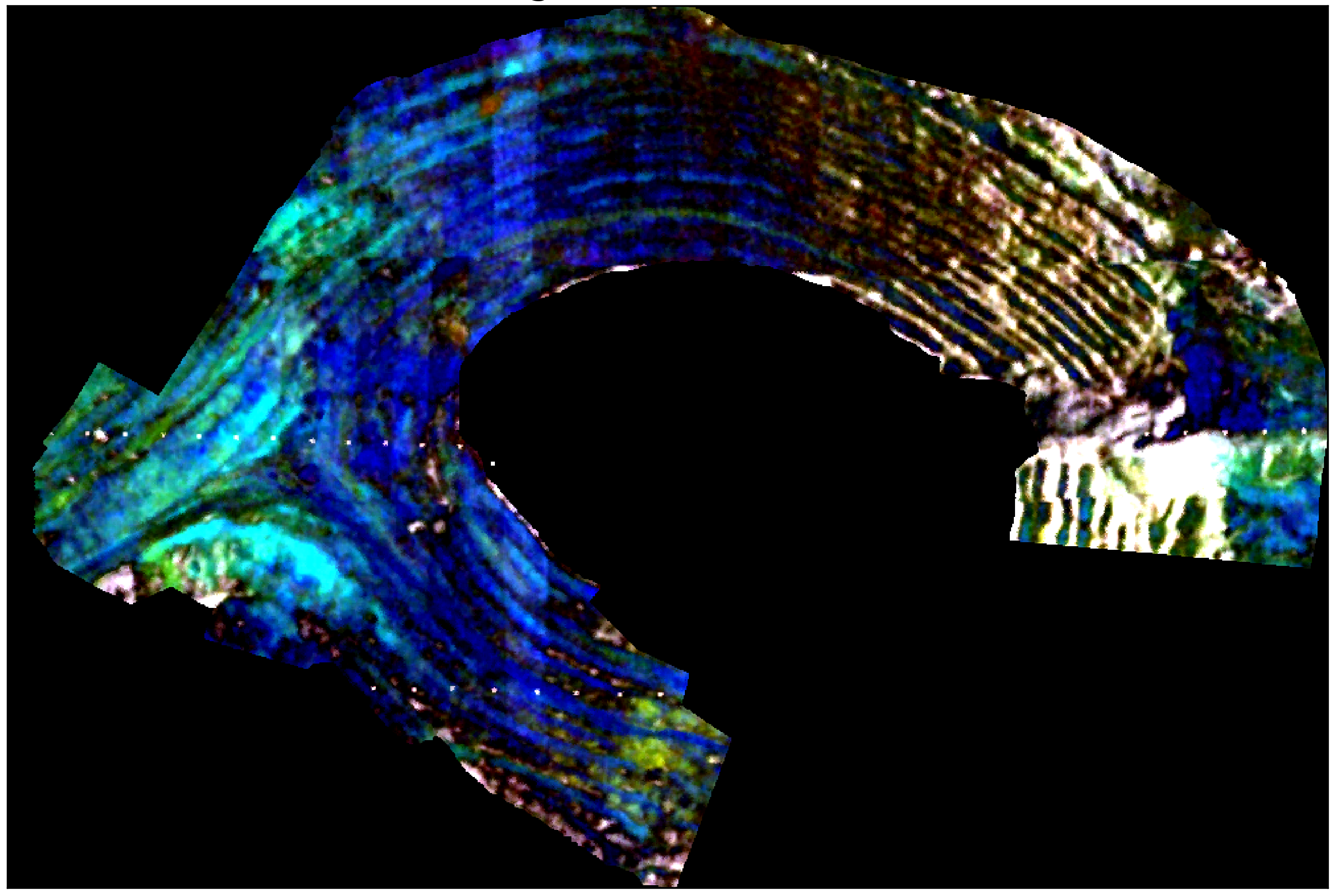} %
    \caption{View1: nadir orthoimage (LWIR: 10114.0, 9181.0, 8545.0 nm.)}
\end{subfigure}
\begin{subfigure}{0.49\textwidth}
    \centering
    \includegraphics[width=0.98\linewidth, height = 5.8 cm]{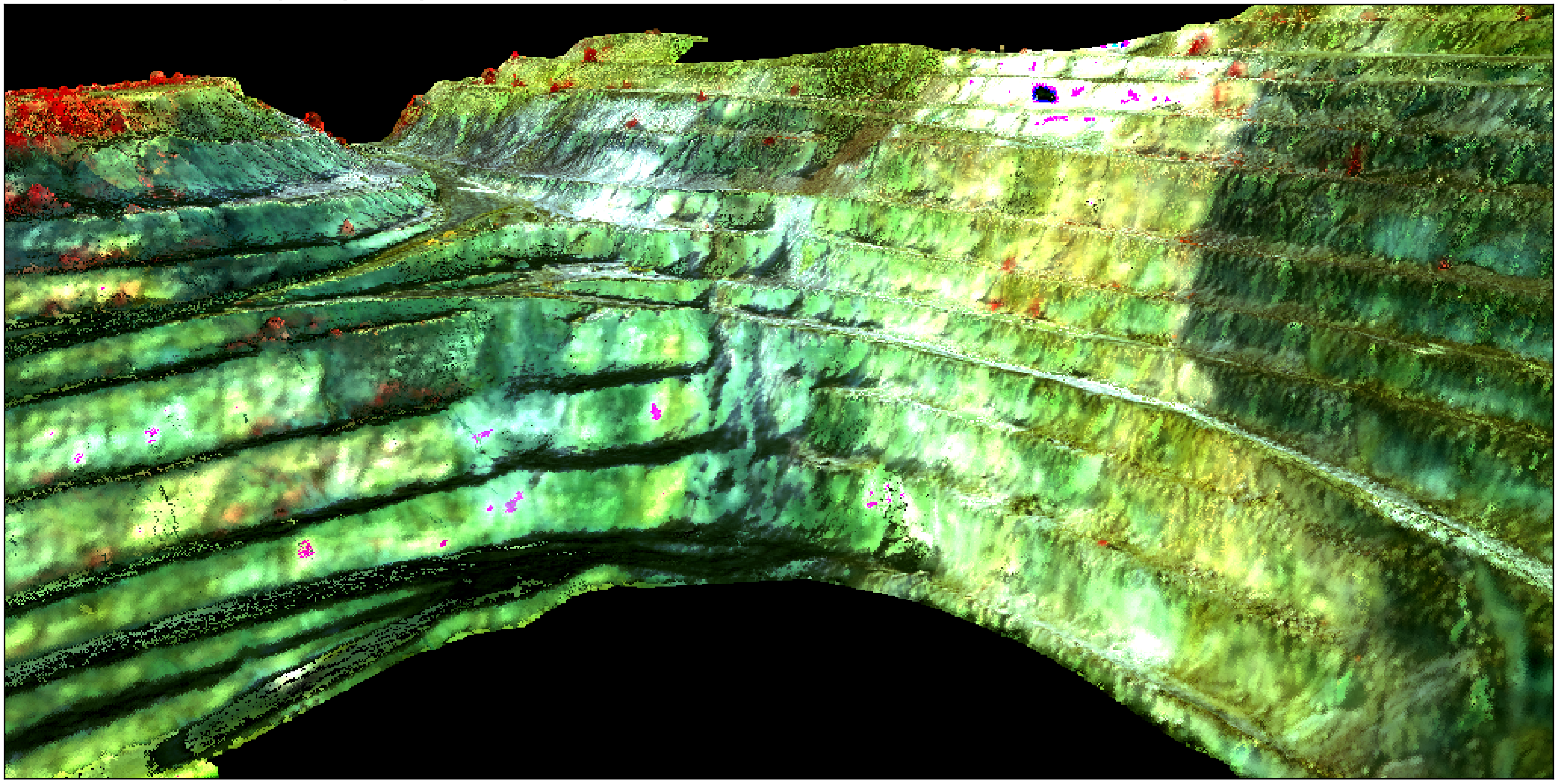} %
    \caption{View2: oblique perspective (VNIR: 850.0, 650.0, 525.0 nm).}
\end{subfigure}%
\vspace{\baselineskip}
\begin{subfigure}{0.98\textwidth}
    \centering
    \includegraphics[width=0.98\linewidth]{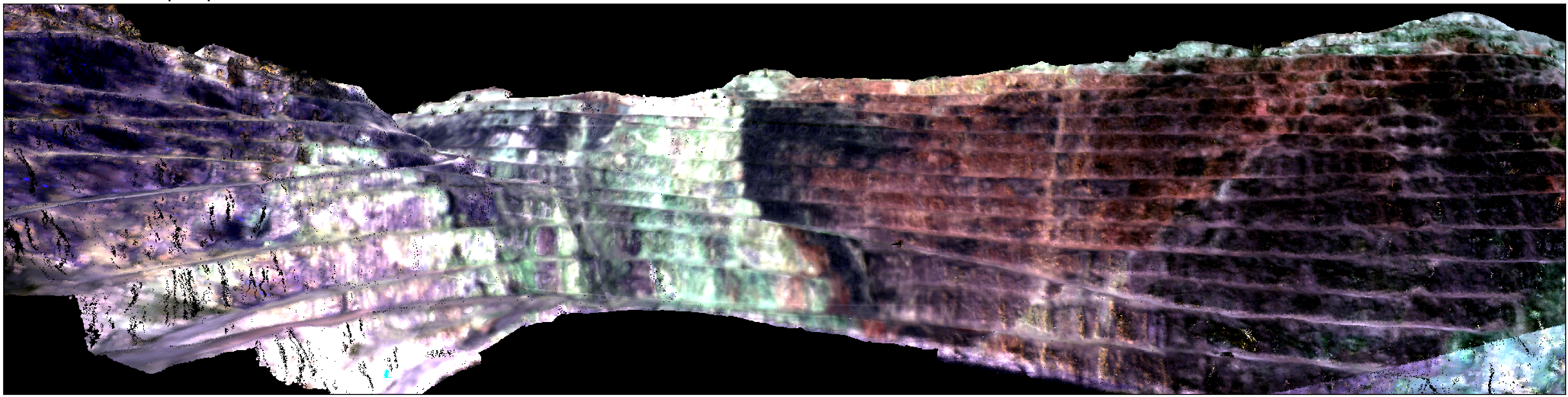} %
    \caption{View3: oblique panorama (SWIR: 2200.0, 2250.0, 2350.0 nm).}
\end{subfigure}%
\caption{False-colour visualisations of the real LWIR (top-left), VNIR (top-right) and SWIR (bottom) hyperspectral datasets from the three viewpoints used to derive the Tinto2D benchmark images.}
\label{false_color}
\end{figure*}

Although this manuscript focuses on 3-D point cloud data attributed with reflectance spectra to create hyperclouds, it is worth noting that we have included a set of 2-D rasters derived by projecting the class labels, real, synthetic and degraded spectra onto nadir, oblique perspective and oblique panoramic views (Figure \ref{false_color}). These will not be discussed further here, but could serve as a useful benchmark for image segmentation or unmixing methods.

\section{Evaluation}
\label{evaluation}
\subsection{Baseline Models}
Many deep learning models designed for processing raw point clouds are primarily focused on classification and segmentation tasks. Point coordinates are typically the most common input to the network, and in some cases, normals and RGB values can also be incorporated. Some models proposed sophisticated layers to effectively process point clouds directly \cite{wu2019pointconv, li2018pointcnn, cai2019spatial}. They learn the geometrical features of the points to perform the classification and segmentation tasks. Tinto dataset includes multiple sources of information, but the emphasis is on learning hyperspectral information for point cloud segmentation. To evaluate the Tinto dataset on deep learning models, different deep learning models designed for point cloud processing were selected and trained from scratch. They are categorized into three different architectures according to the main building layers. The first category is the multi-layer perceptron (MLP) based models \cite{popescu2009multilayer} where the MLPs are the main building layers in the models. The second category is the convolutional-based models where convolutional-like layers are used to process the point clouds. The third category is the Transformers-based models where the Natural Language Processing (NLP) Transformer models are adopted and modified to process raw point clouds for different tasks. Following, we will discuss each category and explain the models related to the category.
%the   was the first model implemented and tested on the dataset. Despite being simpler compared to other advanced deep models proposed for point cloud processing, the MLP performed comparably or even outperformed other models in some scenarios. The second model was PointNet \cite{qi2017pointnet}, the first-ever model proposed for point cloud processing, while the third model was PointNet$++$ \cite{qi2017pointnet2}, its successor capable of extracting the local features of the point cloud. All the baseline models were modified and implemented to accept the hyperspectral information of a point as input for the segmentation task.

\subsubsection{MLP-based Models}
MLP-based models for point cloud processing refer to architectures that utilize MLPs as their fundamental building blocks for analyzing and extracting information from 3D point cloud data. The first baseline model is a 10-layer MLP. The model was implemented to classify the point cloud according to its hyperspectral information. The output layer is equal to the number of classes. Our network consists of $10$ hidden layers between the input and the output layers to extract useful features and help in the classification process. The hidden layers have to extract features of different sizes. The input layer takes the hyperspectral information of the point as input information. The output layer has the same size as the number of classes in the ground-truth. Another MLP-based model is the PointNet model \cite{qi2017pointnet}. It is the first pioneering deep learning model for direct point cloud processing. As the points are unordered in the point cloud data, it processes each point in isolation through a shared MLP to extract local features. Specifically, PointNet applies point-wise operations using several MLP layers to extract independent features separately and uses max-pooling operation to capture the global features of the point cloud. The aggregated global feature extracted from the point cloud can be used for various tasks such as classification and segmentation. The drawback of the PointNet model is that features are learned independently and then the global feature is aggregated. So, the local structure of the point cloud between points is not captured. To overcome this limitation, PointNet$++$ \cite{qi2017pointnet2} was proposed as a hierarchical network. PointNet$++$ consists of three main layers: the sampling layer, the grouping layer, and the PointNet-based learning layer. The sampling layer uses the farthest point sampling algorithm to select centroids. The grouping layer uses the selected centroids to find the nearest neighbor points of each centroid. The PointNet layer is then applied on the local region to learn and extract the feature vector. This process is repeated in a hierarchical form and the points' resolution is reduced as the network goes deeper. In the last layer, the global feature is produced. However, the calculated KNN is not updated as the input goes deeper into the network. The Dynamic Graph CNN (DGCNN) \cite{wang2019dynamic} creates dynamic graphs that capture the relationships between points. It utilizes EdgeConv layers to extract the characteristics of individual points while employing non-linear MLPs on neighboring points. Subsequently, the edges are iteratively adjusted in subsequent layers based on the input from prior layers, creating a dynamic workflow.

\subsubsection{Convolutional-based Models}
The success of CNNs for image classification inspired researchers to use a custom convolutional-like operation for point clouds. PointCNN \cite{li2018pointcnn} is designed to process the point cloud with the ability to efficiently learn and capture patterns from point clouds without relying on predefined grids or structures. It achieves this by introducing a unique convolution operation called X-Conv that adapts to the local geometry of the points by learning the weights of the input features and then permuting the points into canonical order, making it highly effective for tasks like 3D object classification and segmentation. ConvPoint \cite{BOULCH202024} is another model where a customized continuous convolution operation was introduced to process the unordered point clouds. This operation can be extended easily to build a CNN model to process the point clouds similar to 2D CNNs. The convolution operation separates the kernel into spatial and feature parts. A unit sphere is used to select the location of the spacial parts randomly and the weighted function of the layers is learned by a simple MLP.

These proposed approaches and others underscore the adaptability and effectiveness of custom convolutional-like operations in extracting meaningful features from point cloud data, further expanding the horizons of 3D data analysis.

\subsubsection{Transformer-based Models}
Inspired by the popularity of the transformer models in NLP, researchers proposed different Transformer models to process the point clouds. It is under the assumption that the point cloud format suits the self-attention operator due to invariance to permutation and cardinality. Point Transformer (PT) \cite{zhao2021point} proposed a transformer model with self-attention layers as a set operator that can process the point clouds for various tasks. The attention layers learn the relationship between selected central points and the corresponding neighboring points. Those attention layers serve as the backbone for the feature encoder block, which gradually downsamples the number of points in each consecutive layer. Point Cloud Transformer (PCT) \cite{guo2021pct} is permutation invariant and it enhances the input embedding by applying the farthest point sampling for centroids selecting and nearest neighbors calculating. PCT proposed an attention mechanism where the final output features from the attention layer are the offset features that are the difference between the input and the original attention features.

In summary, Transformer-based models tailored for point clouds leverage the self-attention mechanism to exploit the unique characteristics of point cloud data, demonstrating promising potential for a wide range of applications in 3D perception and analysis.
      
\subsection{Implementation Details}
We implemented, trained, and evaluated the MLP model on TensorFlow \cite{abadi2016tensorflow}. The weights of the model were initialized using the Xavier initialization method \cite{glorot2010understanding}. For the remaining models, we used the original implementation codes from the GitHub repositories, all with Adam optimizer \cite{kingma2014adam}, except DGCNN and PCT used Stochastic Gradient Descent (SGD) \cite{10.5555/3042817.3043064} with the momentum of $0.9$. The dataset contains $3,187,785$ points (excluding the Vegetation class). For the complete label scenario, the dataset is split into a training set ($297,968$ points) and a testing set ($2,889,817$ points) with the ratio of $10\%$ and $90\%$, respectively. The inputs were the hyperspectral information of the points. The input size depends on the sensor used to acquire the data (VNIR = $51$ bands, SWIR = $141$ bands, and LWIR = $126$ bands). The learning rate was set to $0.001$, except PCT used $0.1$. All models used a batch size of $16$, except PointNet and PointNet$++$ used $24$ and DGCNN used $32$. The models were implemented in PyTorch \cite{paszke2019pytorch}, and only PointCNN was applied in TensorFlow.

% The weights of the models were initialized using the Xavier initialization method \cite{glorot2010understanding}. We trained the models from scratch using an Adam optimizer \cite{kingma2014adam} to optimize the network parameters with the following settings: the learning rate was set to $0.01$, the momentum was set to $0.9$, and the mini-batch size was $128$ points. We trained the models until the validation accuracy stopped increasing. We excluded the Vegetation class during training and testing.

\subsection{Experimental Results}

\begin{table*}[!hpbt]
    \caption{Performance of the baseline models on the Tinto dataset.}
    \label{results}
    \begin{center}
    \begin{tabular}{|l|c|c|c|c|c|c|c|c|c|}
      \hline
      \multirow{2}{*}{Model} & \multicolumn{3}{c|}{Real Data} & \multicolumn{3}{c|}{Clean Synthetic Data} & \multicolumn{3}{c|}{Noisy Synthetic Data} \\ \cline{2-10}
      %\hline
      & LWIR & SWIR & VNIR & LWIR & SWIR & VNIR &LWIR & SWIR & VNIR \\
      \hline
 %     MLP & 55.8 / 38.8 & 83.6 / 71.5 &74.7 / 60.4 & 98.9 / 92.7 & 99.0 / 93.6 & 98.9 / 90.5 & 87.7 / 75.1 & 86.7 / 77.2 & 84.6 / 73.1 \\
      MLP & 38.8 & 71.5 & 60.4 & 92.7 & 93.6 & 90.5 & 75.1 & 77.2 & 73.1 \\
      \hline
%      PointNet & 55.1 / 39.7 & 80.1/ 66.3 & 72.2 / 59.0 & 98.8 / 93.1 & 99.0 / 93.4 & 98.8 / 93.4 & 87.5 / 77.3 & 86.8 / 77.6 & 79.0 / 74.6 \\
      PointNet \cite{qi2017pointnet} & 43.2 & 67.7 & \textbf{62.1} & 96.7 & 96.4 & 96.0 & 82.5 & 84.9 & 83.2 \\
      \hline
%      PointNet$++$ & 47.6 / 30.3 & 76.6 / 62.6 & 67.5 / 56.3 & 97.5 / 94.7 & 98.6 / 96.0 & 90.6 / 82.3 & 82.5 / 72.0 & 81.5 / 70.4 & 72.3 / 60.4 \\
%      PointNet$++$ & 30.3 & 62.6 & 56.3 & 94.7 & 96.0 & 82.3 & 72.0 & 70.4 & 60.4 \\
      PointNet$++$ \cite{qi2017pointnet2} & 43.0 & 71.5 & 60.7 & 96.8 & 95.9 & 96.2 & \textbf{83.9} & 85.3 & 83.1 \\
      \hline
       DGCNN \cite{wang2019dynamic} & 42.7 & 68.9 & 59.1 & \textbf{97.1} & \textbf{96.9} & \textbf{96.8} & 83.5 & \textbf{86.0} & \textbf{83.5} \\
      \hline
      PointCNN \cite{li2018pointcnn} & \textbf{47.3} & 59.7 & 58.0 & 90.1 & 95.0 & 87.7 & 76.9 & 69.7 & 69.3  \\
      \hline
      ConvPoint \cite{BOULCH202024} & 45.1 & 67.7 & 55.4 & 92.9 & 89.7 & 92.9 & 76.8 & 77.3 & 74.1 \\
      \hline
      PT \cite{zhao2021point} & 45.3 & 60.2 & 51.2 & 84.7 & 95.3 & 80.1 & 64.9 & 80.4 & 71.0 \\
      \hline
      PCT \cite{guo2021pct} & 43.3 & \textbf{71.8} & 61.2 & 96.9 & 96.4 & 94.0 & 82.6 & 85.2 & 82.8 \\
      \hline
    \end{tabular}
    \end{center}
\end{table*}

We conducted experiments on the testing split of the Tinto dataset to assess the performance of the baseline models. All hyperspectral point clouds of the VNIR, SWIR, and LWIR (the real, clean synthetic, and noisy synthetic) are utilized for the quantitative and qualitative evaluation. The baseline models trained and tested on the Tinto dataset are MLP, PointNet \cite{qi2017pointnet}, PointNet$++$ \cite{qi2017pointnet2}, DGCNN \cite{wang2019dynamic}, PointCNN \cite{li2018pointcnn}, ConvPoint \cite{BOULCH202024}, PT \cite{zhao2021point}, and PCT \cite{guo2021pct}.

Firstly, we trained the baseline models separately on each sensor data using the training set of real data with complete labels. The trained models were then evaluated on the testing set and their accuracies were computed. Table \ref{results} (Real Data) reports the overall accuracy of the baseline models on the real data for the complete labels. The results indicated that the PointCNN model achieved the highest accuracy on the LWIR data, the PCT model achieved the highest accuracy on the SWIR data, and the PointNet model achieved the highest accuracy on the VNIR data. Most baseline models were proposed and designed to capture the geometric information from 3D point clouds of shapes and performed well with objects and scenes that can be segmented into grids. These models mainly trained on the coordinates of the points and other information (e.g., normals and RGB values) as input features to extract geometric information from the point cloud. However, we modified the baseline models and trained them on the hyperspectral data only. Their performance is lower on this dataset compared to the models' performance on other datasets. Moreover, it's worth mentioning that the accuracy of the labels associated with the real data is not guaranteed to be 100\%, which can impact the models' performance to some extent. This is because of the nature of the dataset as there are no sharp boundaries between the classes as the rocks are highly overlapped in reality.

To address the issue of inaccurate ground truth labels, the dataset has a synthetic part where each point in the point cloud has a synthetic hyperspectral feature and is associated with a correct class label. We trained the baseline models on the training set and evaluated them on the testing set of various sensors. Table \ref{results} (Clean Synthetic Data) reports the overall accuracy of the baseline models on the clean synthetic data for the complete labels. We found that the performance of all models on the clean synthetic data scored higher accuracies compared to the performance on the real data with a large margin. Interestingly, the majority of models performed similarly on clean synthetic data from different hyperspectral sensors. Our observation is that precise ground truth data can greatly enhance model performance and enable accurate segmentation of the point cloud. Also, Table \ref{results} shows that the models that consider the neighboring points when extracting the features scored the best accuracies. The DGCNN model, which relies on the local neighboring points to create and update the graph for learning the point features, outperforms other models on all sensors data. Then, it is followed by PCT and PointNet$++$ models. Both models learn the local features by considering the neighboring points and merge them with the global features computed by the self-attention layer for PCT and the max-pooling layer for PointNet$++$. Our experiment demonstrated that those models are better suited for our synthetic dataset.

%The DGCNN model outperforms other models on all sensors’ data as it extracts the features of each point by considering the neighboring points and update the graph during the training process. Furthermore, PCT and PointNet$++$ models outperform other models due to their incorporation of neighboring points for extracting both local and global features during training. Therefore, it holds great significance to take into account the neighboring points within this dataset in order to achieve improved performance.

In order to increase the realism of the synthetic data and challenge the models further, we added real noise information (sensor noise and processing artifacts) to the synthetic point cloud. We evaluated the baseline models on the noisy synthetic data and found that their accuracy decreased compared to those trained on clean synthetic data. Table \ref{results} (Noisy Synthetic Data) reports the overall accuracy of the baseline models on the noisy synthetic data for the basic and complete labels. While all models achieved higher performance compared to the models trained on the real data, DGCNN and PointNet$++$ models are consistently the leading models on the synthetic data. The DGCNN model achieved the highest accuracy on the SWIR and VNIR data while PointNet$++$ achieved the highest accuracy on the LWIR data.

%While all models achieved higher performance compared to the models trained on the real data, the models utilizing the neighboring points outperform other models as they extract local and global features to segment the point cloud on all sensors’ data. The DGCNN model achieved the highest accuracy on the SWIR and VNIR data while PointNet$++$ achieved the highest accuracy on the LWIR data.

In conclusion, the baseline models proved that they can be adapted to learn hyperspectral information and perform the point cloud segmentation task on the geological data. This opens a new direction of applying deep learning models to generate segmented maps on the geological data using hyperspectral information and propose new models that can fuse information from different sources. 

\begin{figure*}[!htpb]
\centering
\begin{subfigure}{0.25\textwidth}
    \centering
    \includegraphics[width=0.98\linewidth, trim = {0.5cm 0.5cm 0.5cm 0.5cm}, clip]{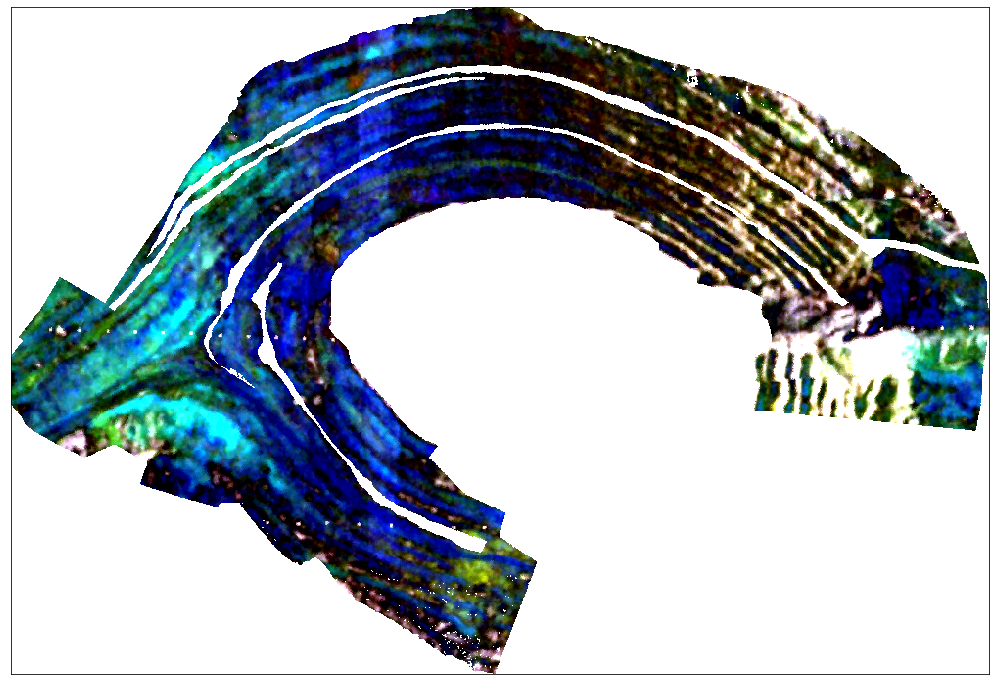}
    \caption{LWIR}
\end{subfigure}%
\hspace{0.5 cm}
\begin{subfigure}{0.25\textwidth}
    \centering
    \includegraphics[width=0.98\linewidth, trim = {0.5cm 0.5cm 0.5cm 0.5cm}, clip]{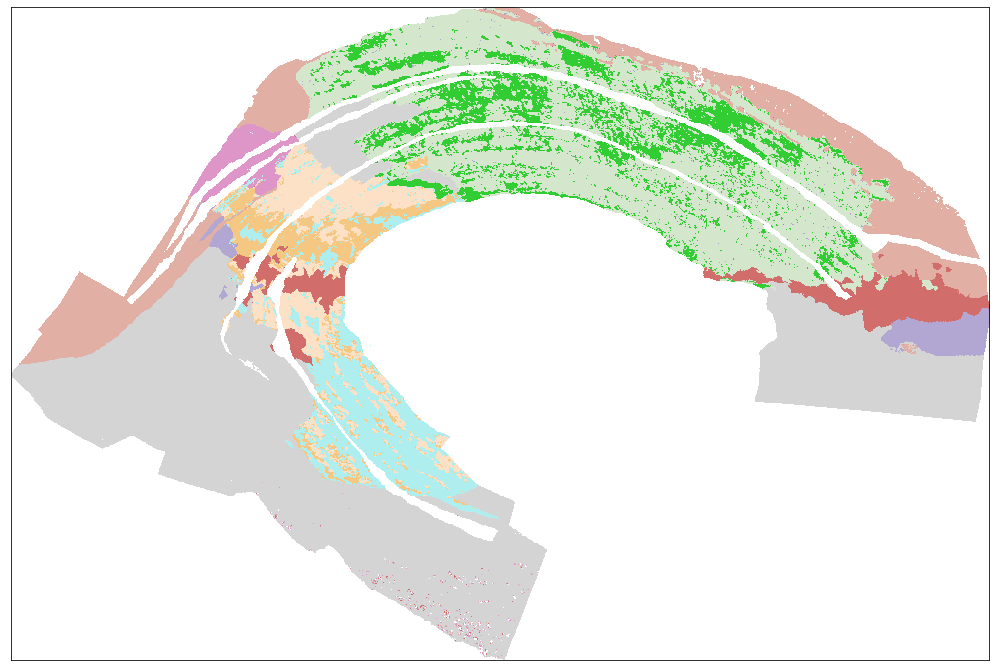} %
    \caption{ground truth}
\end{subfigure}

\begin{subfigure}{0.24\textwidth}
    \centering
    \includegraphics[width=0.98\linewidth, trim = {0.5cm 0.5cm 0.5cm 0.5cm}, clip]{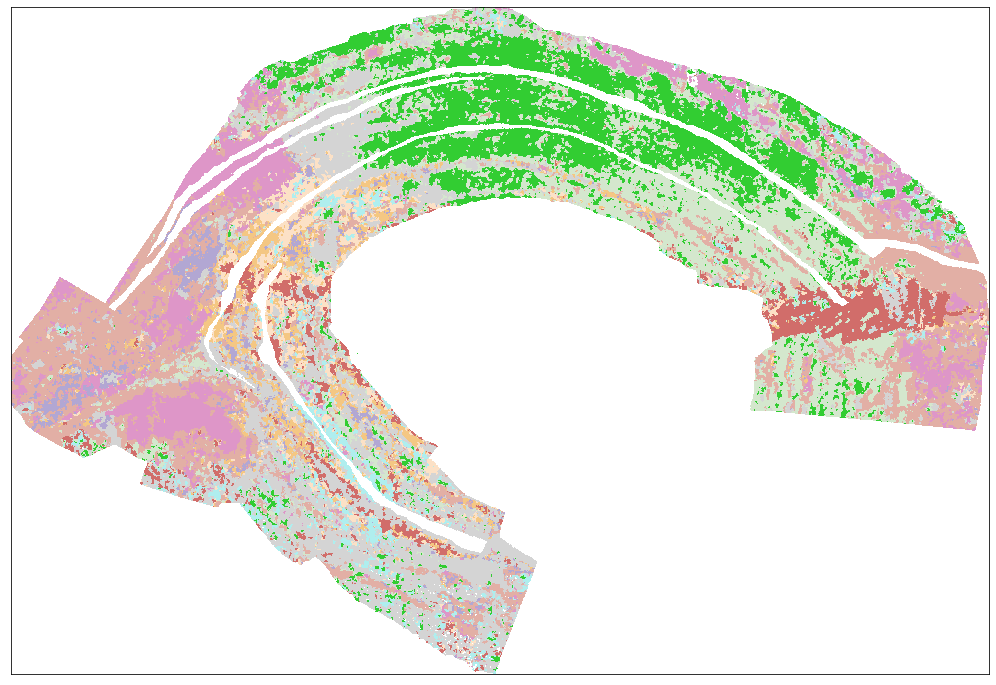} %
    \caption{LWIR MLP real}
\end{subfigure}%
\begin{subfigure}{0.24\textwidth}
    \centering
    \includegraphics[width=0.98\linewidth, trim = {0.5cm 0.5cm 0.5cm 0.5cm}, clip]{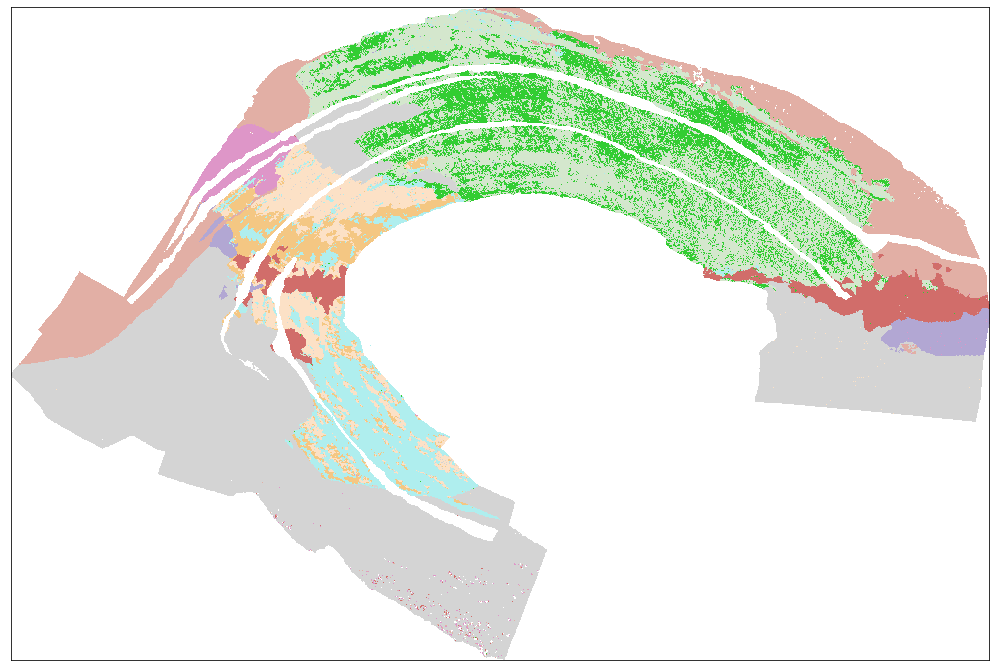}
    \caption{LWIR MLP clean}
\end{subfigure}%
\begin{subfigure}{0.24\textwidth}
    \centering
    \includegraphics[width=0.98\linewidth, trim = {0.5cm 0.5cm 0.5cm 0.5cm}, clip]{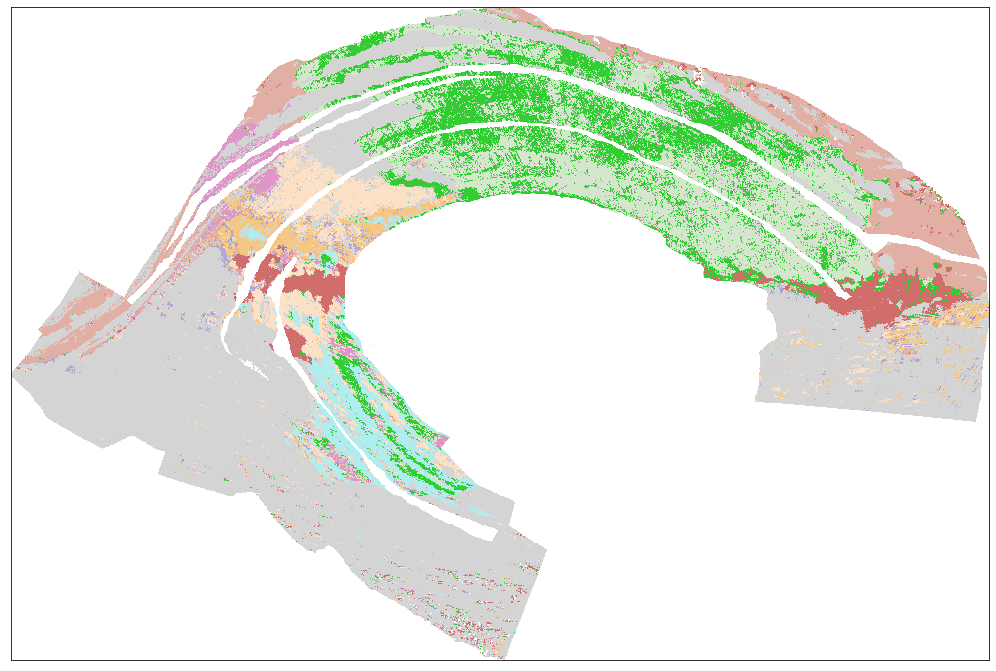}
    \caption{LWIR MLP noisy}
\end{subfigure}%
\begin{subfigure}{0.24\textwidth}
    \centering
    \includegraphics[width=0.98\linewidth]{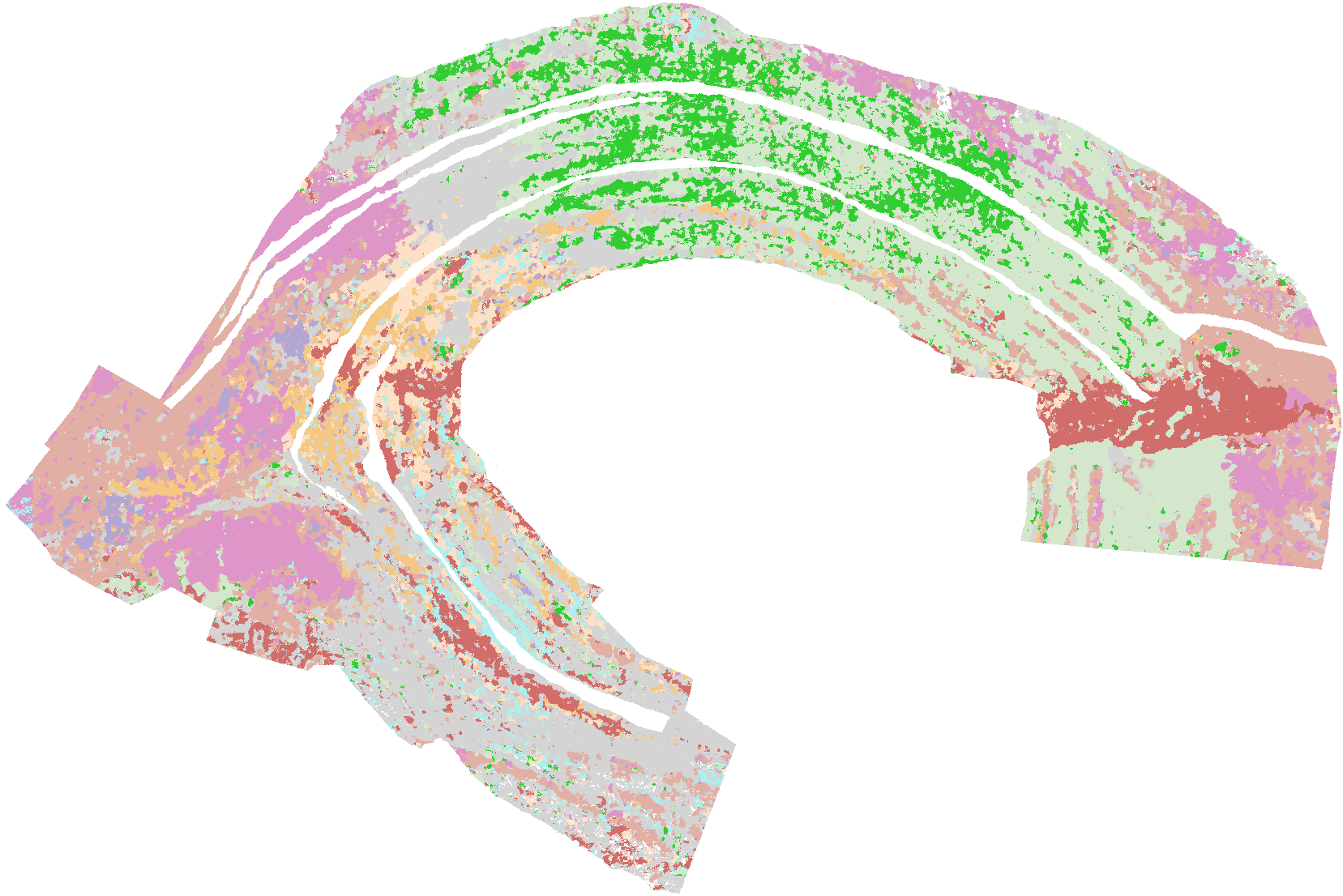} %
    \caption{LWIR PointNet real}
\end{subfigure}%

\begin{subfigure}{0.24\textwidth}
    \centering
    \includegraphics[width=0.98\linewidth]{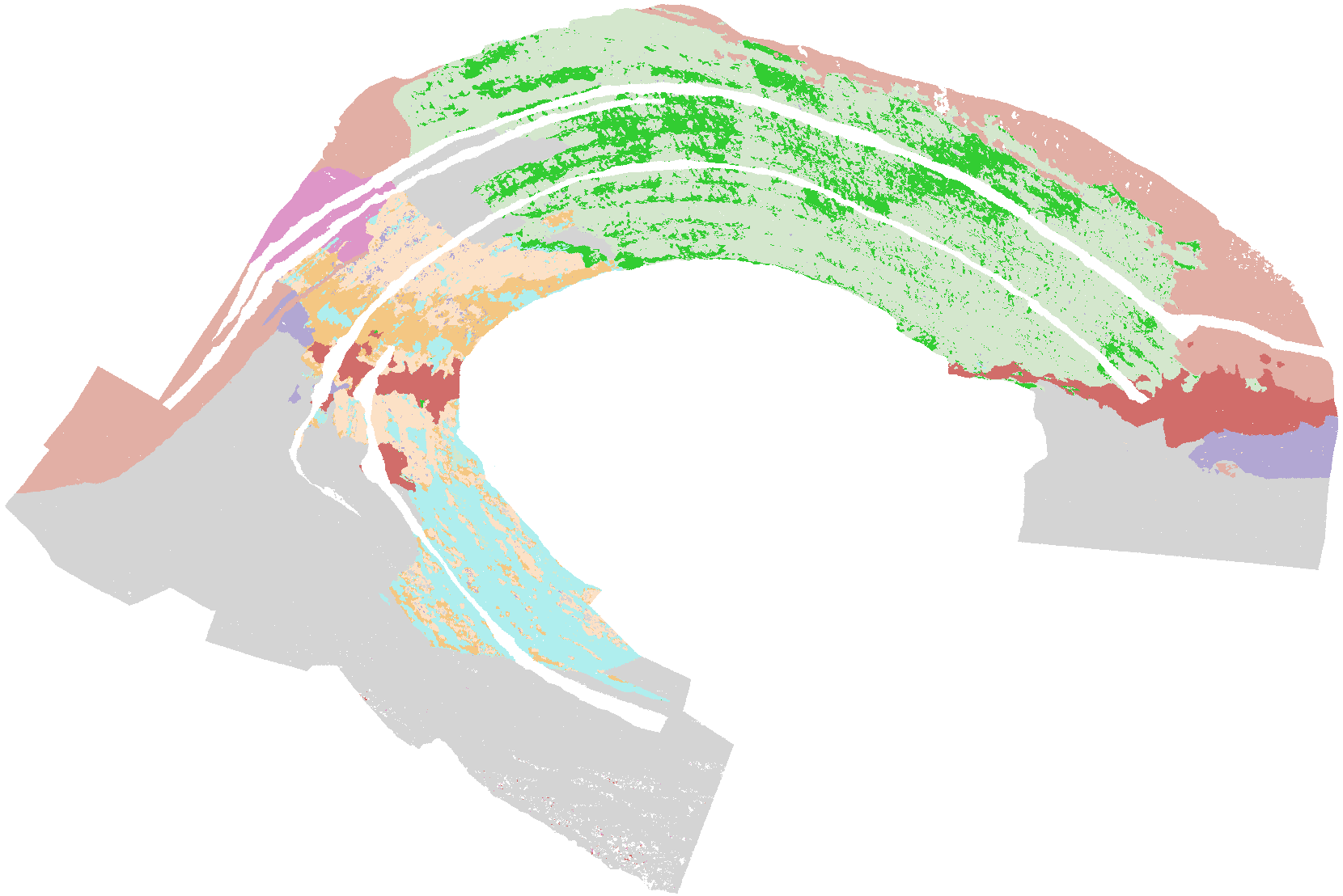}
    \caption{LWIR PointNet clean}
\end{subfigure}%
\begin{subfigure}{0.24\textwidth}
    \centering
    \includegraphics[width=0.98\linewidth]{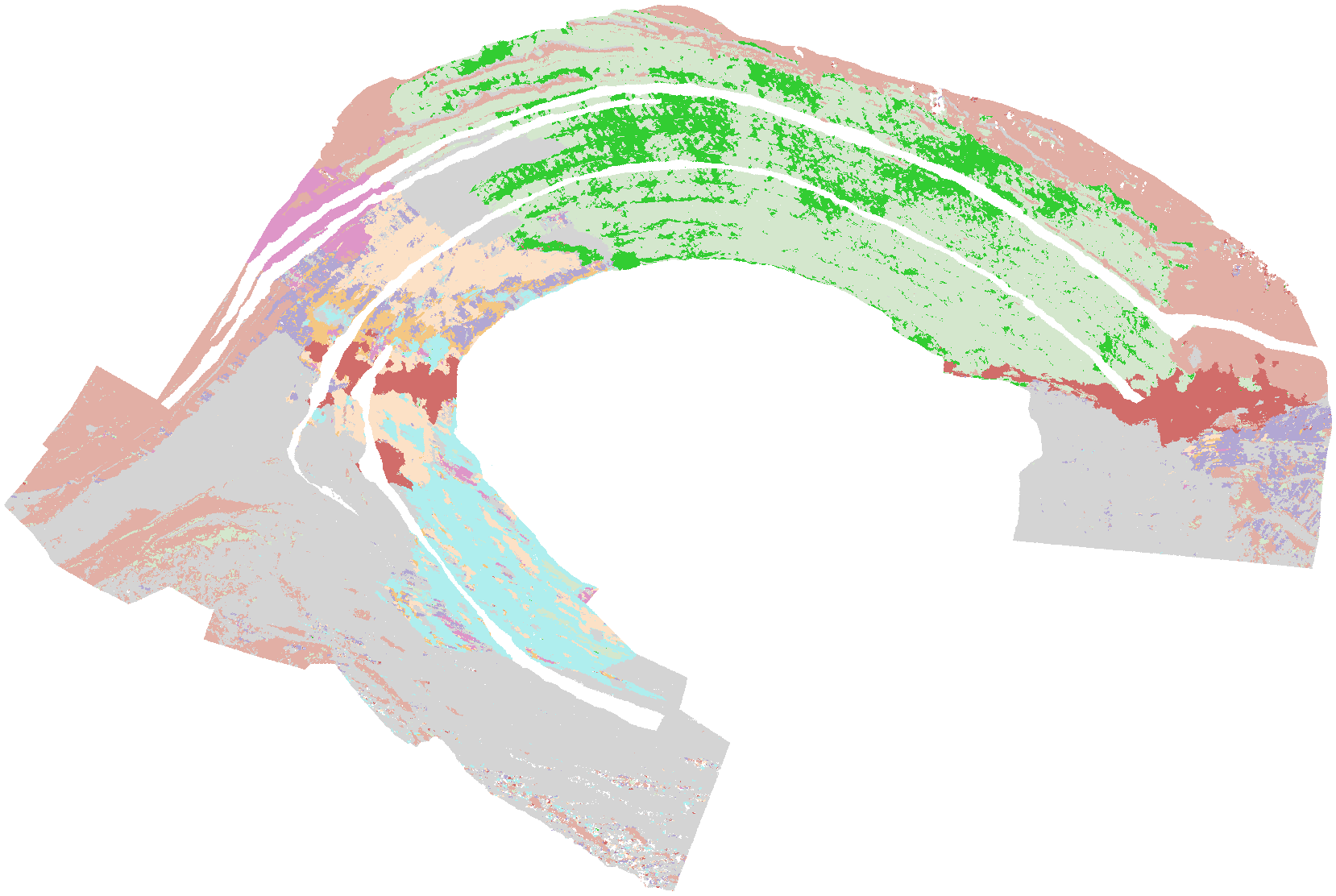}
    \caption{LWIR PointNet noisy}
\end{subfigure}%
\begin{subfigure}{0.24\textwidth}
    \centering
    \includegraphics[width=0.98\linewidth]{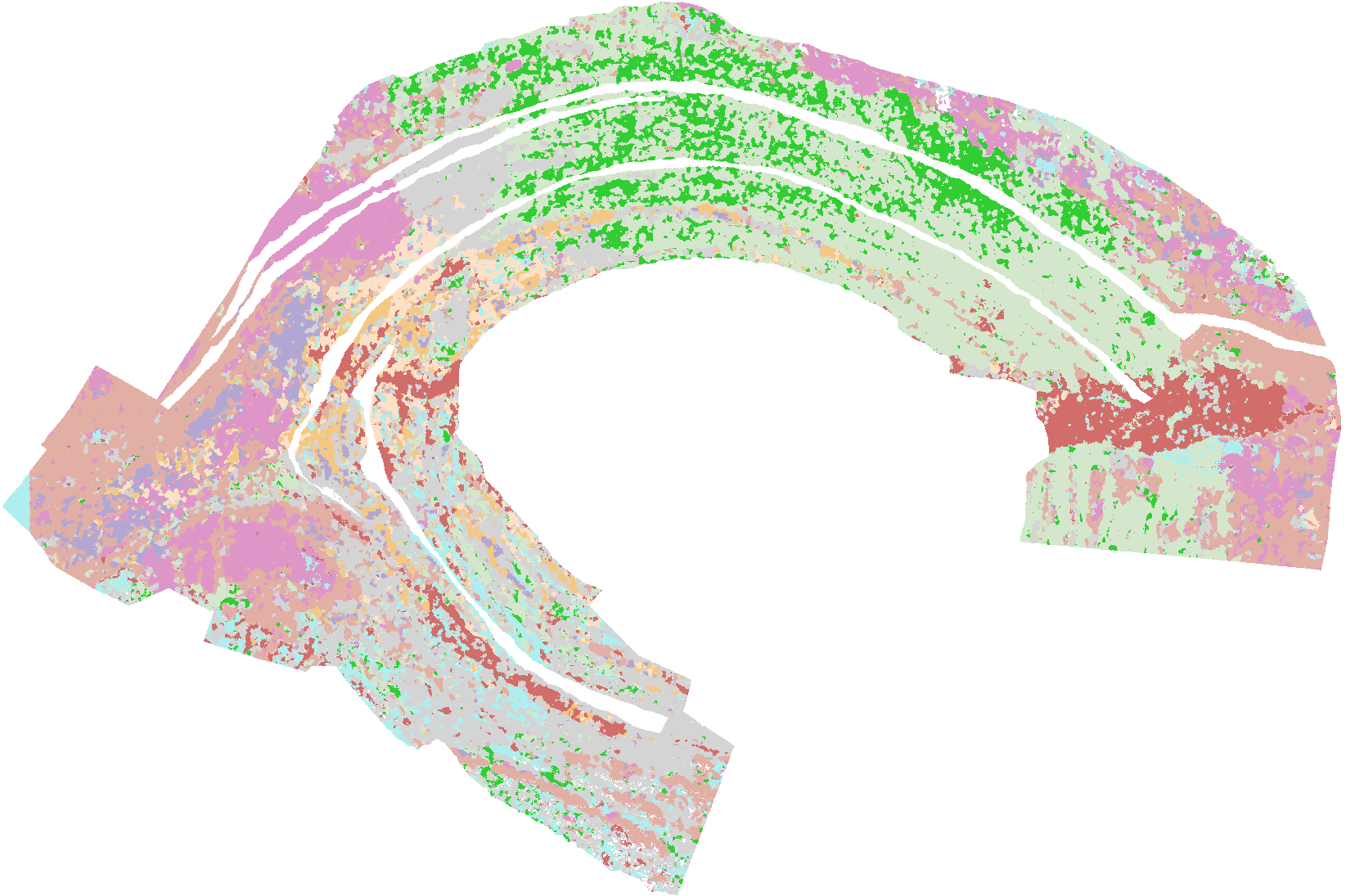} %
    \caption{LWIR PointNet2 real}
\end{subfigure}%
\begin{subfigure}{0.24\textwidth}
    \centering
    \includegraphics[width=0.98\linewidth]{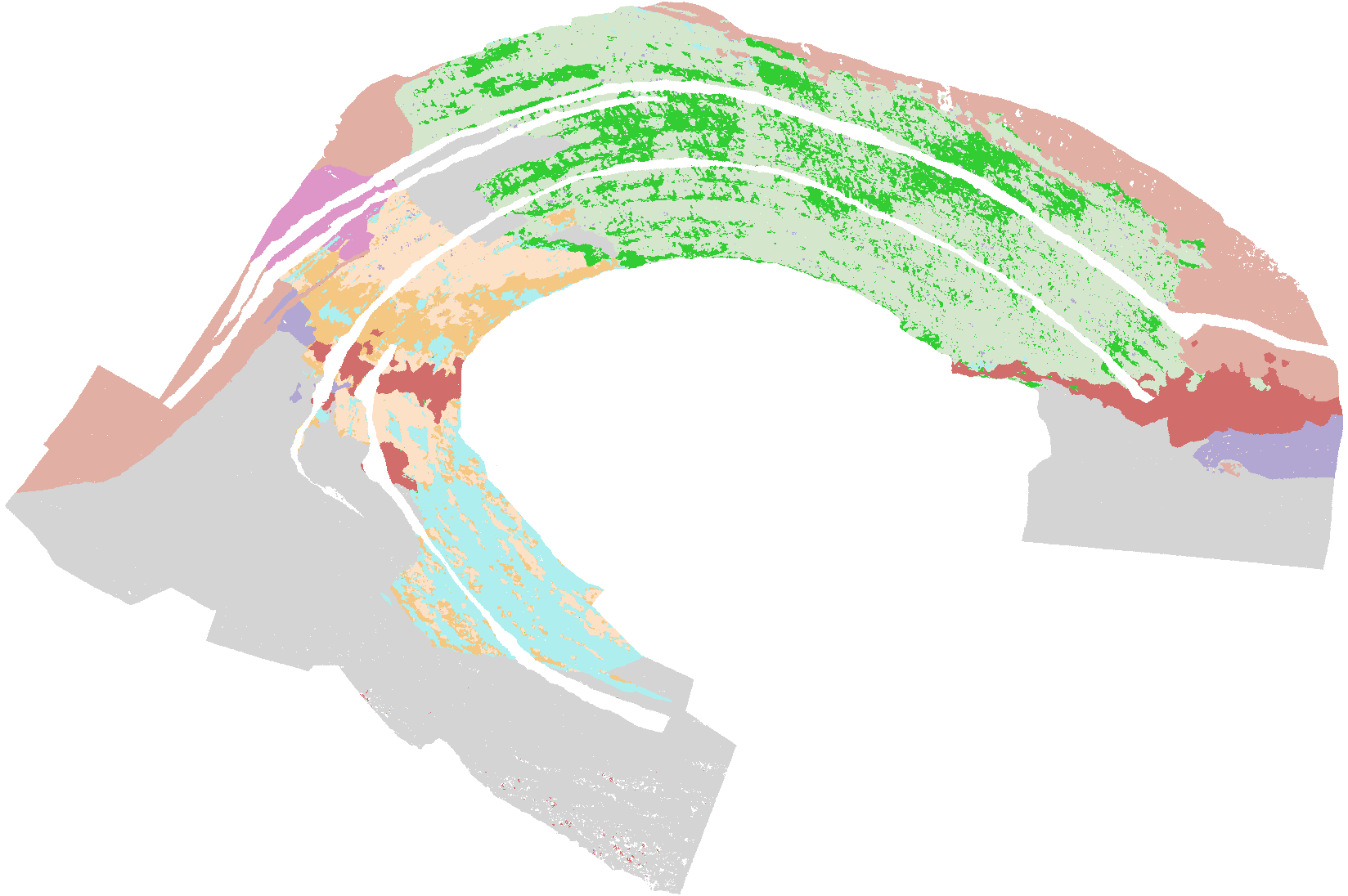}
    \caption{LWIR PointNet2 clean}
\end{subfigure}%

\begin{subfigure}{0.24\textwidth}
    \centering
    \includegraphics[width=0.98\linewidth]{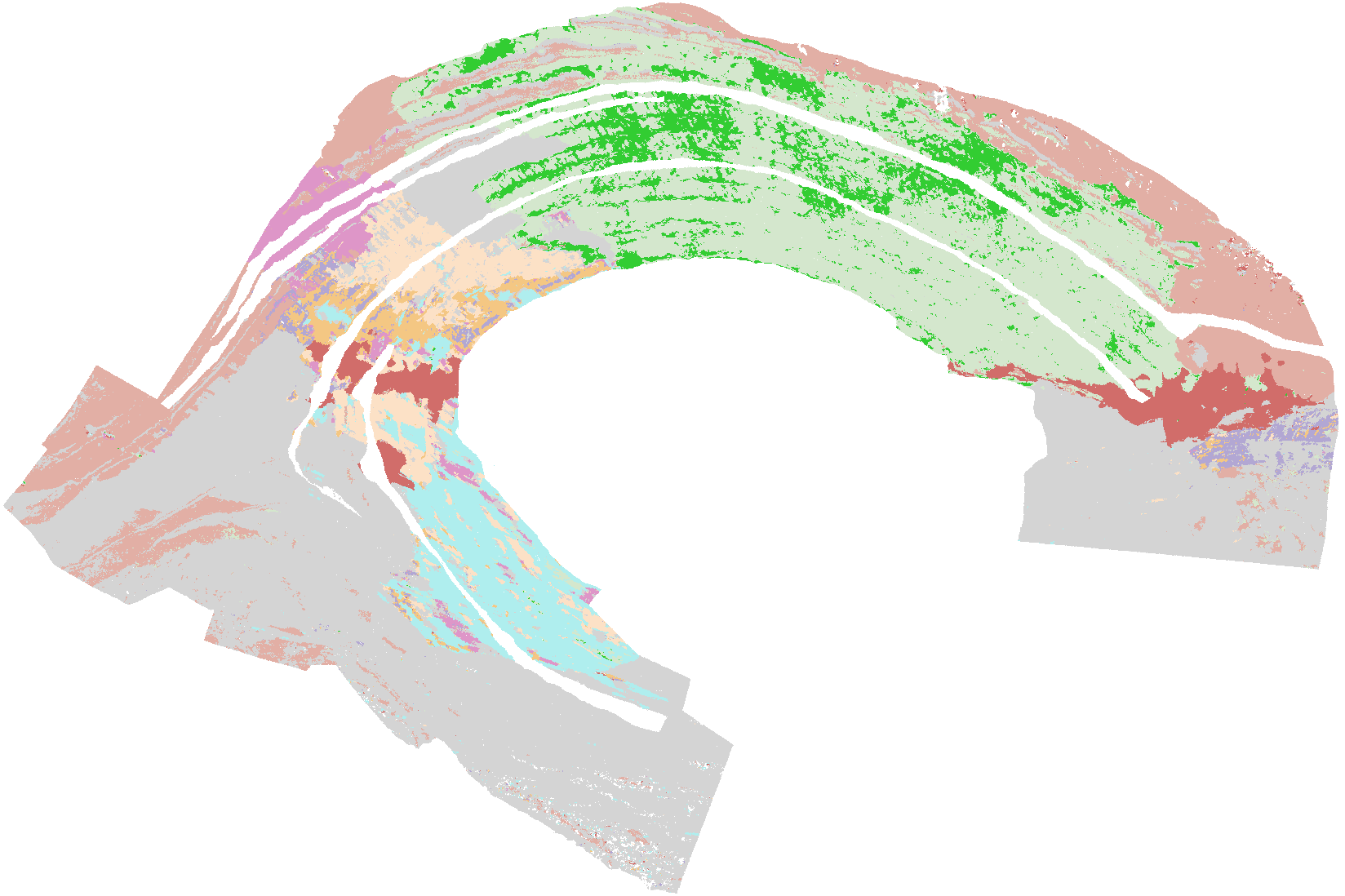}
    \caption{LWIR PointNet2 noisy}
\end{subfigure}%
\begin{subfigure}{0.24\textwidth}
    \centering
    \includegraphics[width=0.98\linewidth]{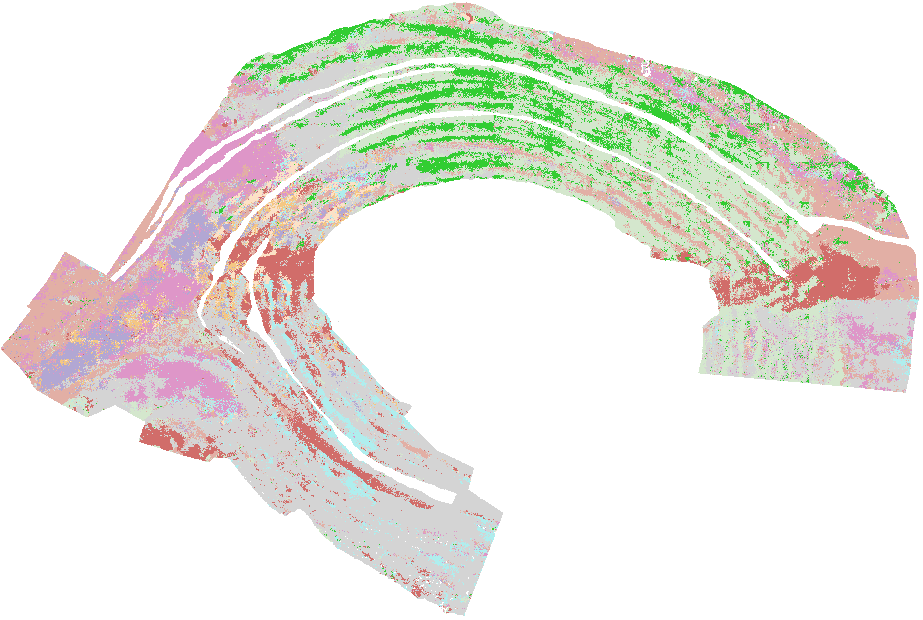} %
    \caption{LWIR PointCNN real}
\end{subfigure}%
\begin{subfigure}{0.24\textwidth}
    \centering
    \includegraphics[width=0.98\linewidth]{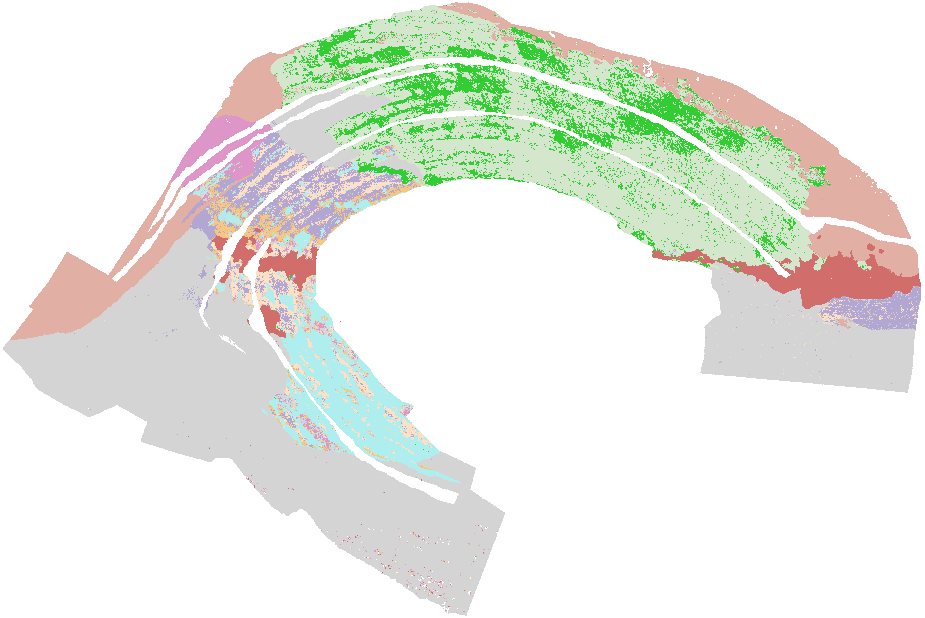} %
    \caption{LWIR PointCNN clean}
\end{subfigure}%
\begin{subfigure}{0.24\textwidth}
    \centering
    \includegraphics[width=0.98\linewidth]{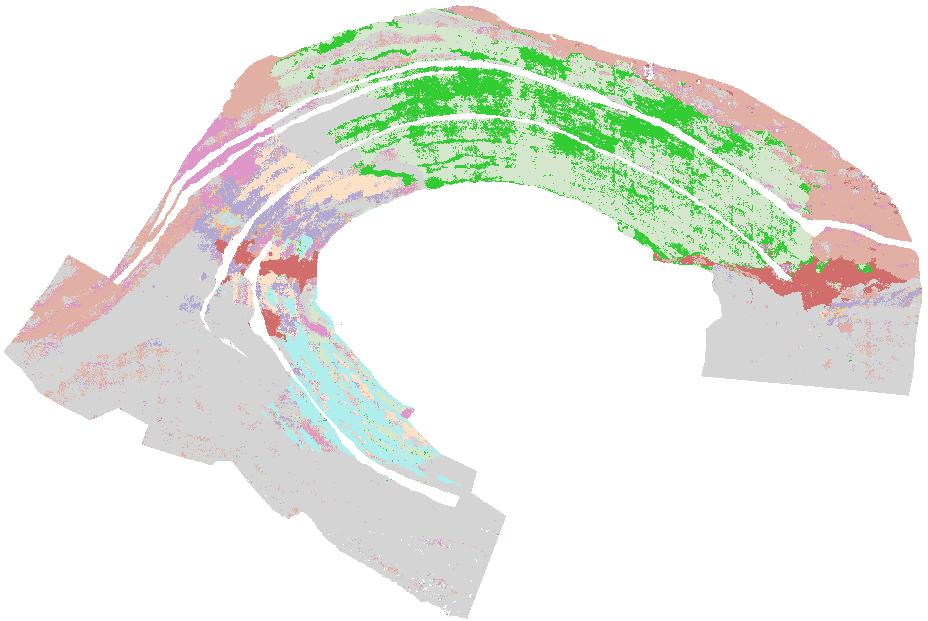} %
    \caption{LWIR PointCNN noisy}
\end{subfigure}%

\begin{subfigure}{0.24\textwidth}
    \centering
    \includegraphics[width=0.98\linewidth]{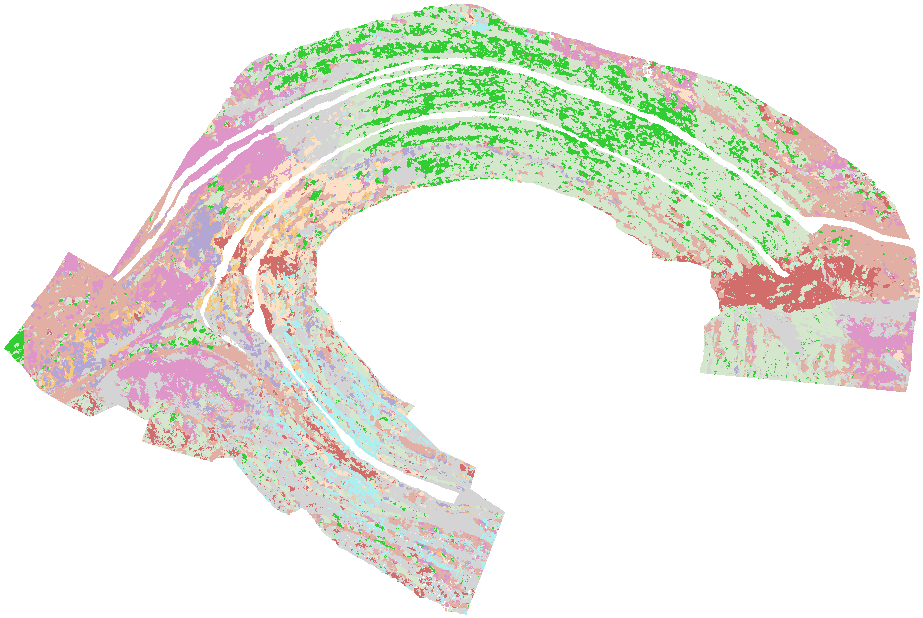} %
    \caption{LWIR ConvPoint real}
\end{subfigure}%
\begin{subfigure}{0.24\textwidth}
    \centering
    \includegraphics[width=0.98\linewidth]{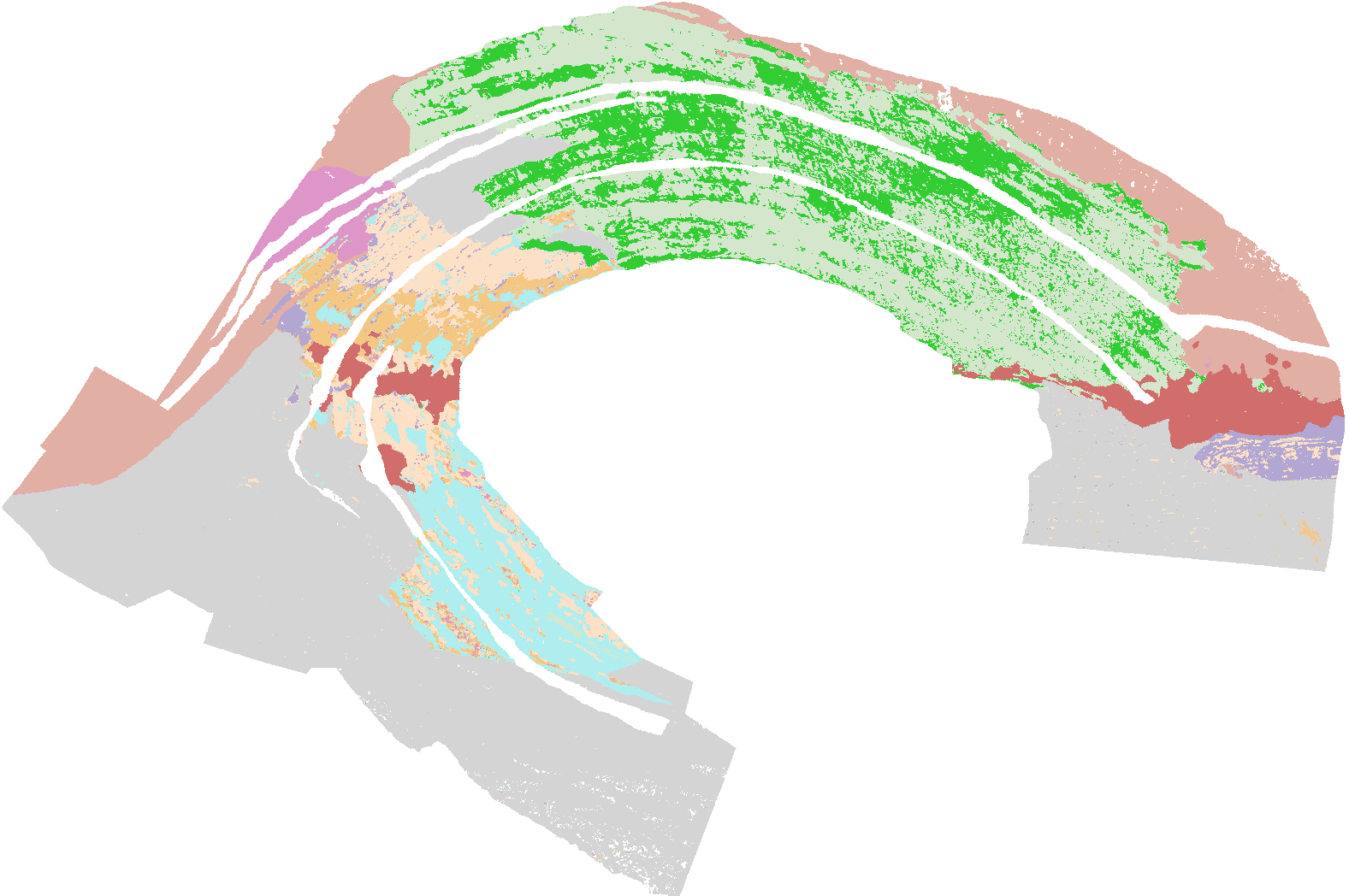} %
    \caption{LWIR ConvPoint clean}
\end{subfigure}%
\begin{subfigure}{0.24\textwidth}
    \centering
    \includegraphics[width=0.98\linewidth]{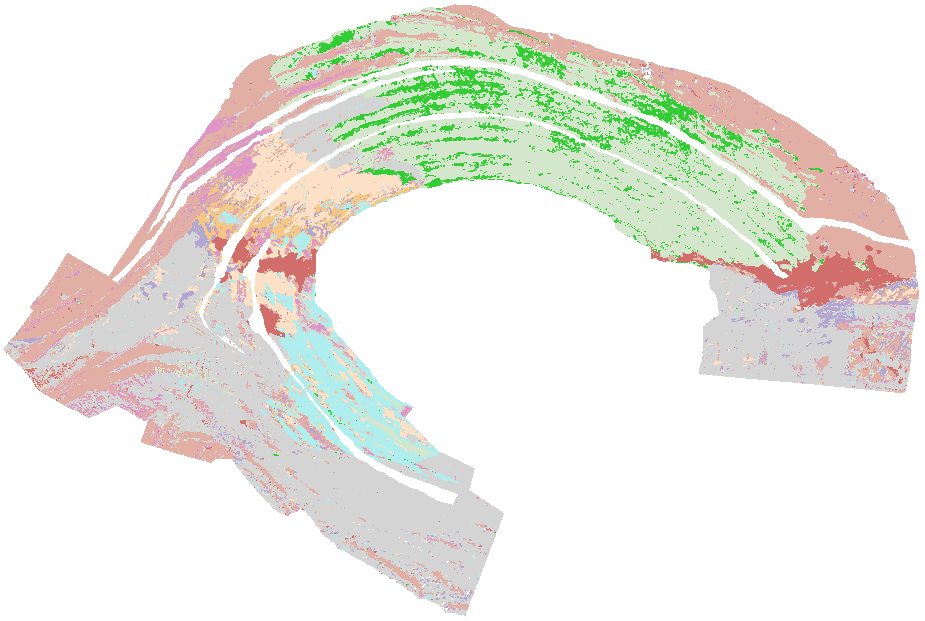} %
    \caption{LWIR ConvPoint noisy}
\end{subfigure}%
\begin{subfigure}{0.24\textwidth}
    \centering
    \includegraphics[width=0.98\linewidth]{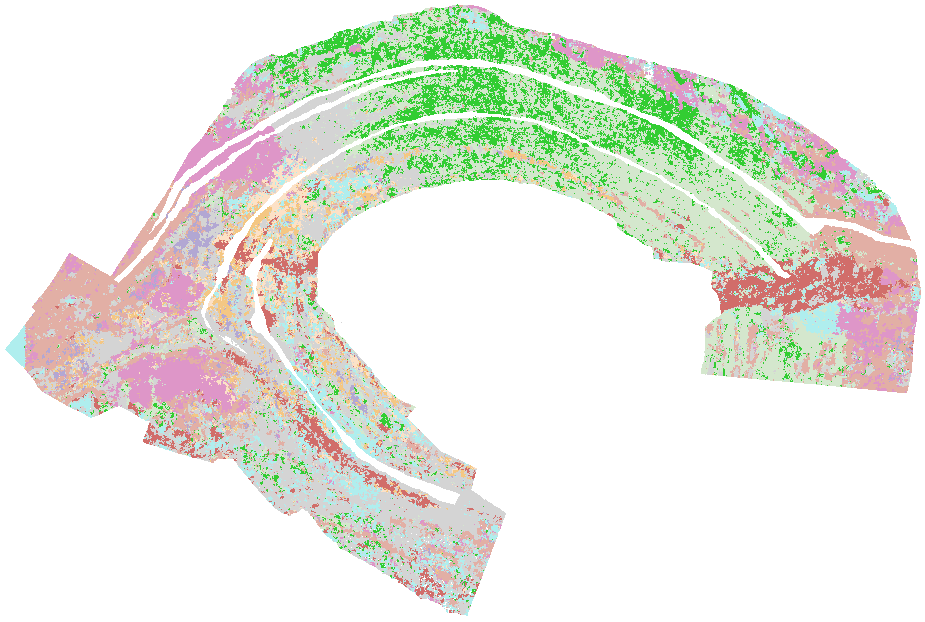} %
    \caption{LWIR DGCNN real}
\end{subfigure}%

\begin{subfigure}{0.24\textwidth}
    \centering
    \includegraphics[width=0.98\linewidth]{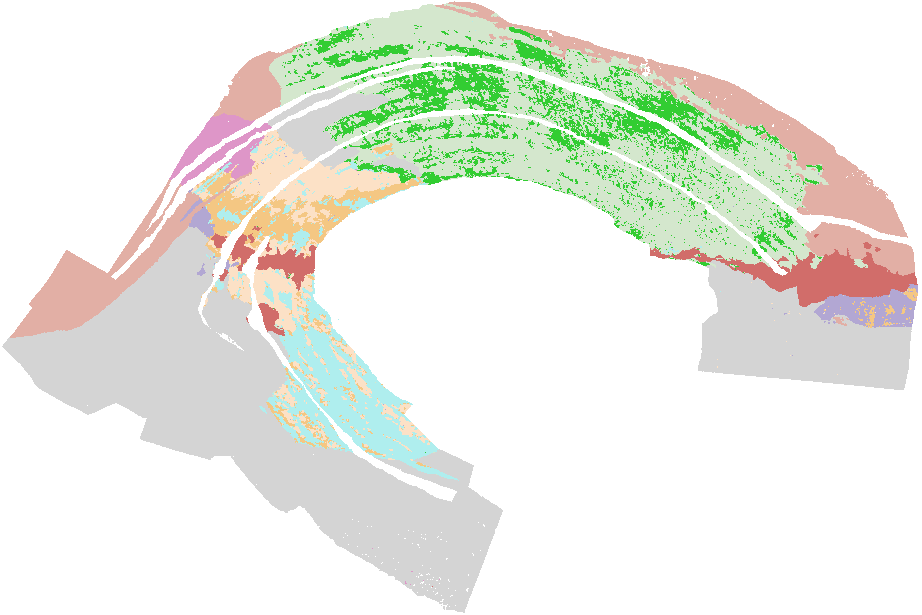} %
    \caption{LWIR DGCNN clean}
\end{subfigure}%
\begin{subfigure}{0.24\textwidth}
    \centering
    \includegraphics[width=0.98\linewidth]{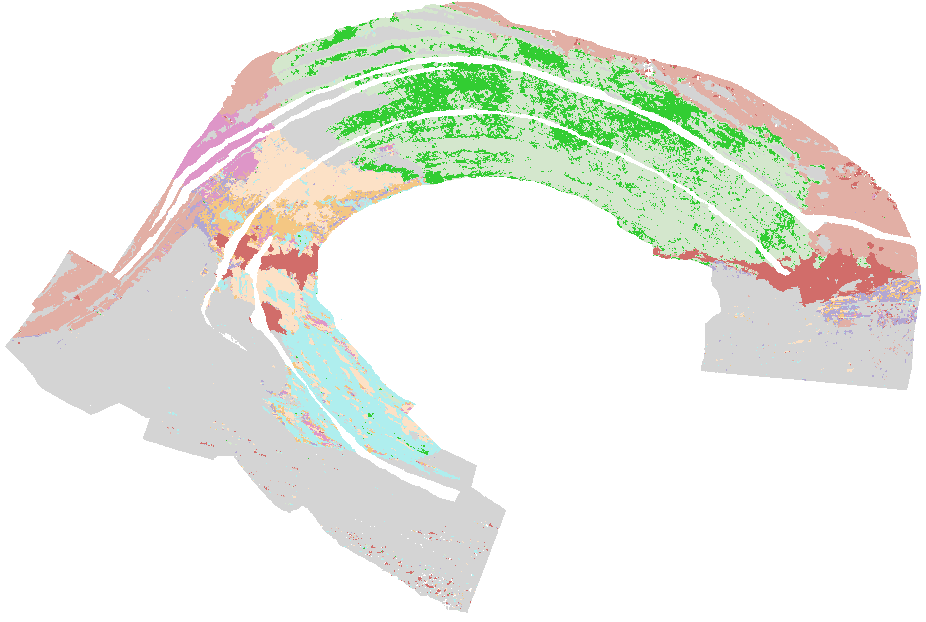} %
    \caption{LWIR DGCNN noisy}
\end{subfigure}%
\begin{subfigure}{0.24\textwidth}
    \centering
    \includegraphics[width=0.98\linewidth]{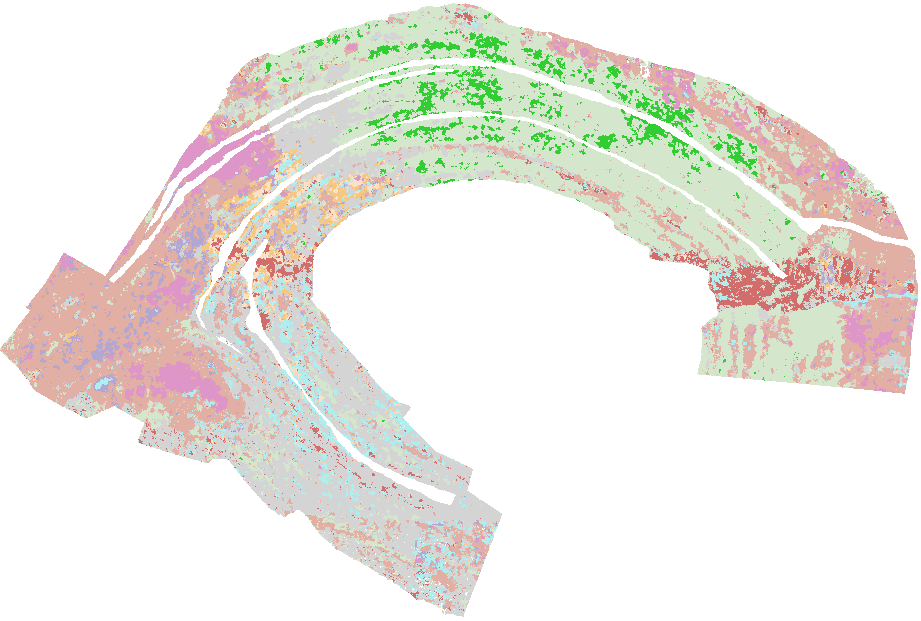} %
    \caption{LWIR PT real}
\end{subfigure}%
\begin{subfigure}{0.24\textwidth}
    \centering
    \includegraphics[width=0.98\linewidth]{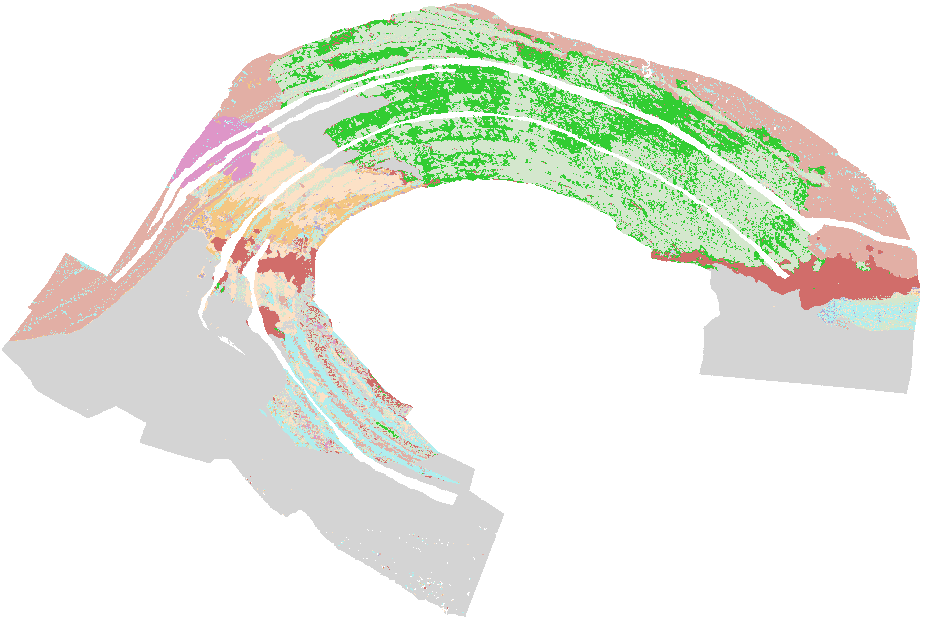} %
    \caption{LWIR PT clean}
\end{subfigure}%

\begin{subfigure}{0.24\textwidth}
    \centering
    \includegraphics[width=0.98\linewidth]{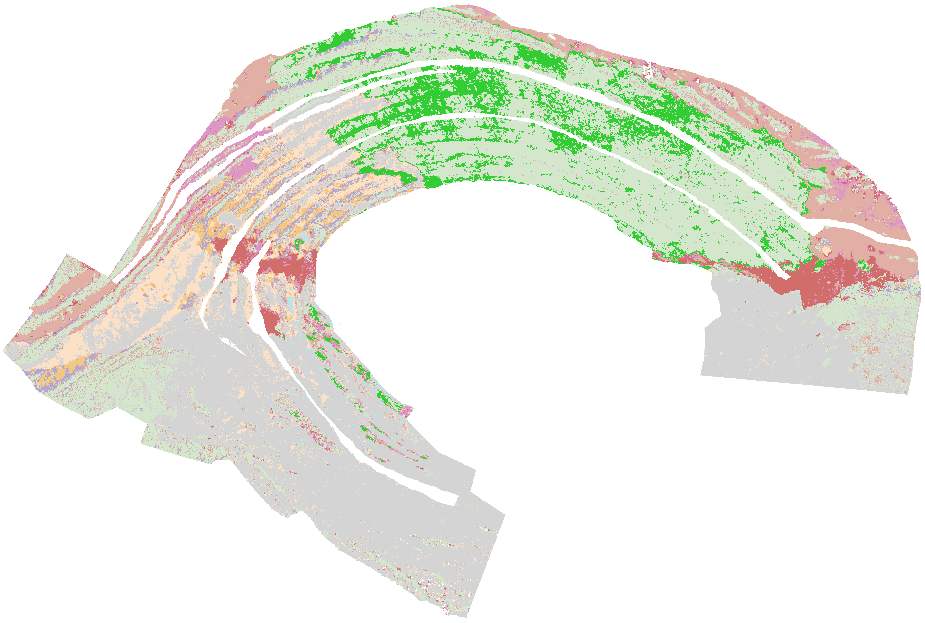} %
    \caption{LWIR PT noisy}
\end{subfigure}%
\begin{subfigure}{0.24\textwidth}
    \centering
    \includegraphics[width=0.98\linewidth]{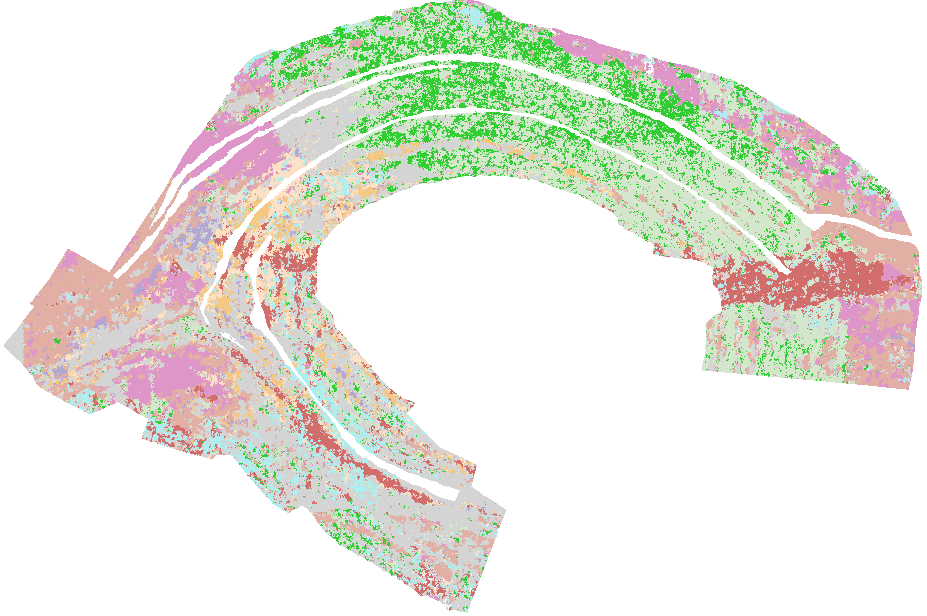} %
    \caption{LWIR PCT real}
\end{subfigure}%
\begin{subfigure}{0.24\textwidth}
    \centering
    \includegraphics[width=0.98\linewidth]{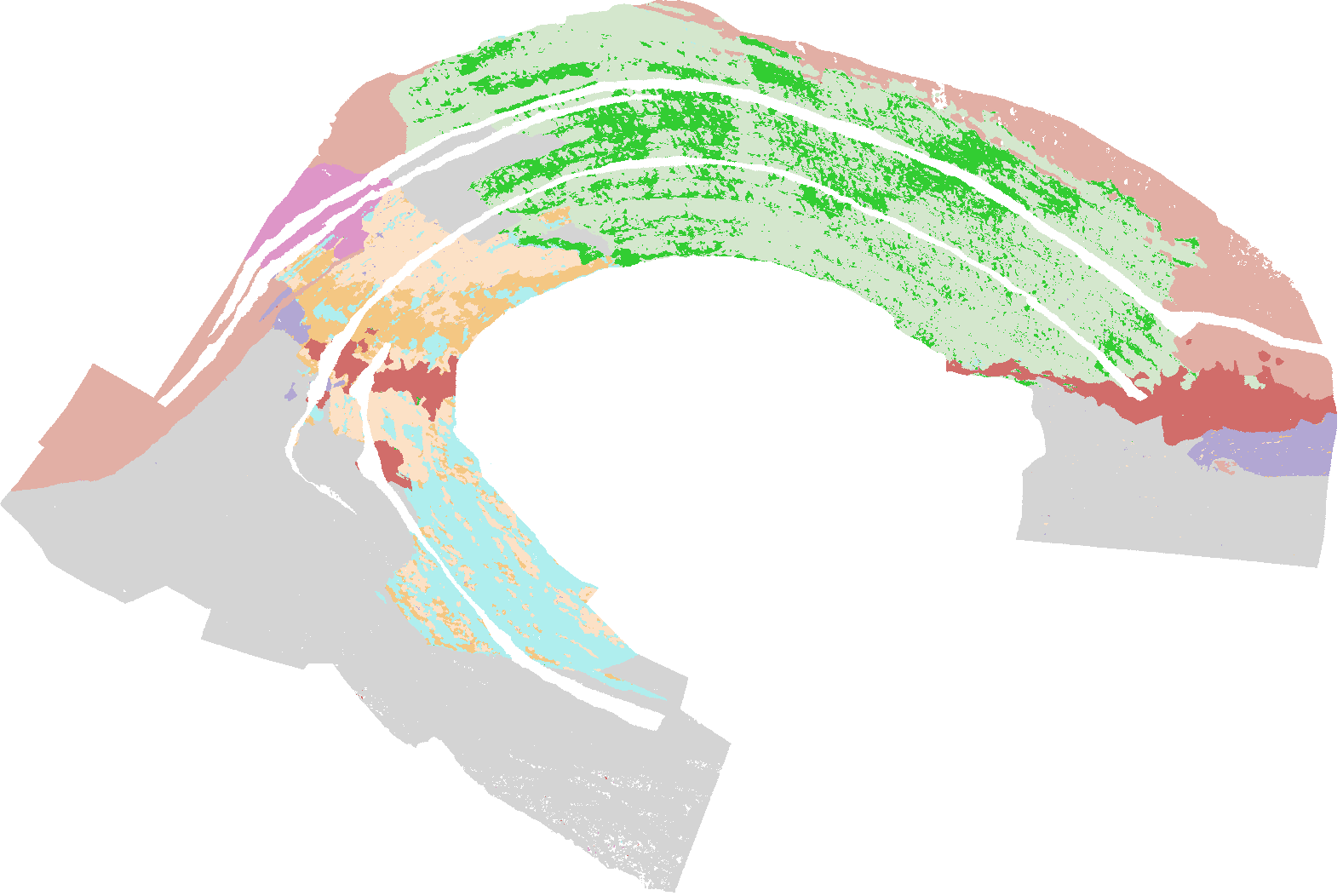} %
    \caption{LWIR PCT clean}
\end{subfigure}%
\begin{subfigure}{0.24\textwidth}
    \centering
    \includegraphics[width=0.98\linewidth]{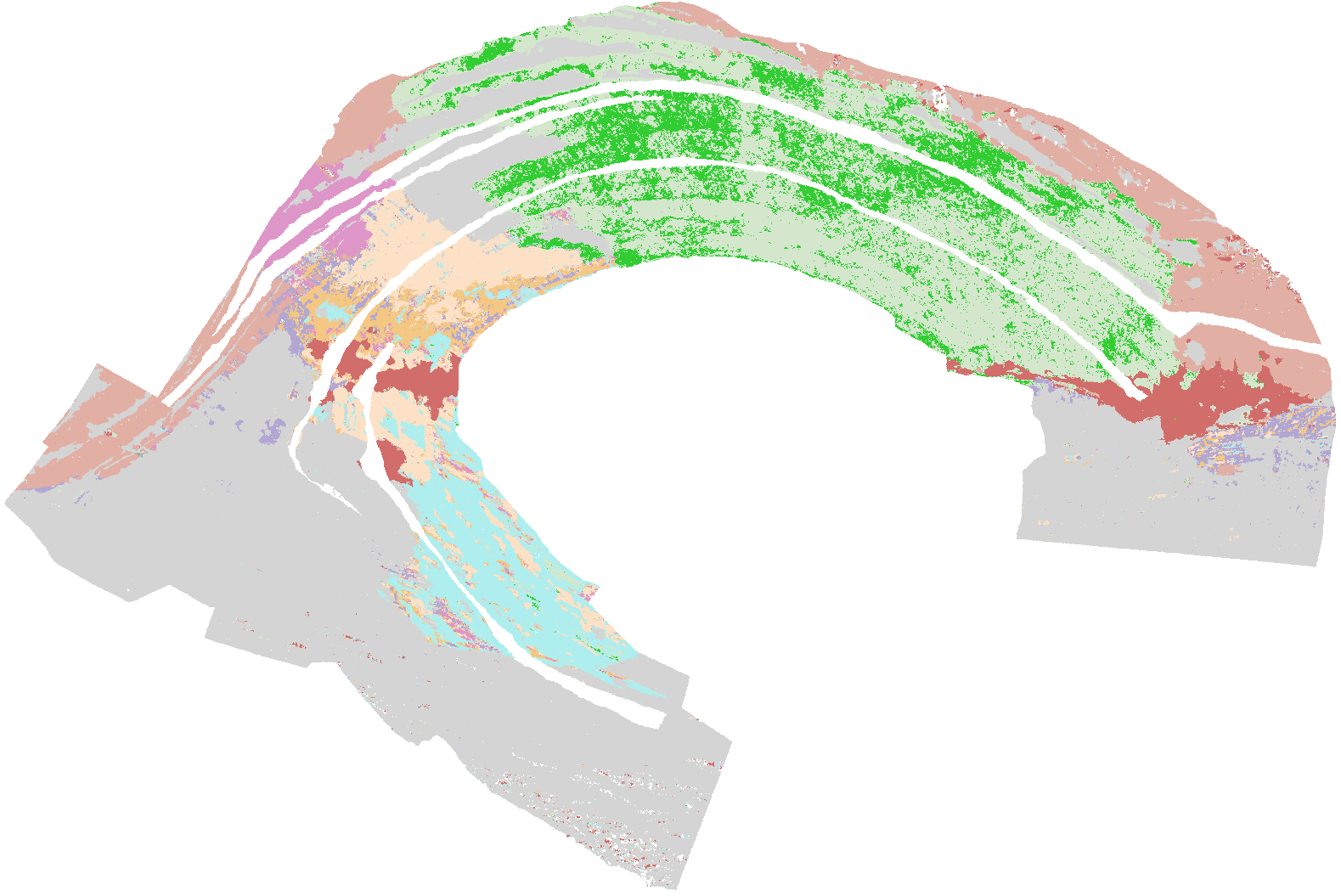} %
    \caption{LWIR PCT noisy}
\end{subfigure}%

\caption{Segmentation results of the baseline models on the LWIR test in various scenarios.}
\label{LWIR_results}
\end{figure*}

\begin{figure*}[!htpb]
\centering
\begin{subfigure}{0.25\textwidth}
    \centering
    \includegraphics[width=0.98\linewidth, trim = {0.5cm 0.5cm 0.5cm 0.5cm}, clip]{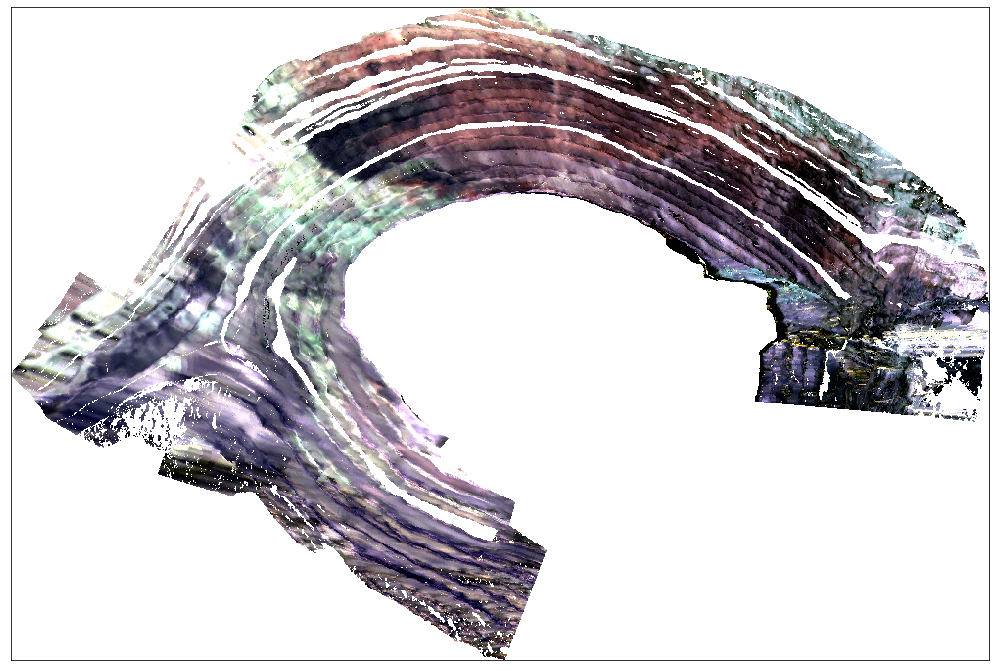}
    \caption{SWIR}
\end{subfigure}%
\hspace{0.5 cm}
\begin{subfigure}{0.25\textwidth}
    \centering
    \includegraphics[width=0.98\linewidth, trim = {0.5cm 0.5cm 0.5cm 0.5cm}, clip]{figures/gt-complete.png} %
    \caption{ground truth}
\end{subfigure}%

\begin{subfigure}{0.24\textwidth}
    \centering
    \includegraphics[width=0.98\linewidth, trim = {0.5cm 0.5cm 0.5cm 0.5cm}, clip]{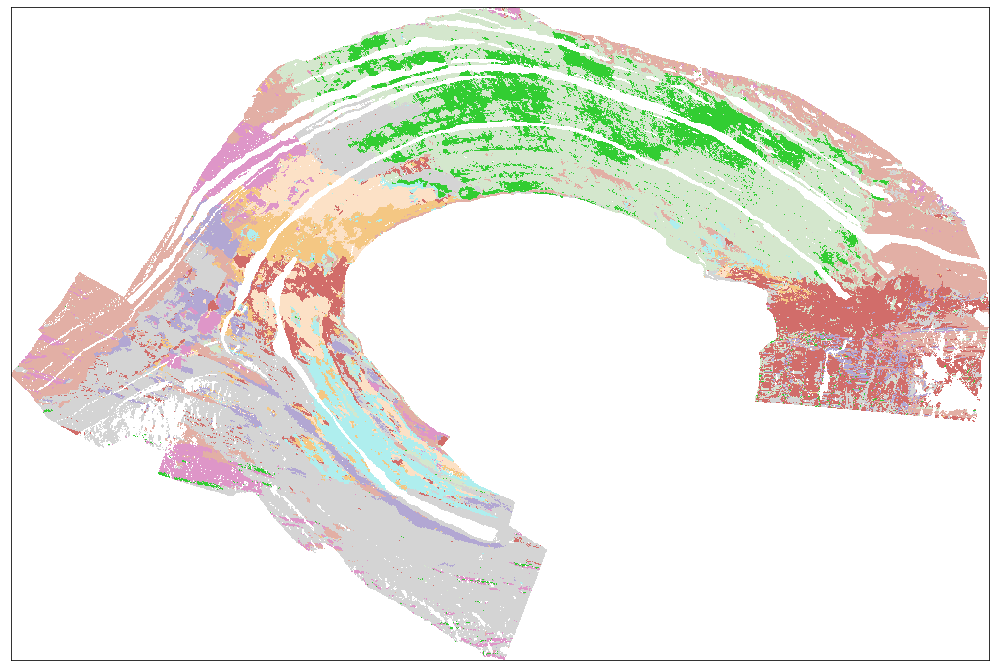} %
    \caption{SWIR MLP real}
\end{subfigure}%
\begin{subfigure}{0.24\textwidth}
    \centering
    \includegraphics[width=0.98\linewidth, trim = {0.5cm 0.5cm 0.5cm 0.5cm}, clip]{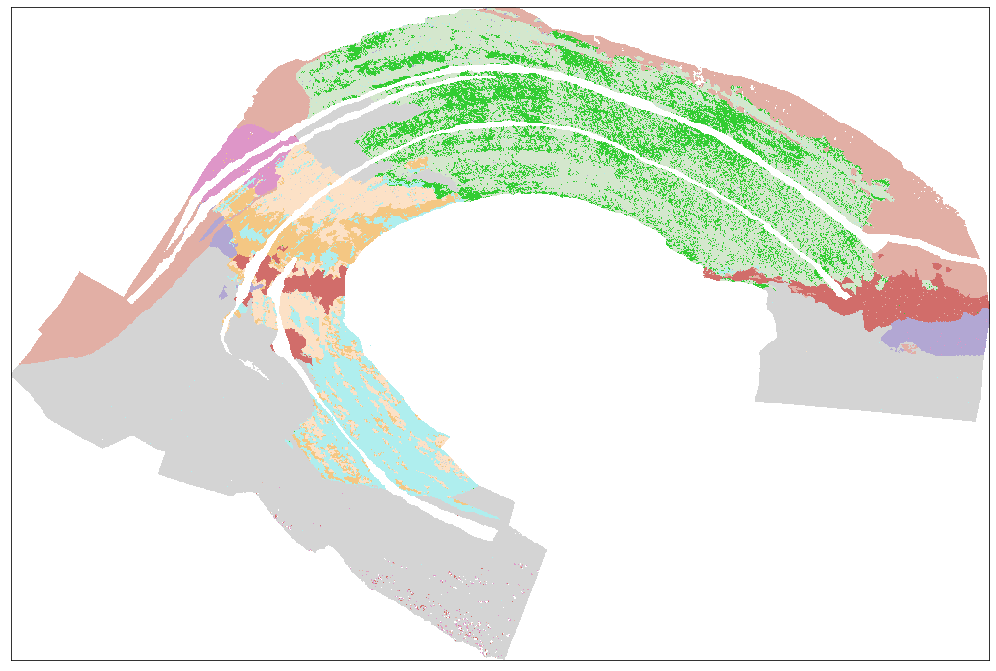}
    \caption{SWIR MLP clean}
\end{subfigure}%
\begin{subfigure}{0.24\textwidth}
    \centering
    \includegraphics[width=0.98\linewidth, trim = {0.5cm 0.5cm 0.5cm 0.5cm}, clip]{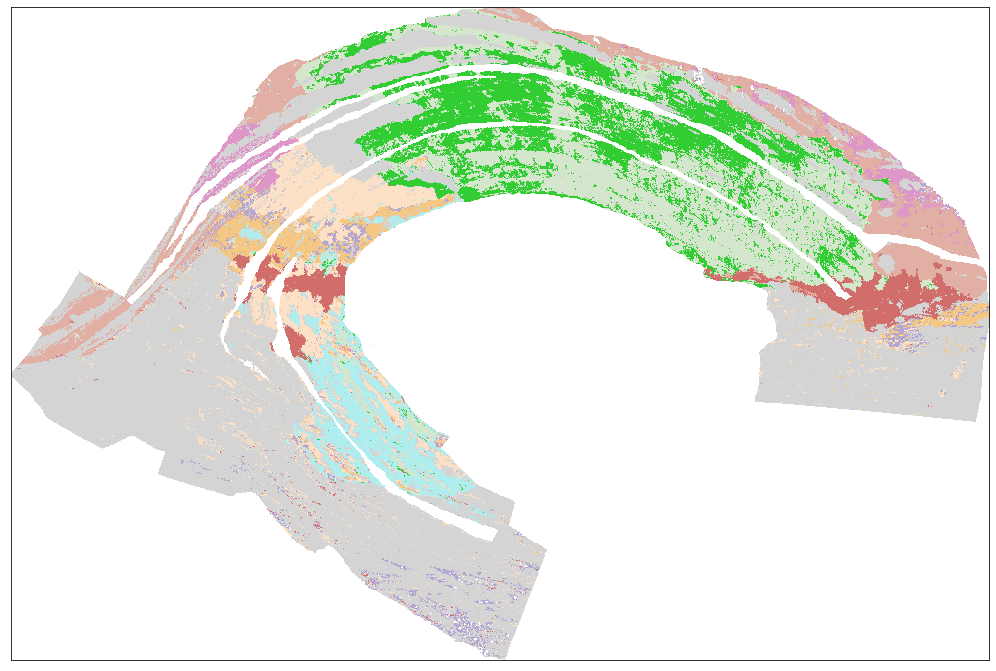}
    \caption{SWIR MLP noisy}
\end{subfigure}%
\begin{subfigure}{0.24\textwidth}
    \centering
    \includegraphics[width=0.98\linewidth]{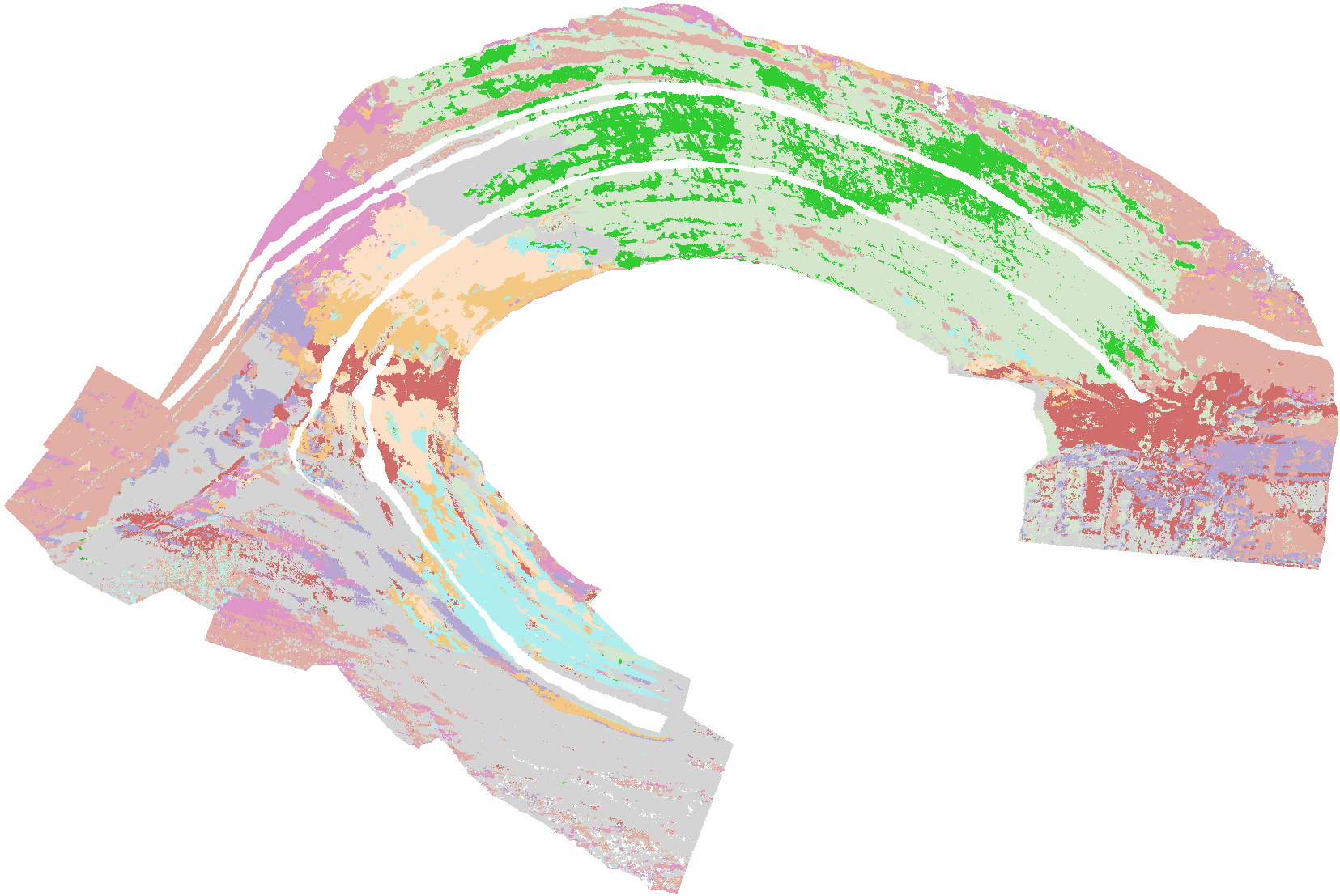} %
    \caption{SWIR PointNet real}
\end{subfigure}%

\begin{subfigure}{0.24\textwidth}
    \centering
    \includegraphics[width=0.98\linewidth]{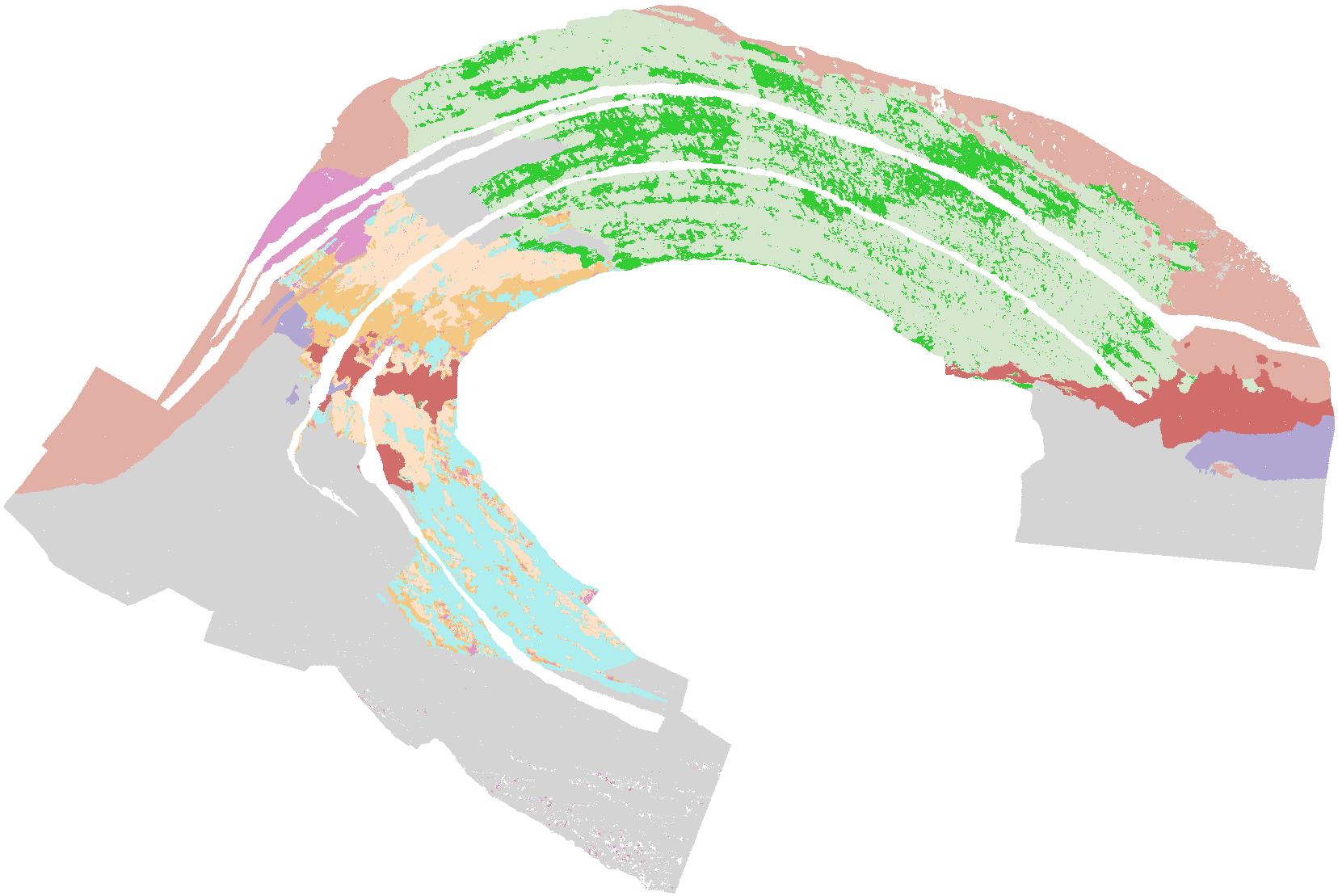}
    \caption{SWIR PointNet clean}
\end{subfigure}%
\begin{subfigure}{0.24\textwidth}
    \centering
    \includegraphics[width=0.98\linewidth]{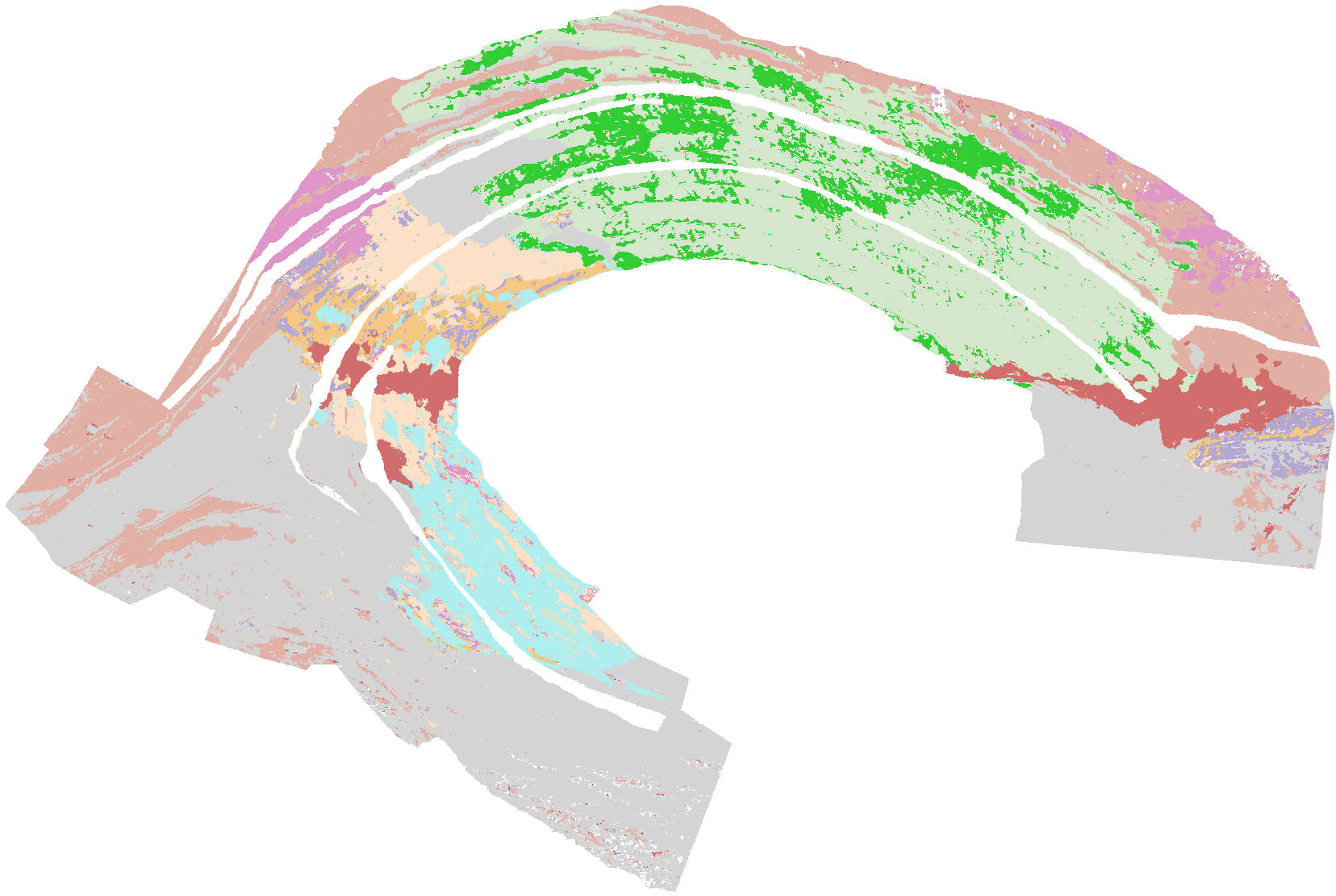}
    \caption{SWIR PointNet noisy}
\end{subfigure}%
\begin{subfigure}{0.24\textwidth}
    \centering
    \includegraphics[width=0.98\linewidth]{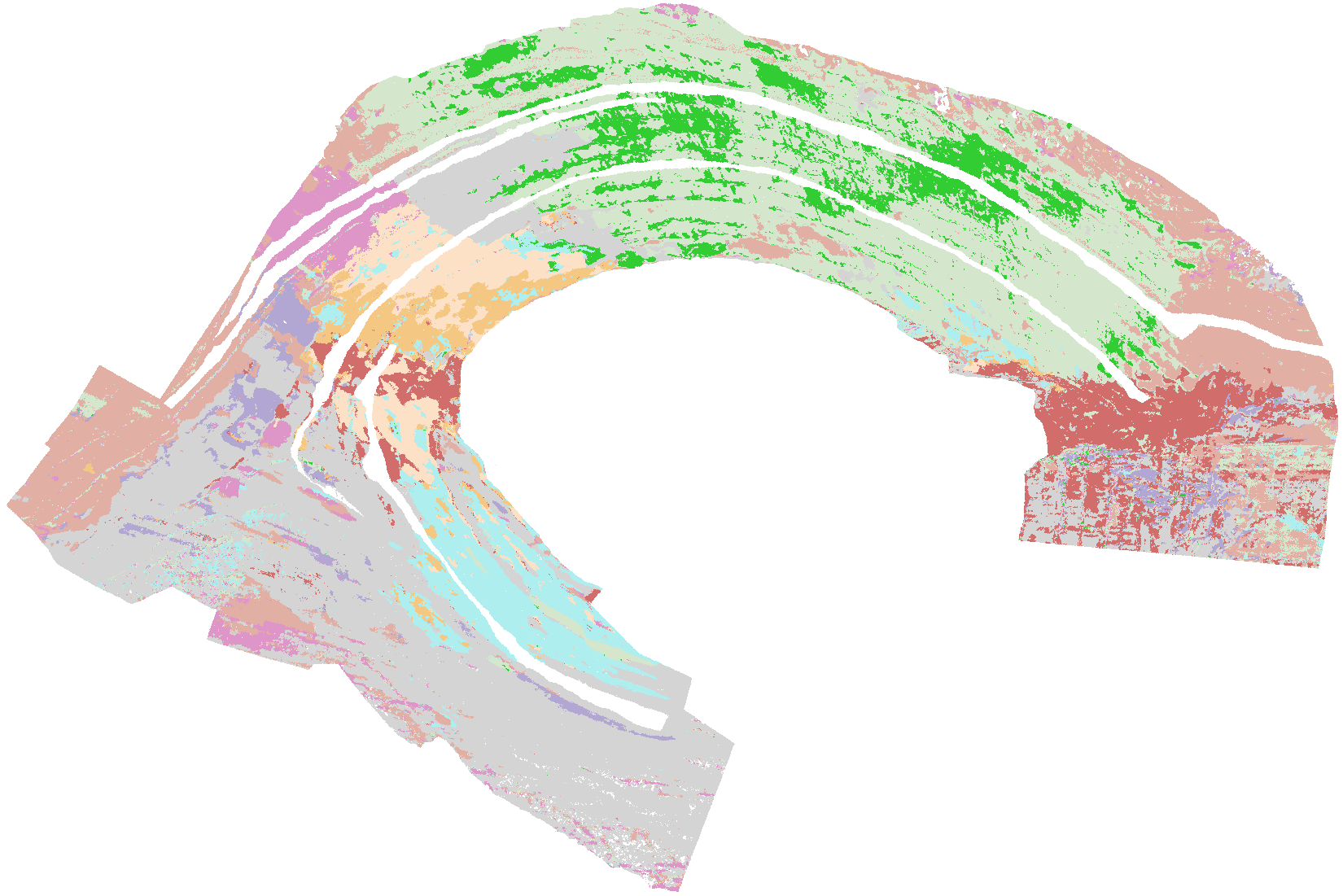} %
    \caption{SWIR PointNet2 real}
\end{subfigure}%
\begin{subfigure}{0.24\textwidth}
    \centering
    \includegraphics[width=0.98\linewidth]{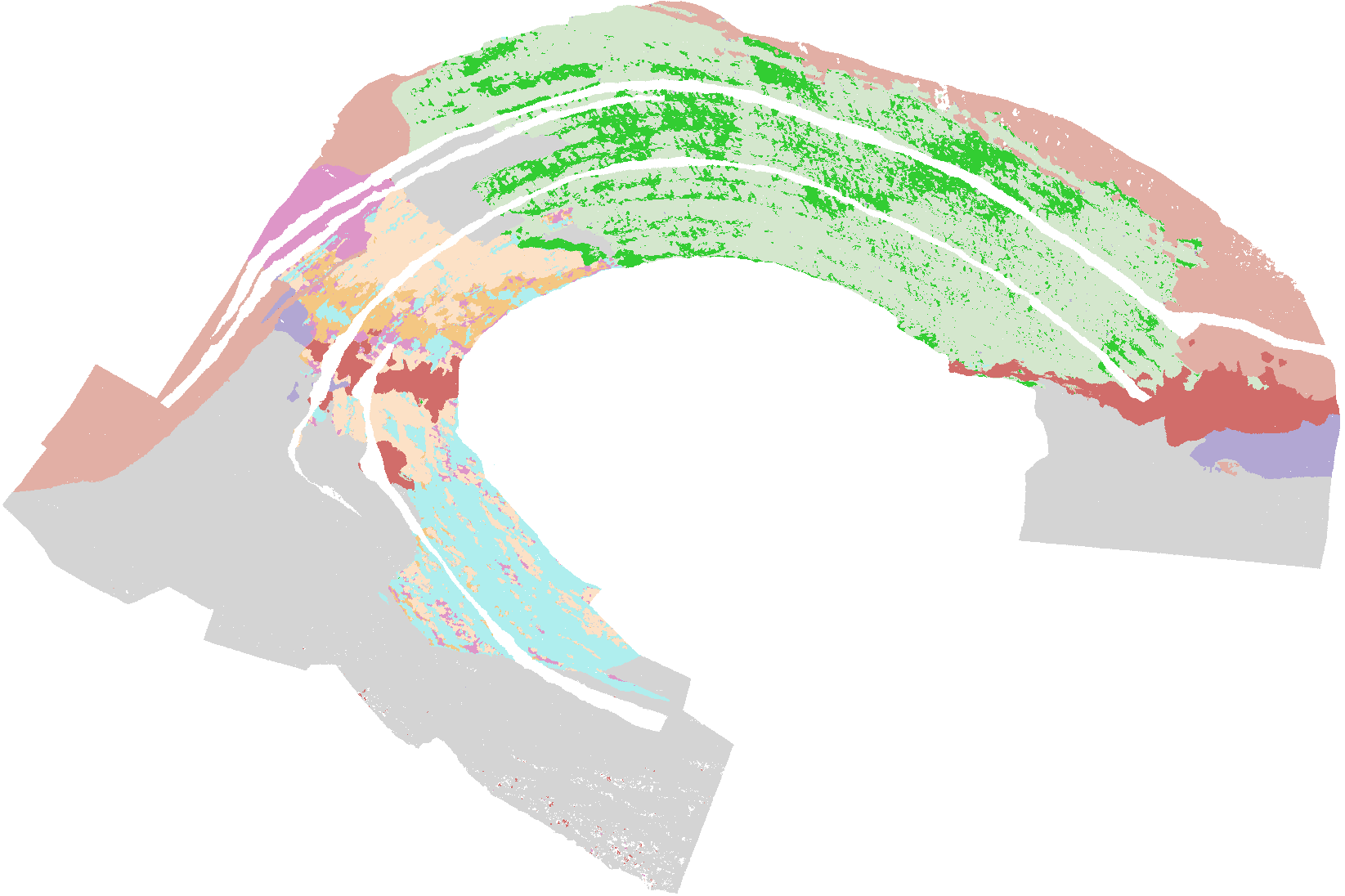}
    \caption{SWIR PointNet2 clean}
\end{subfigure}%

\begin{subfigure}{0.24\textwidth}
    \centering
    \includegraphics[width=0.98\linewidth]{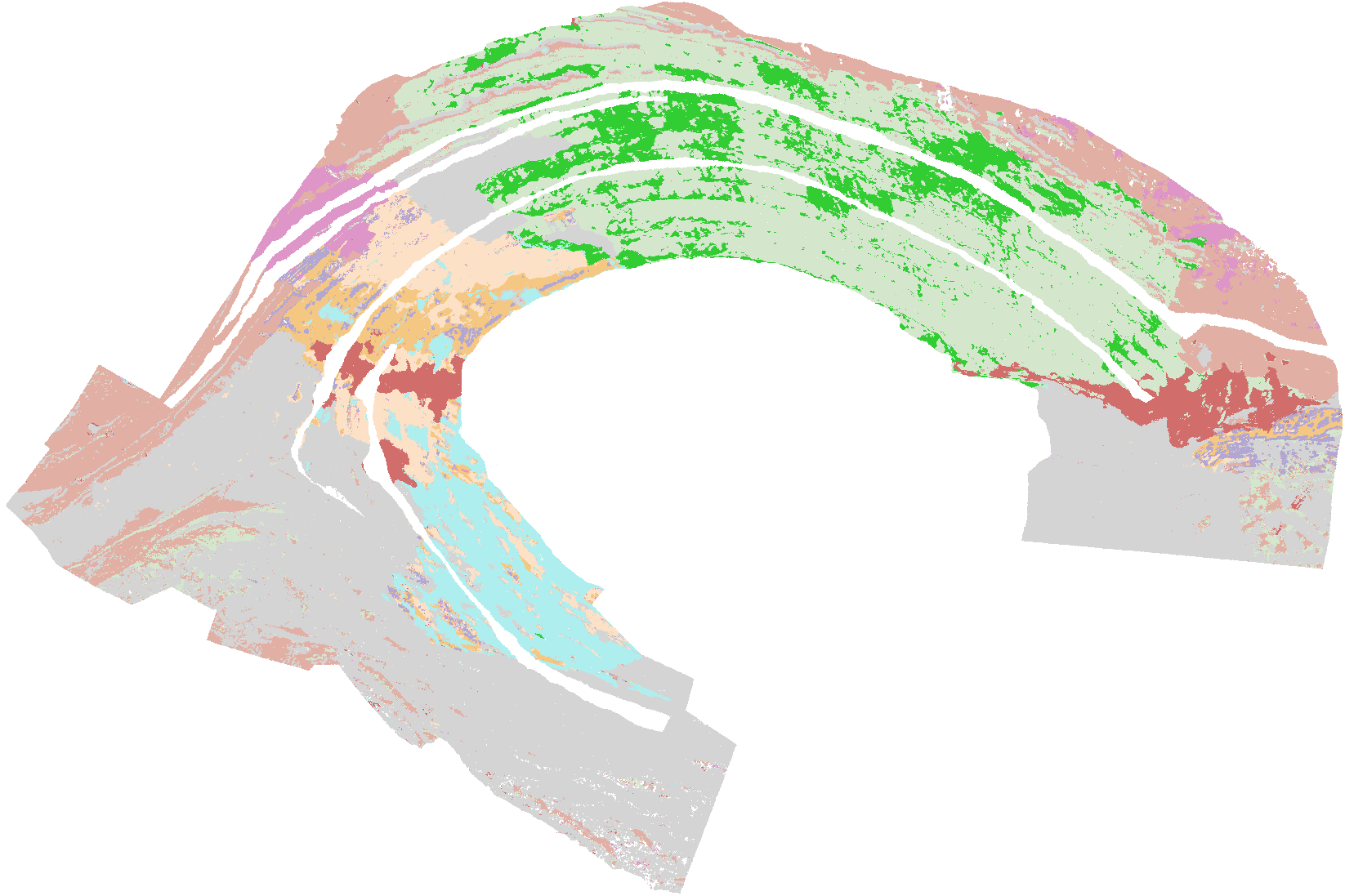}
    \caption{SWIR PointNet2 noisy}
\end{subfigure}%
\begin{subfigure}{0.24\textwidth}
    \centering
    \includegraphics[width=0.98\linewidth]{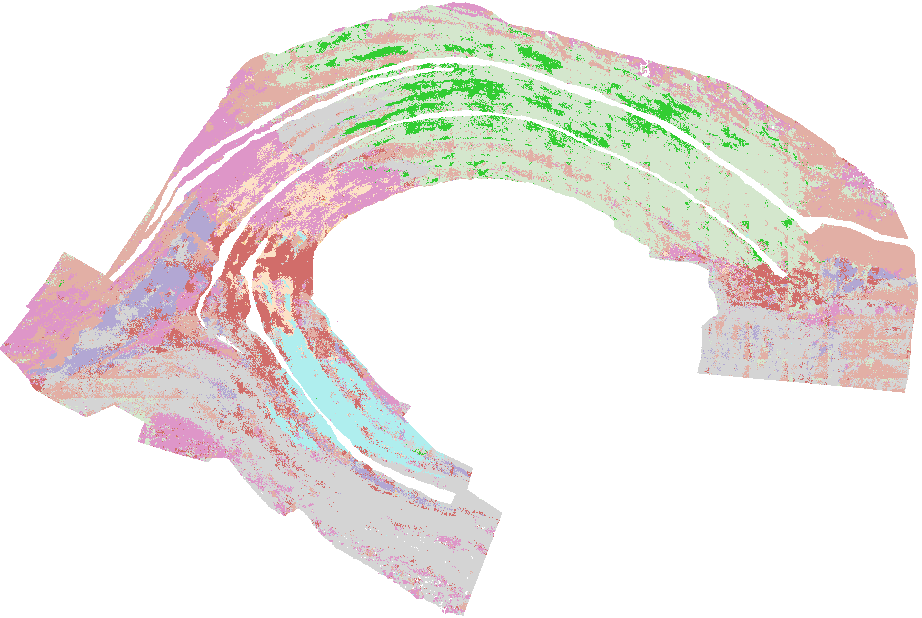} %
    \caption{SWIR PointCNN real}
\end{subfigure}%
\begin{subfigure}{0.24\textwidth}
    \centering
    \includegraphics[width=0.98\linewidth]{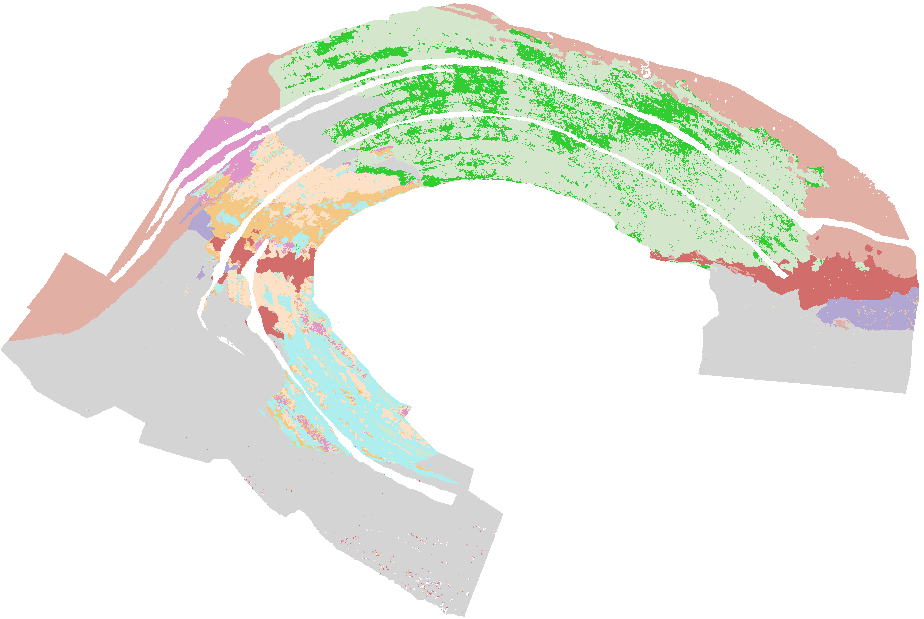} %
    \caption{SWIR PointCNN clean}
\end{subfigure}%
\begin{subfigure}{0.24\textwidth}
    \centering
    \includegraphics[width=0.98\linewidth]{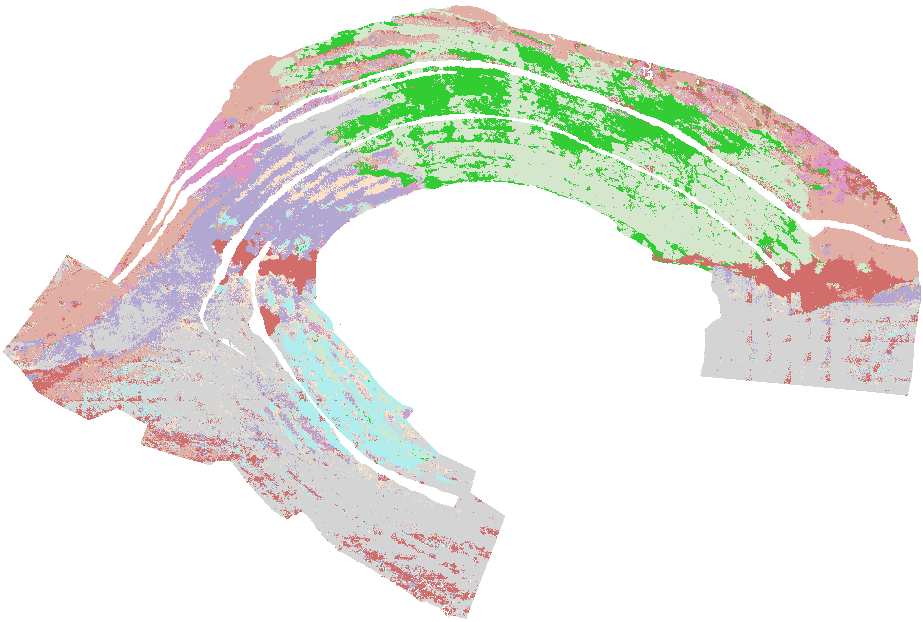} %
    \caption{SWIR PointCNN noisy}
\end{subfigure}%

\begin{subfigure}{0.24\textwidth}
    \centering
    \includegraphics[width=0.98\linewidth]{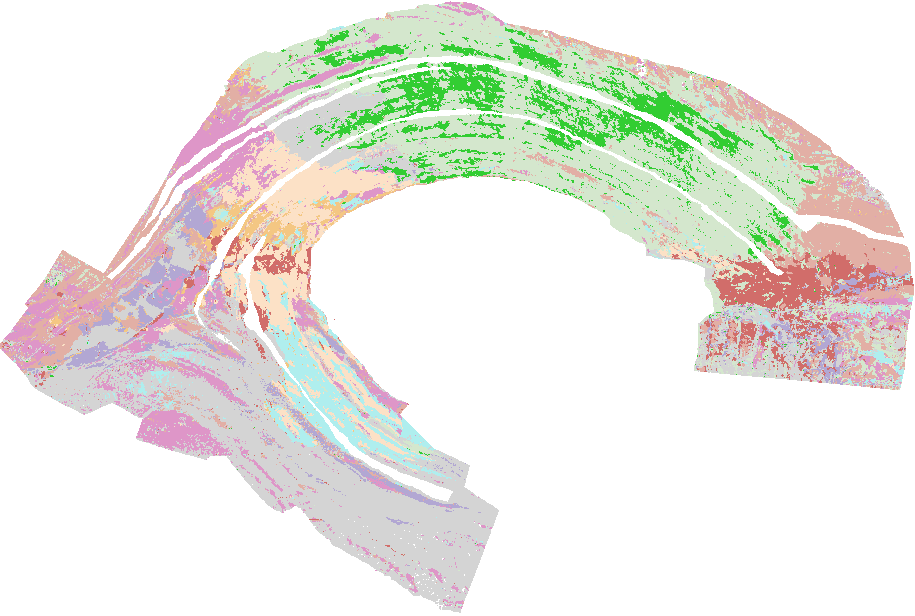} %
    \caption{SWIR ConvPoint real}
\end{subfigure}%
\begin{subfigure}{0.24\textwidth}
    \centering
    \includegraphics[width=0.98\linewidth]{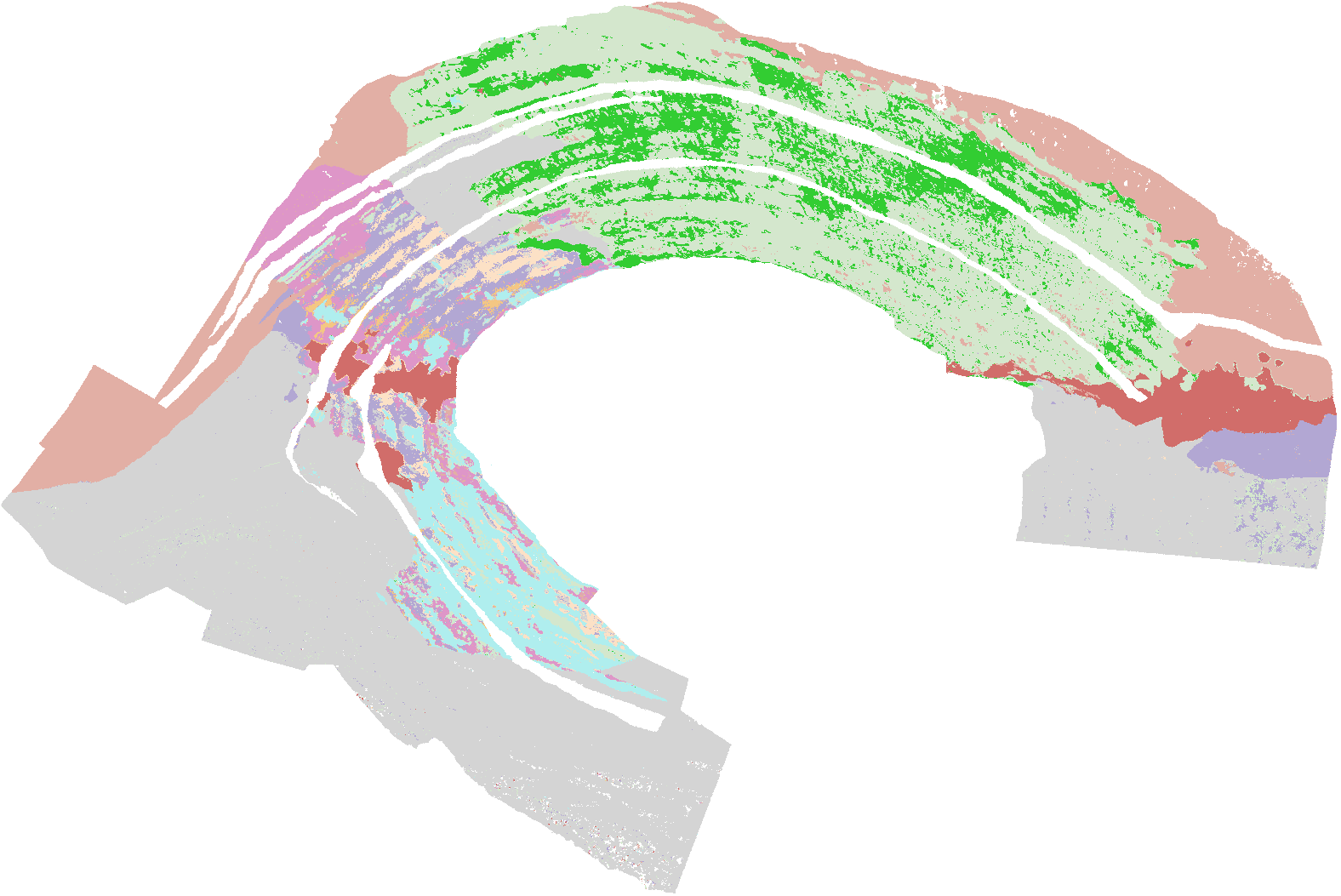} %
    \caption{SWIR ConvPoint clean}
\end{subfigure}%
\begin{subfigure}{0.24\textwidth}
    \centering
    \includegraphics[width=0.98\linewidth]{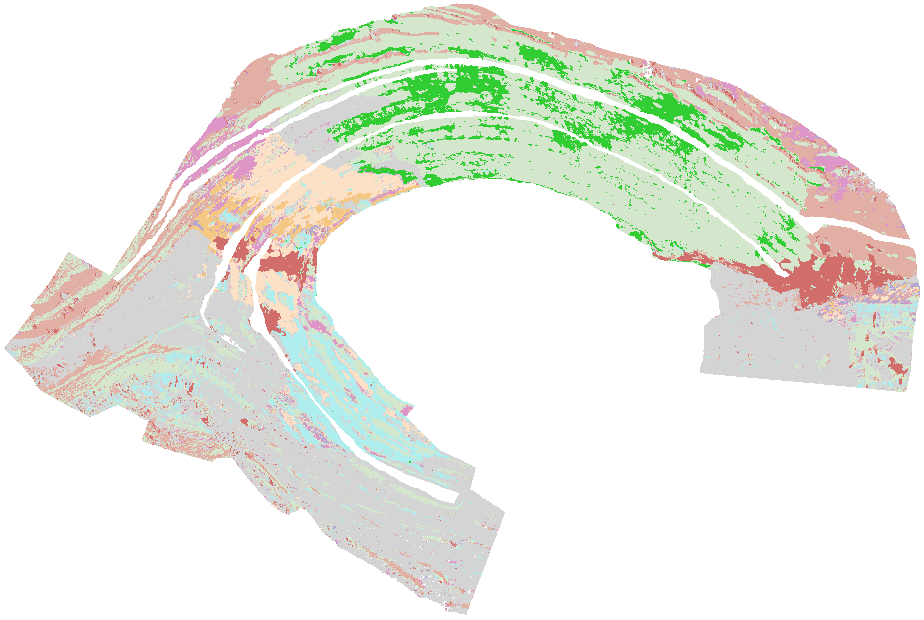} %
    \caption{SWIR ConvPoint noisy}
\end{subfigure}%
\begin{subfigure}{0.24\textwidth}
    \centering
    \includegraphics[width=0.98\linewidth]{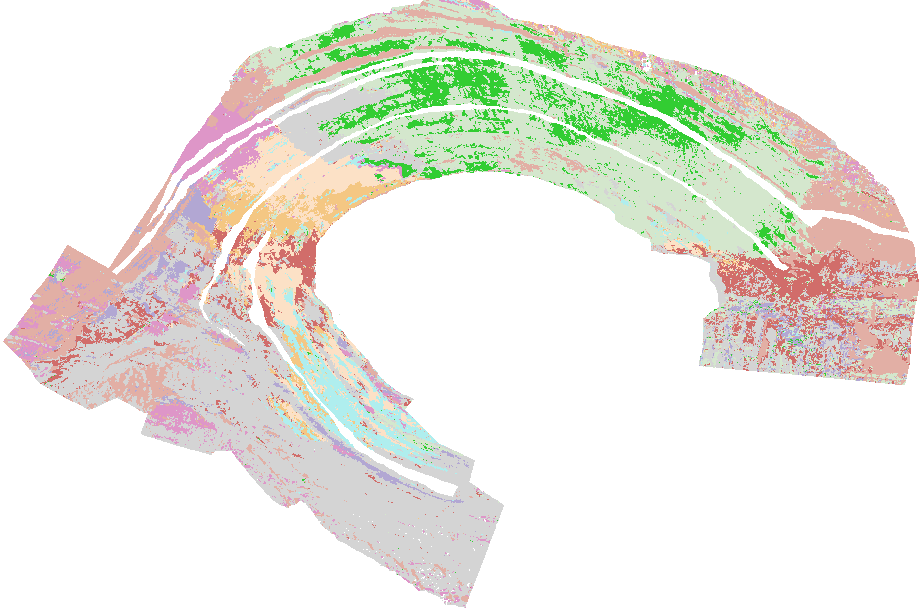} %
    \caption{SWIR DGCNN real}
\end{subfigure}%

\begin{subfigure}{0.24\textwidth}
    \centering
    \includegraphics[width=0.98\linewidth]{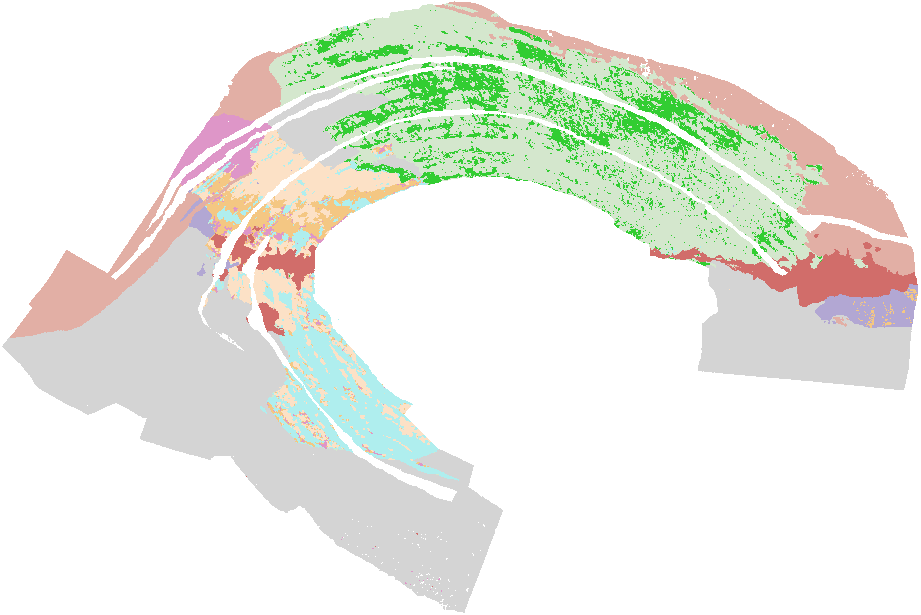} %
    \caption{SWIR DGCNN clean}
\end{subfigure}%
\begin{subfigure}{0.24\textwidth}
    \centering
    \includegraphics[width=0.98\linewidth]{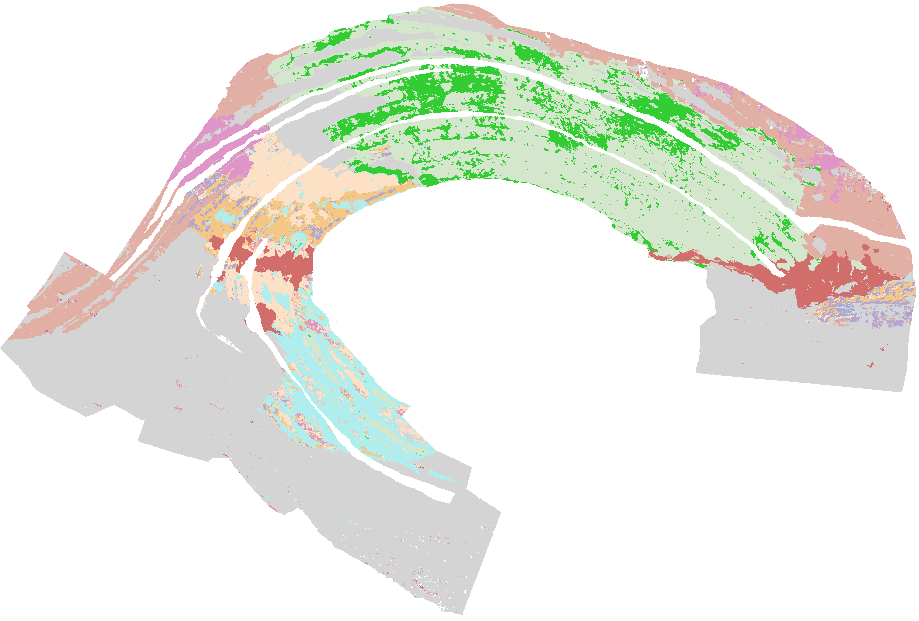} %
    \caption{SWIR DGCNN noisy}
\end{subfigure}%
\begin{subfigure}{0.24\textwidth}
    \centering
    \includegraphics[width=0.98\linewidth]{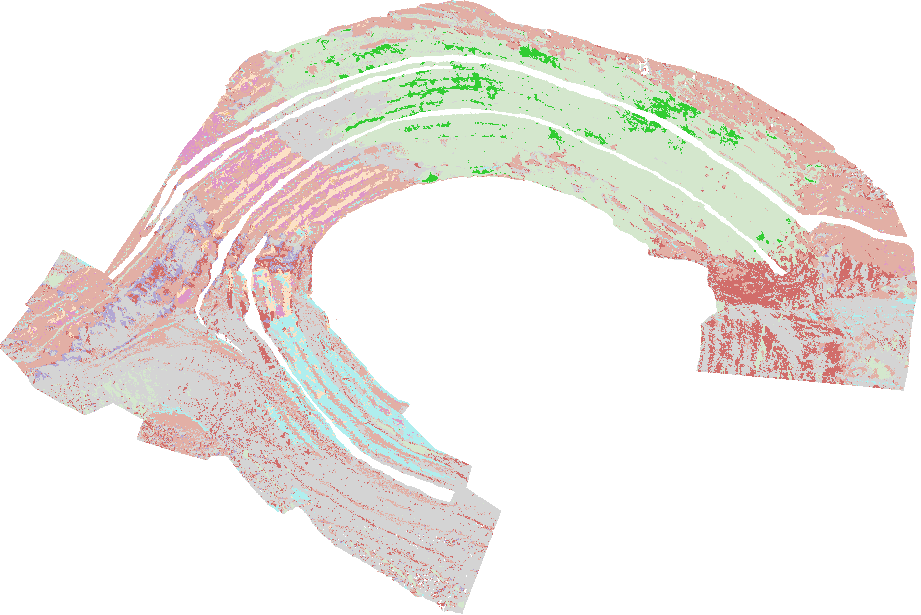} %
    \caption{SWIR PT real}
\end{subfigure}%
\begin{subfigure}{0.24\textwidth}
    \centering
    \includegraphics[width=0.98\linewidth]{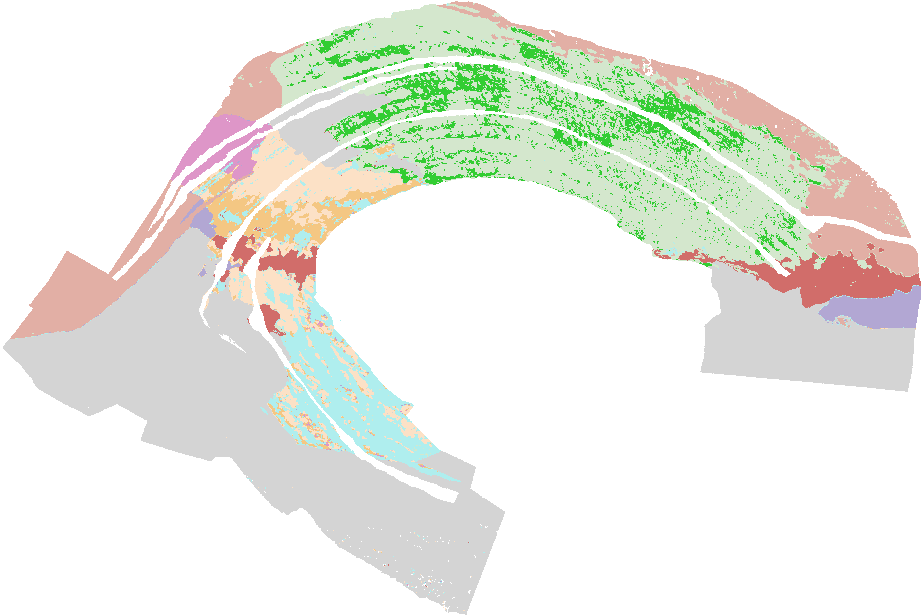} %
    \caption{SWIR PT clean}
\end{subfigure}%

\begin{subfigure}{0.24\textwidth}
    \centering
    \includegraphics[width=0.98\linewidth]{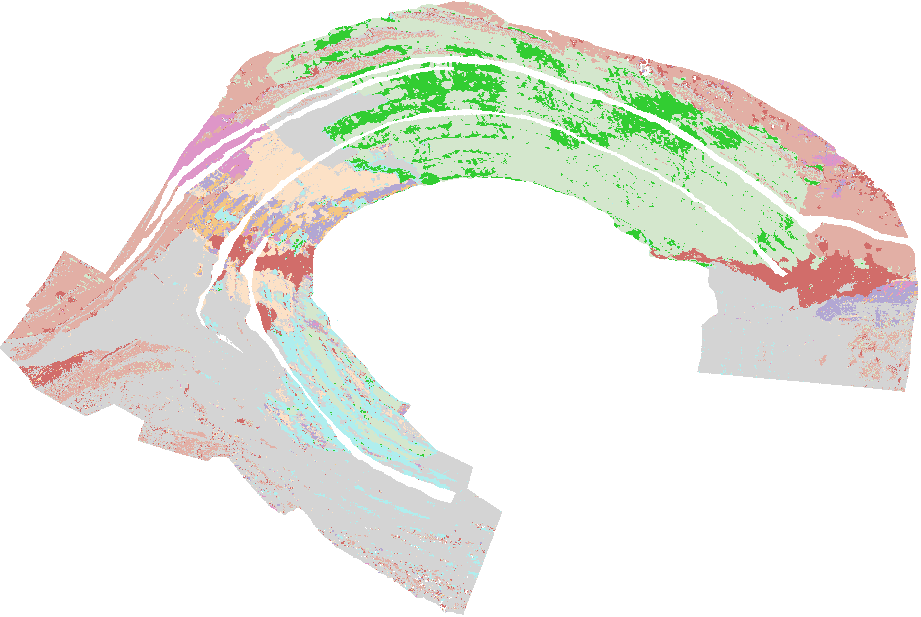} %
    \caption{SWIR PT noisy}
\end{subfigure}%
\begin{subfigure}{0.24\textwidth}
    \centering
    \includegraphics[width=0.98\linewidth]{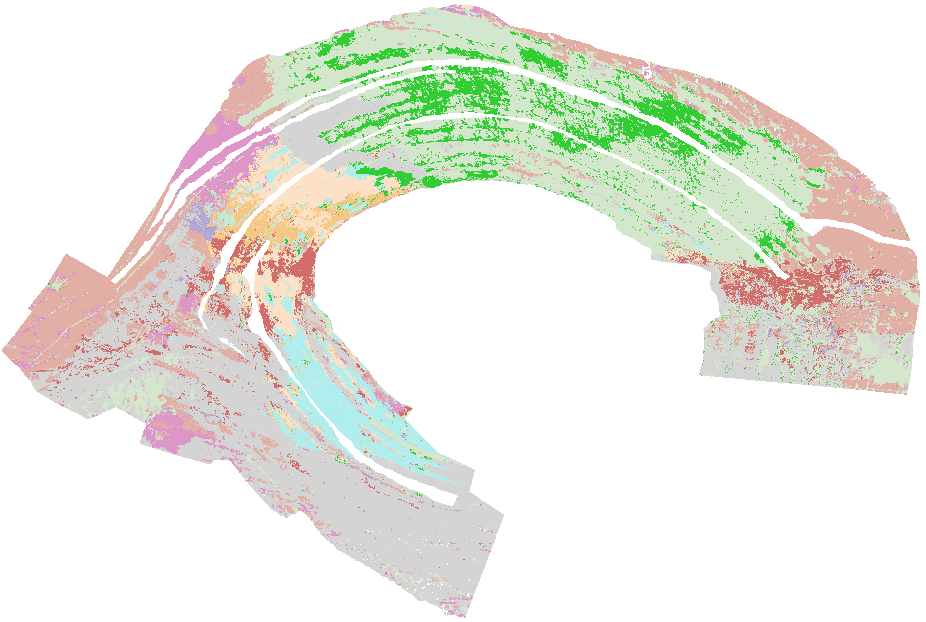} %
    \caption{SWIR PCT real}
\end{subfigure}%
\begin{subfigure}{0.24\textwidth}
    \centering
    \includegraphics[width=0.98\linewidth]{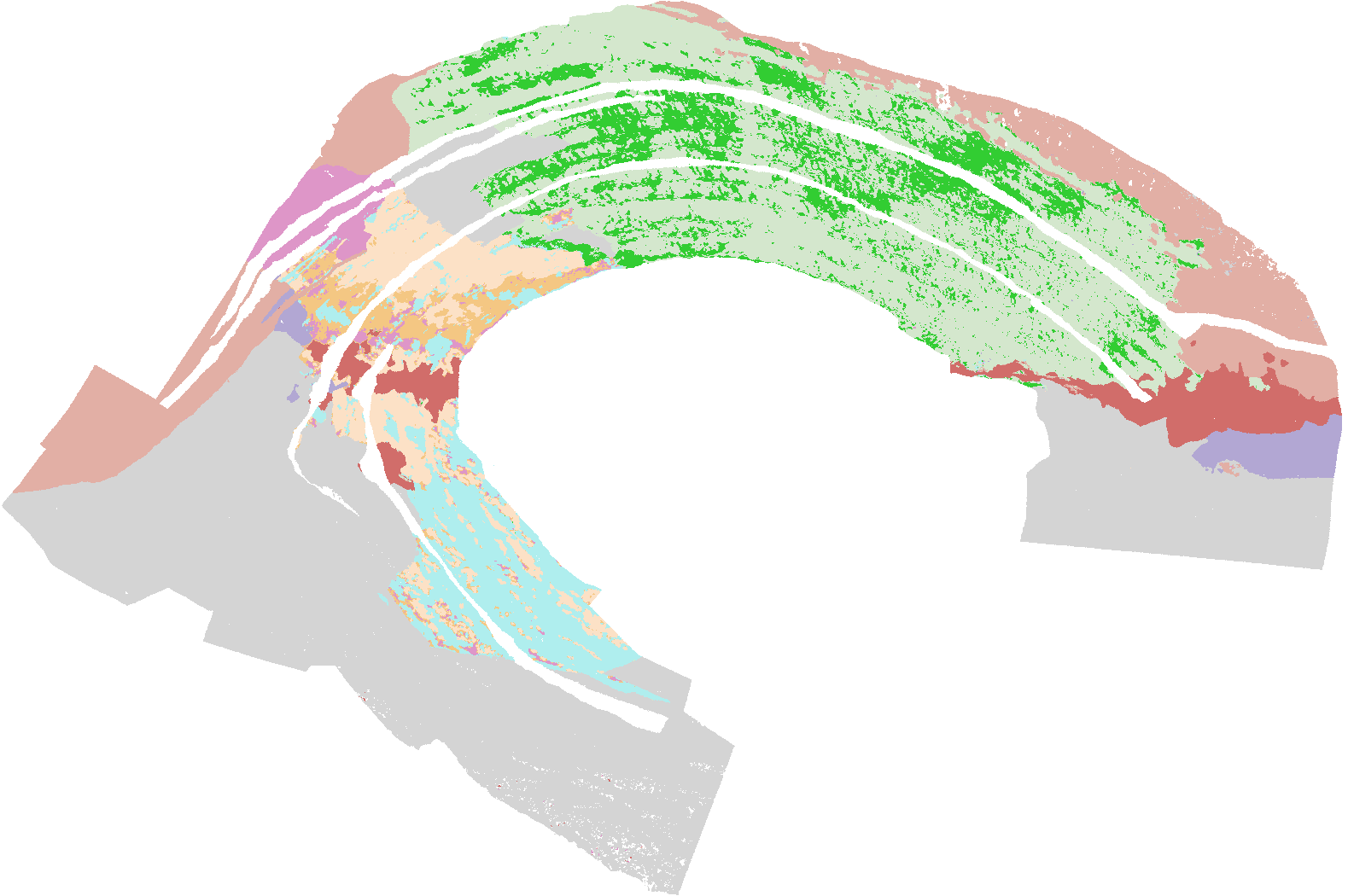} %
    \caption{SWIR PCT clean}
\end{subfigure}%
\begin{subfigure}{0.24\textwidth}
    \centering
    \includegraphics[width=0.98\linewidth]{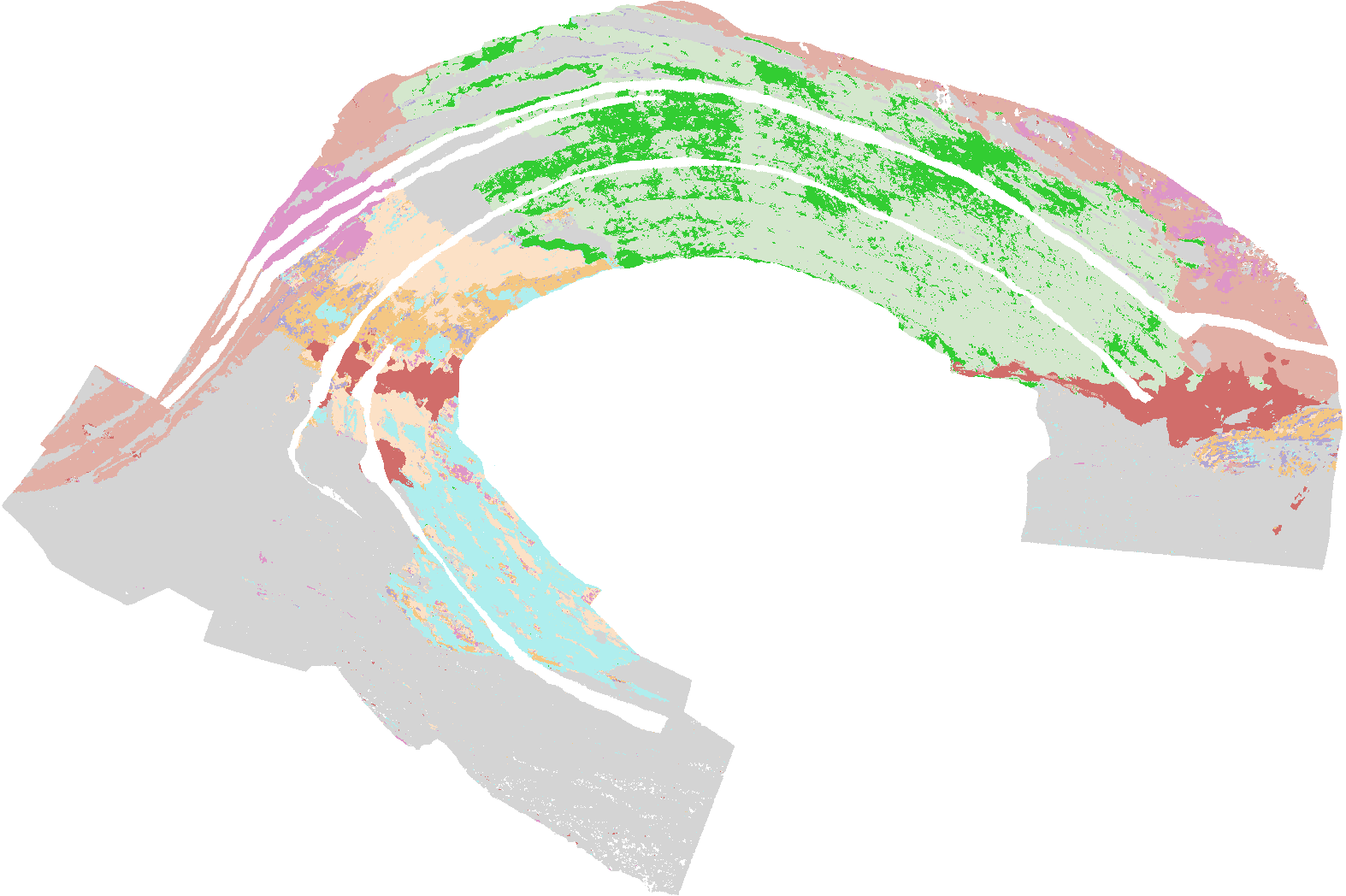} %
    \caption{SWIR PCT noisy}
\end{subfigure}%

\caption{Segmentation results of the baseline models on the SWIR test in various scenarios.}
\label{SWIR_results}
\end{figure*}

\begin{figure*}[!htpb]
\centering
\begin{subfigure}{0.25\textwidth}
    \centering
    \includegraphics[width=0.98\linewidth, trim = {0.5cm 0.5cm 0.5cm 0.5cm}, clip]{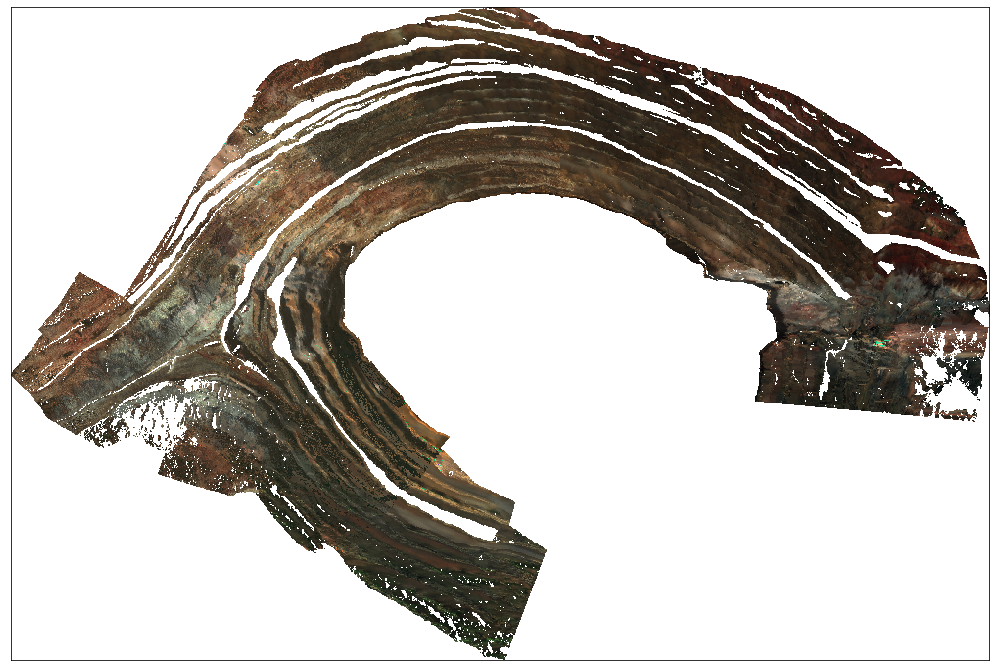}
    \caption{VNIR}
\end{subfigure}%
\hspace{0.5 cm}
\begin{subfigure}{0.25\textwidth}
    \centering
    \includegraphics[width=0.98\linewidth, trim = {0.5cm 0.5cm 0.5cm 0.5cm}, clip]{figures/gt-complete.png} %
    \caption{ground truth}
\end{subfigure}%

\begin{subfigure}{0.24\textwidth}
    \centering
    \includegraphics[width=0.98\linewidth, trim = {0.5cm 0.5cm 0.5cm 0.5cm}, clip]{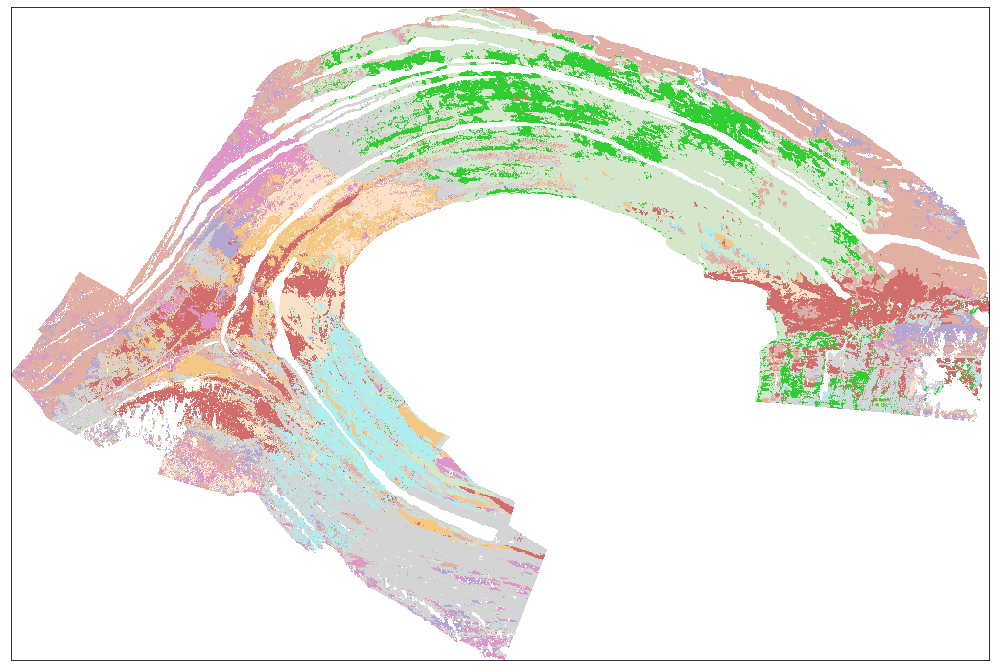} %
    \caption{VNIR MLP real}
\end{subfigure}%
\begin{subfigure}{0.24\textwidth}
    \centering
    \includegraphics[width=0.98\linewidth, trim = {0.5cm 0.5cm 0.5cm 0.5cm}, clip]{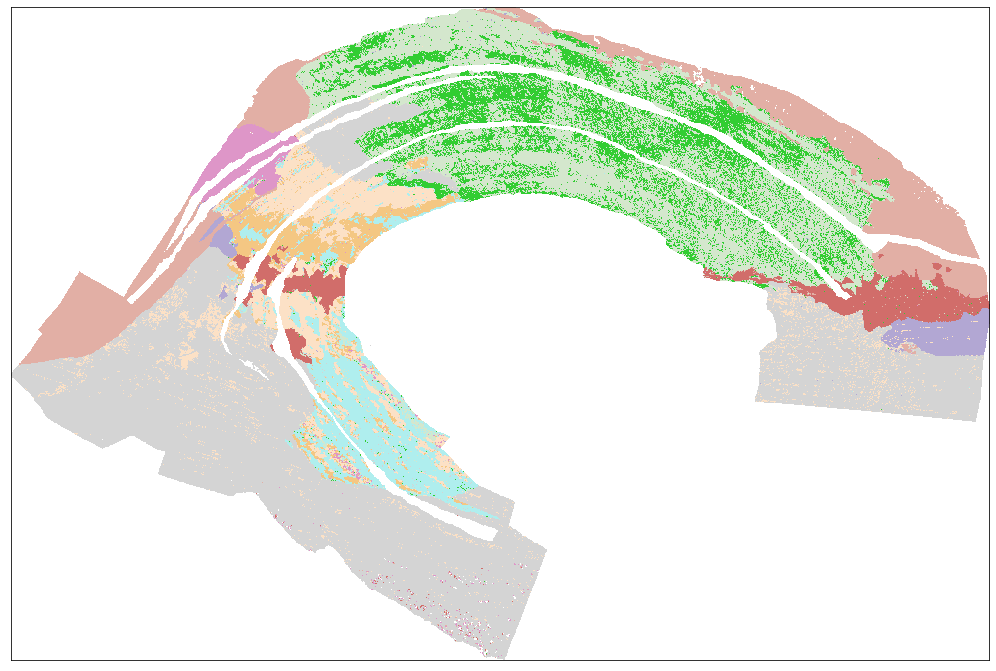}
    \caption{VNIR MLP clean}
\end{subfigure}%
\begin{subfigure}{0.24\textwidth}
    \centering
    \includegraphics[width=0.98\linewidth, trim = {0.5cm 0.5cm 0.5cm 0.5cm}, clip]{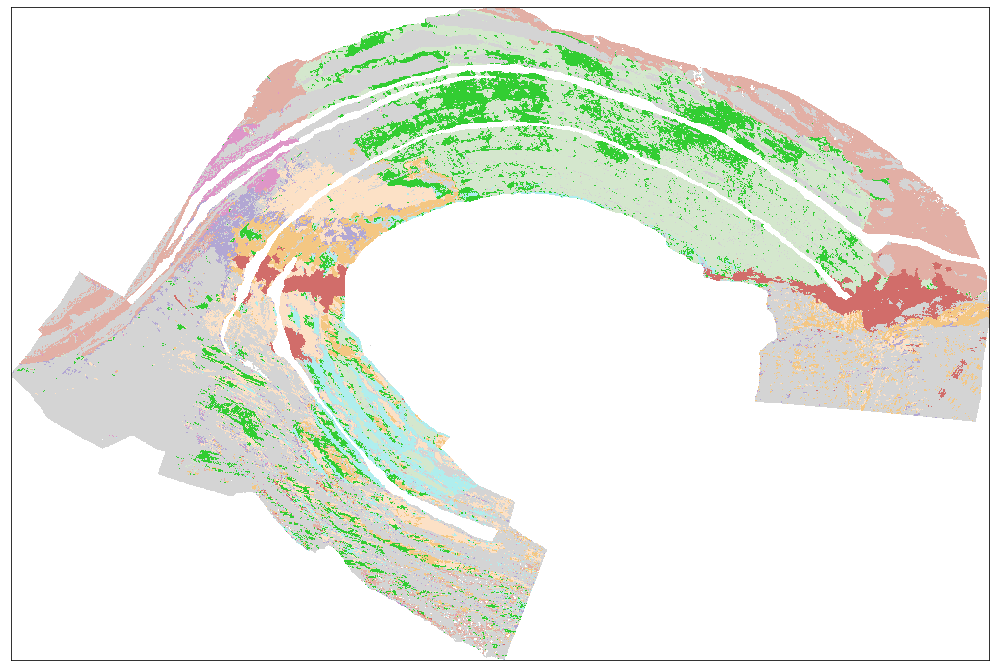}
    \caption{VNIR MLP noisy}
\end{subfigure}%
\begin{subfigure}{0.24\textwidth}
    \centering
    \includegraphics[width=0.98\linewidth]{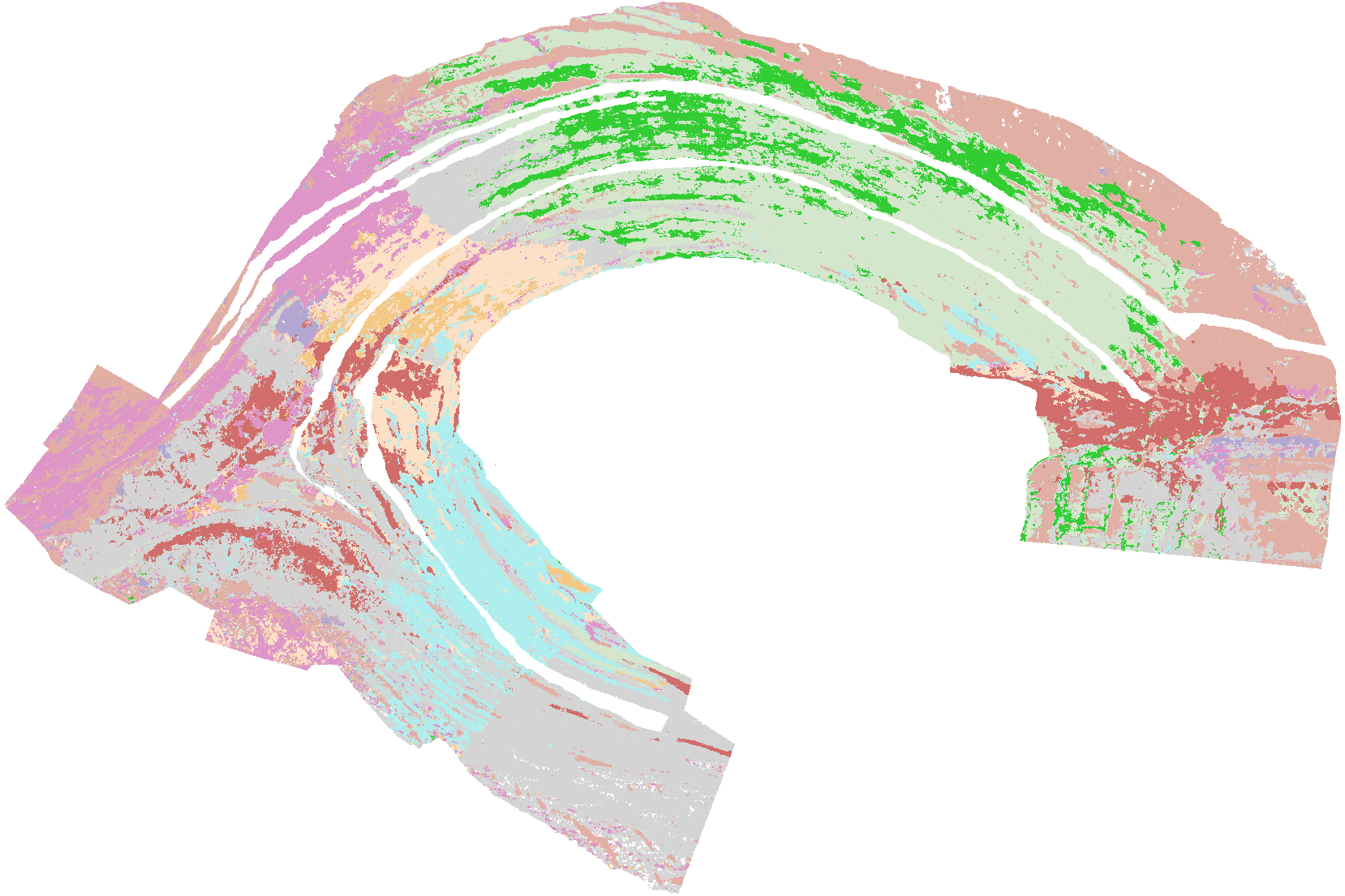} %
    \caption{VNIR PointNet real}
\end{subfigure}%

\begin{subfigure}{0.24\textwidth}
    \centering
    \includegraphics[width=0.98\linewidth]{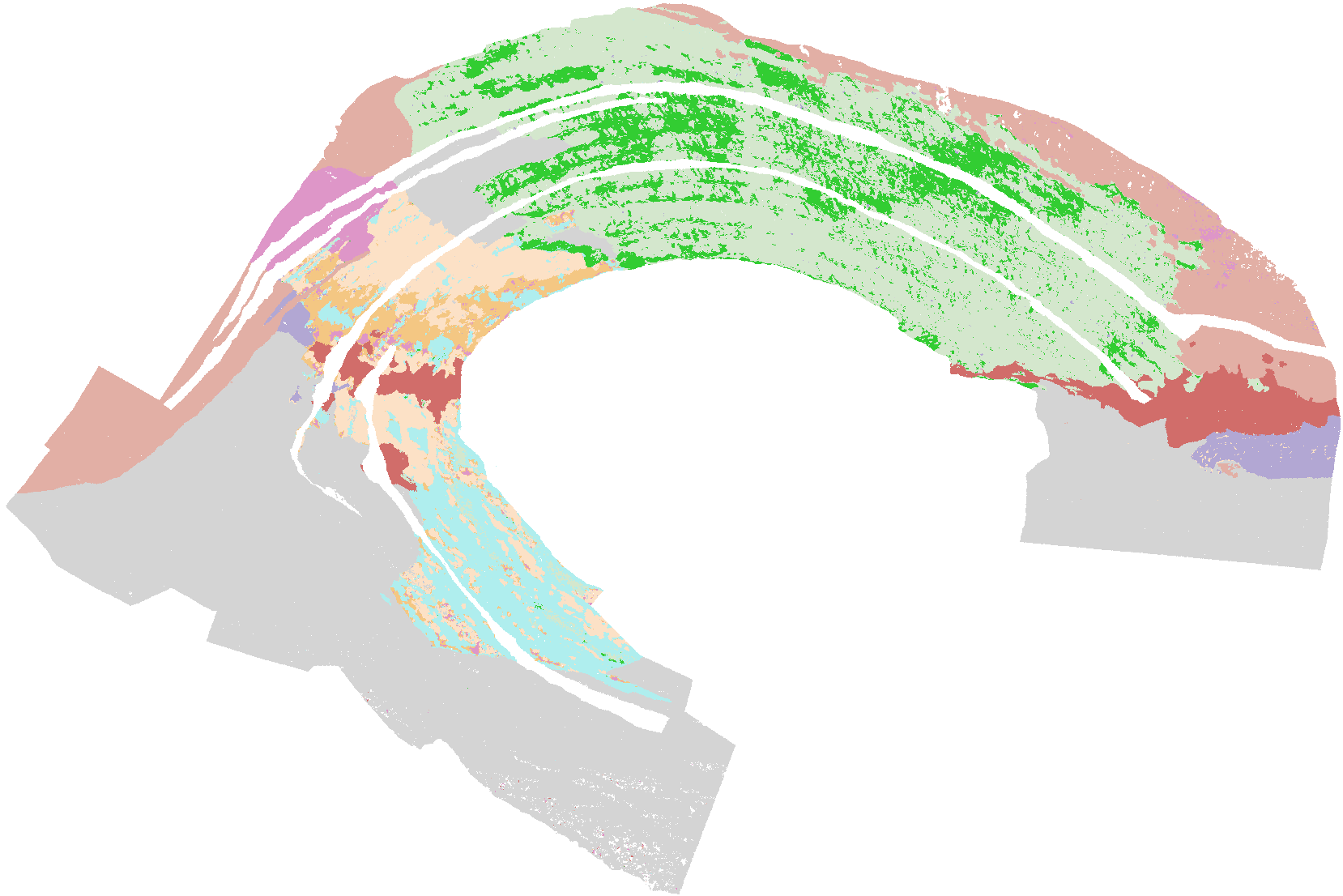}
    \caption{VNIR PointNet clean}
\end{subfigure}%
\begin{subfigure}{0.24\textwidth}
    \centering
    \includegraphics[width=0.98\linewidth]{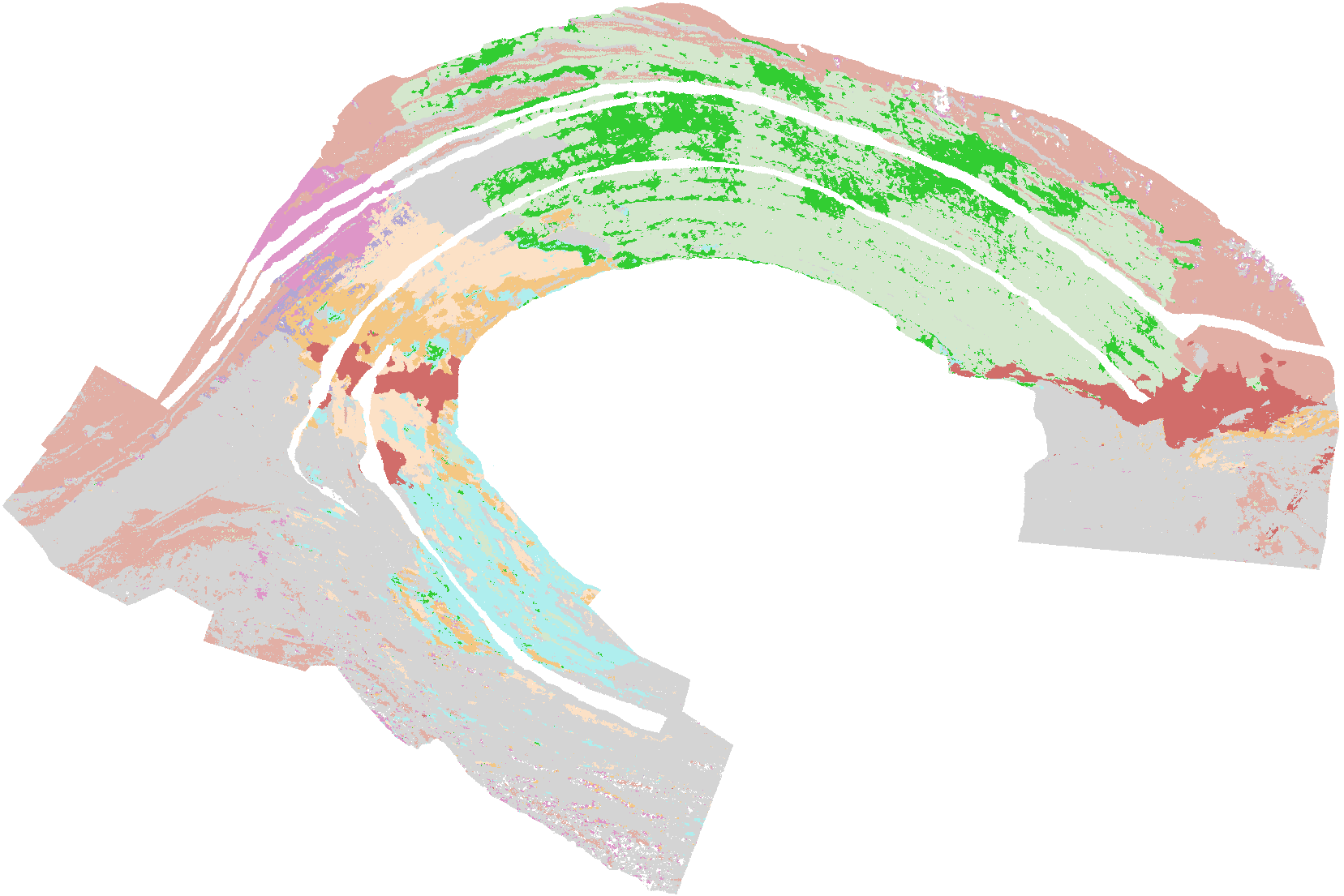}
    \caption{VNIR PointNet noisy}
\end{subfigure}%
\begin{subfigure}{0.24\textwidth}
    \centering
    \includegraphics[width=0.98\linewidth]{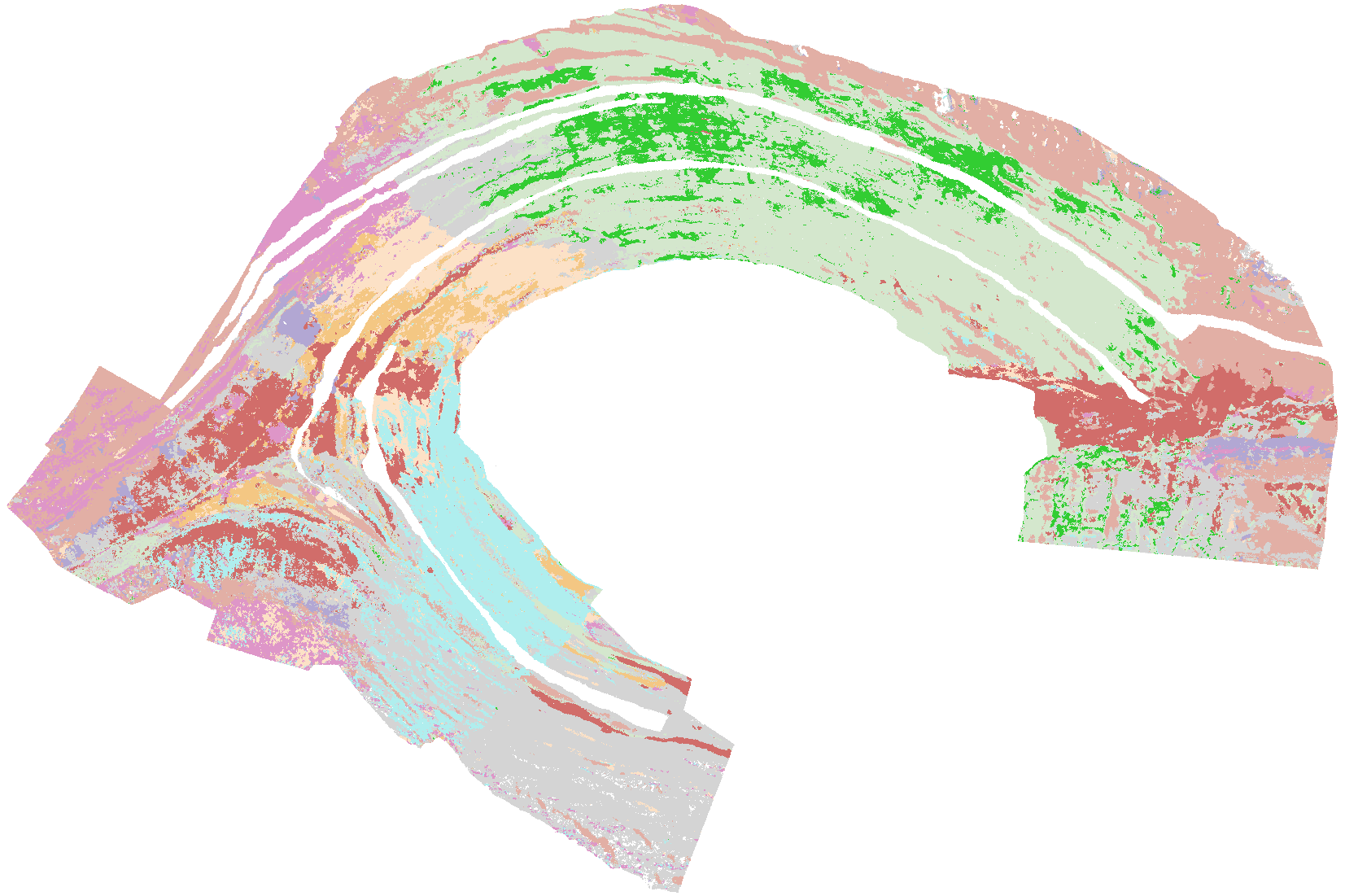}
    \caption{VNIR PointNet2 real}
\end{subfigure}%
\begin{subfigure}{0.24\textwidth}
    \centering
    \includegraphics[width=0.98\linewidth]{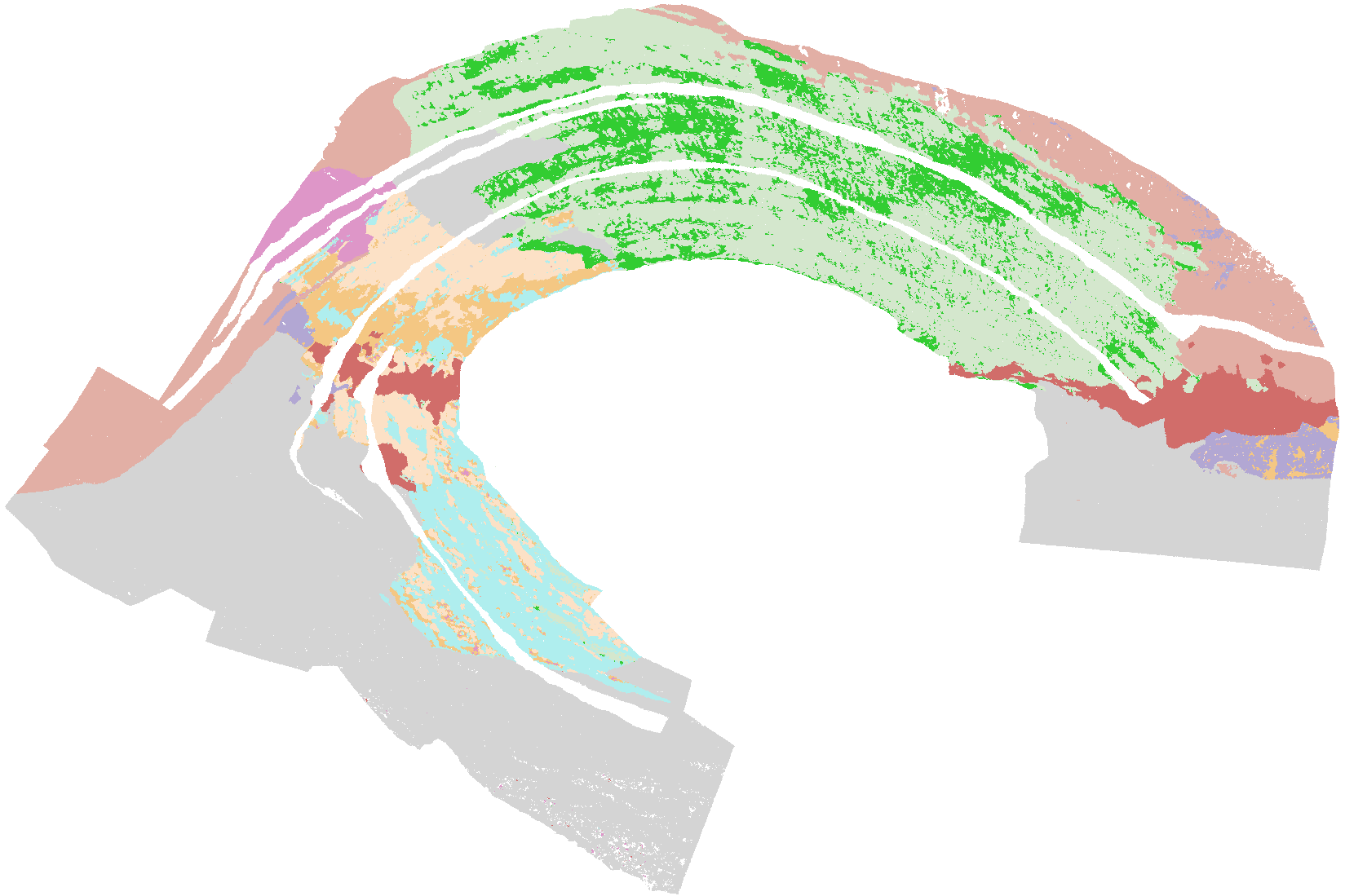}
    \caption{VNIR PointNet2 clean}
\end{subfigure}%

\begin{subfigure}{0.24\textwidth}
    \centering
    \includegraphics[width=0.98\linewidth, trim = {0.5cm 0.5cm 0.5cm 0.5cm}, clip]{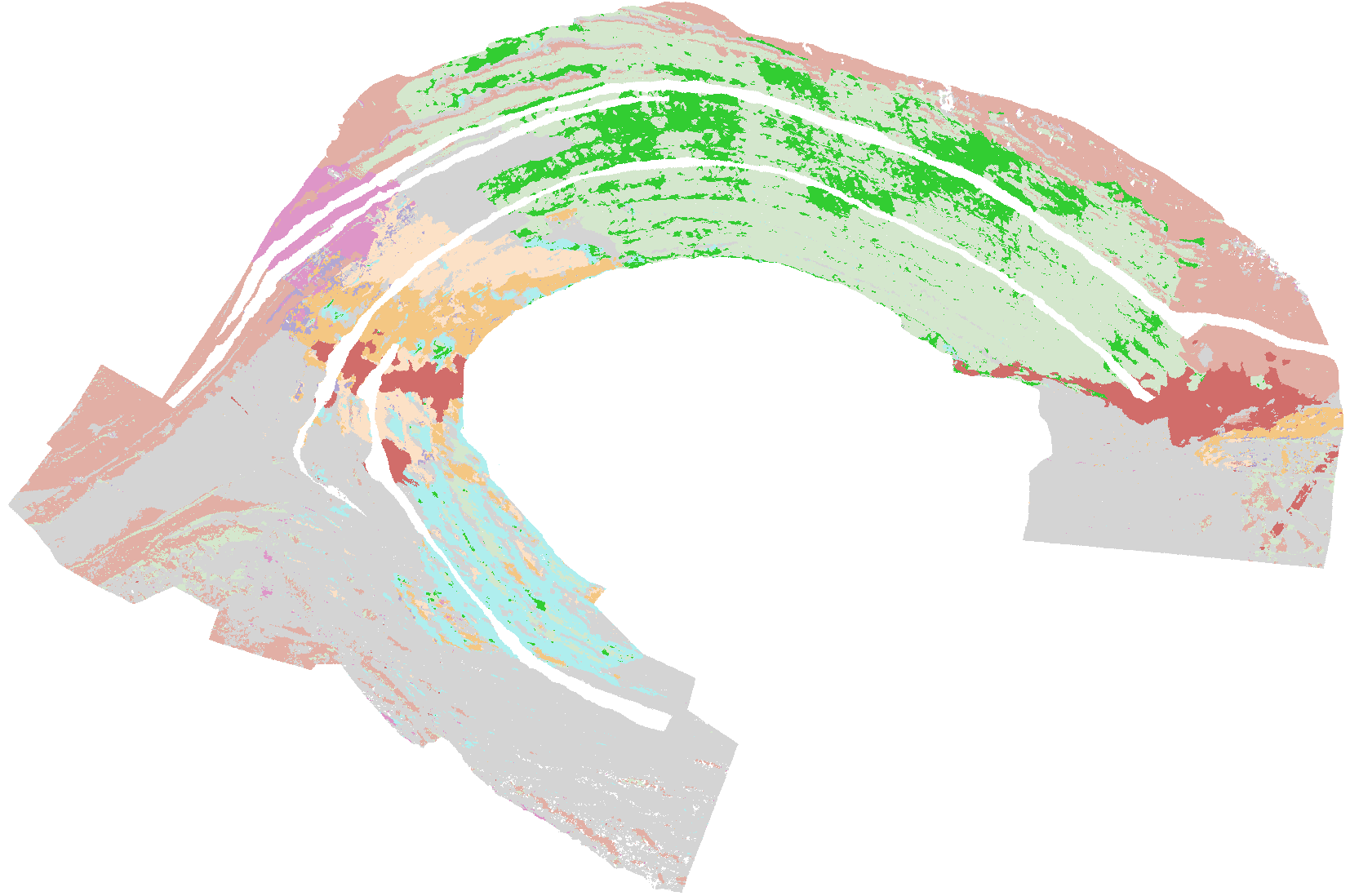}
    \caption{VNIR PointNet2 noisy}
\end{subfigure}%
\begin{subfigure}{0.24\textwidth}
    \centering
    \includegraphics[width=0.98\linewidth]{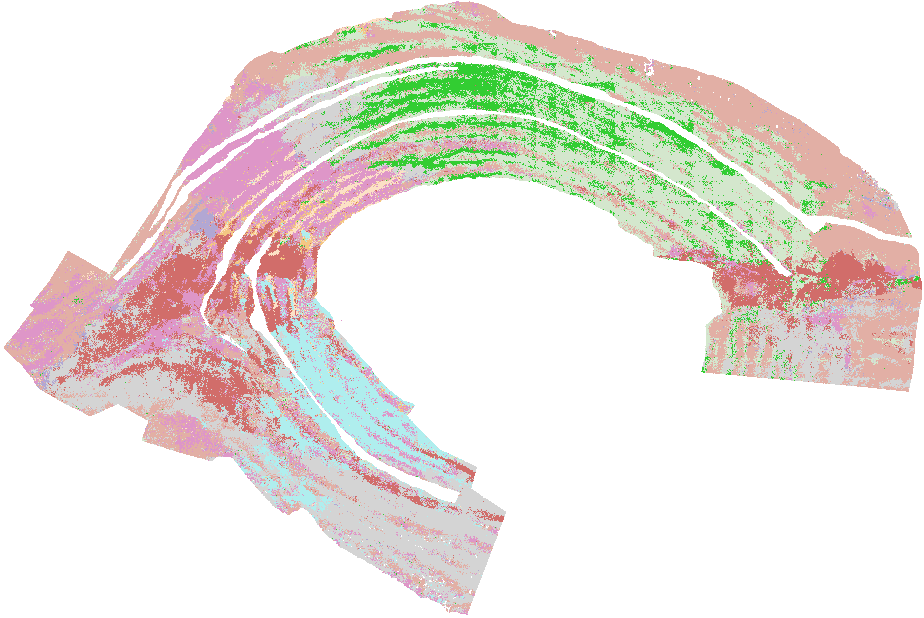} %
    \caption{VNIR PointCNN real}
\end{subfigure}%
\begin{subfigure}{0.24\textwidth}
    \centering
    \includegraphics[width=0.98\linewidth]{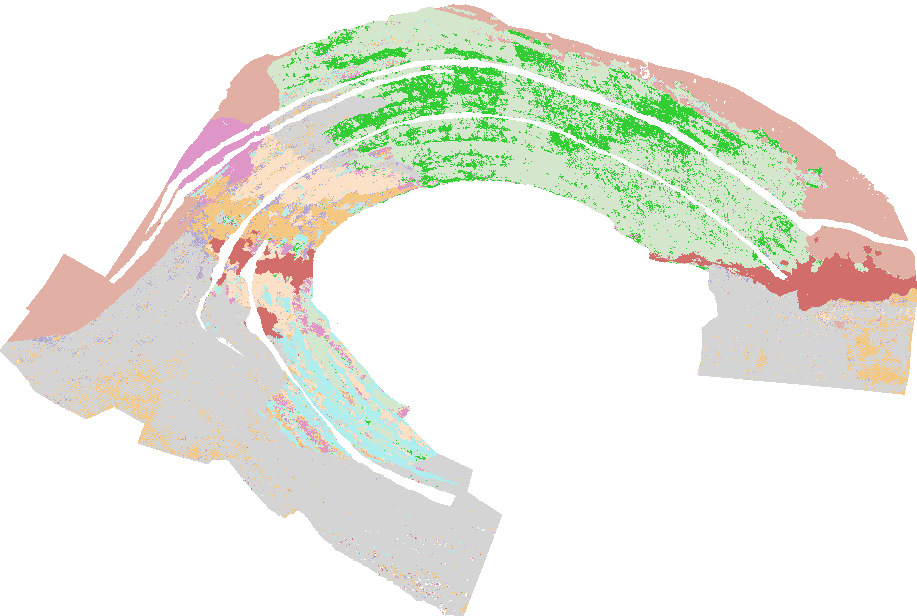} %
    \caption{VNIR PointCNN clean}
\end{subfigure}%
\begin{subfigure}{0.24\textwidth}
    \centering
    \includegraphics[width=0.98\linewidth]{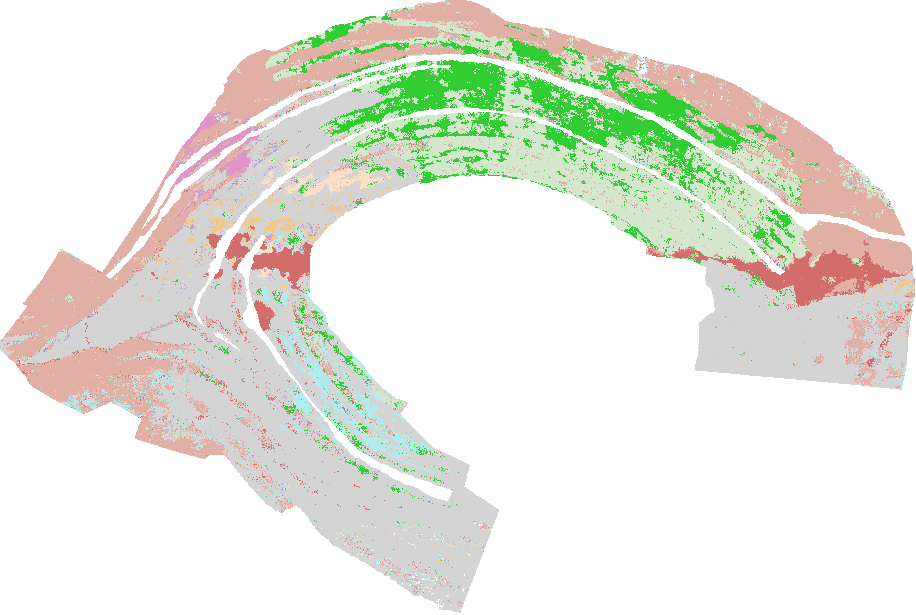} %
    \caption{VNIR PointCNN noisy}
\end{subfigure}%

\begin{subfigure}{0.24\textwidth}
    \centering
    \includegraphics[width=0.98\linewidth]{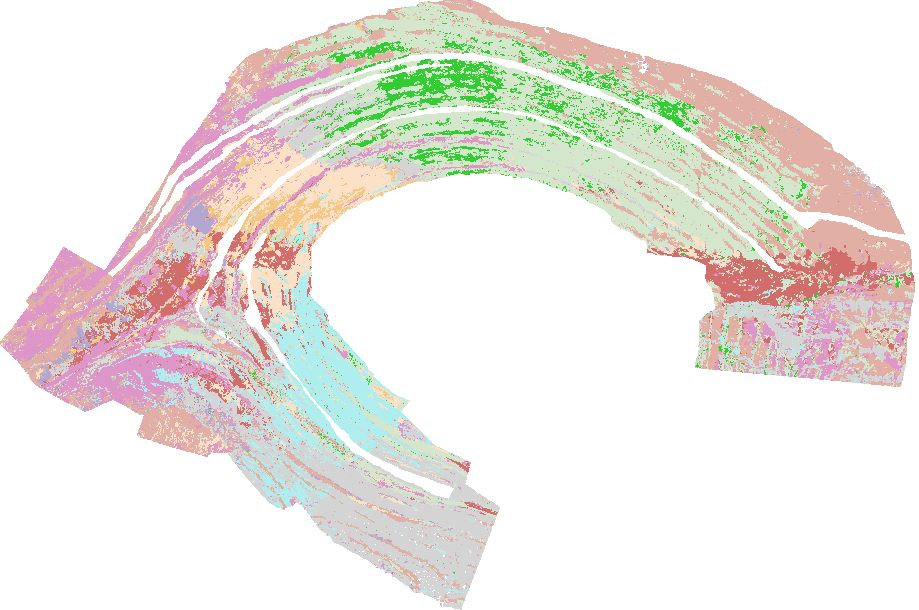} %
    \caption{VNIR ConvPoint real}
\end{subfigure}%
\begin{subfigure}{0.24\textwidth}
    \centering
    \includegraphics[width=0.98\linewidth]{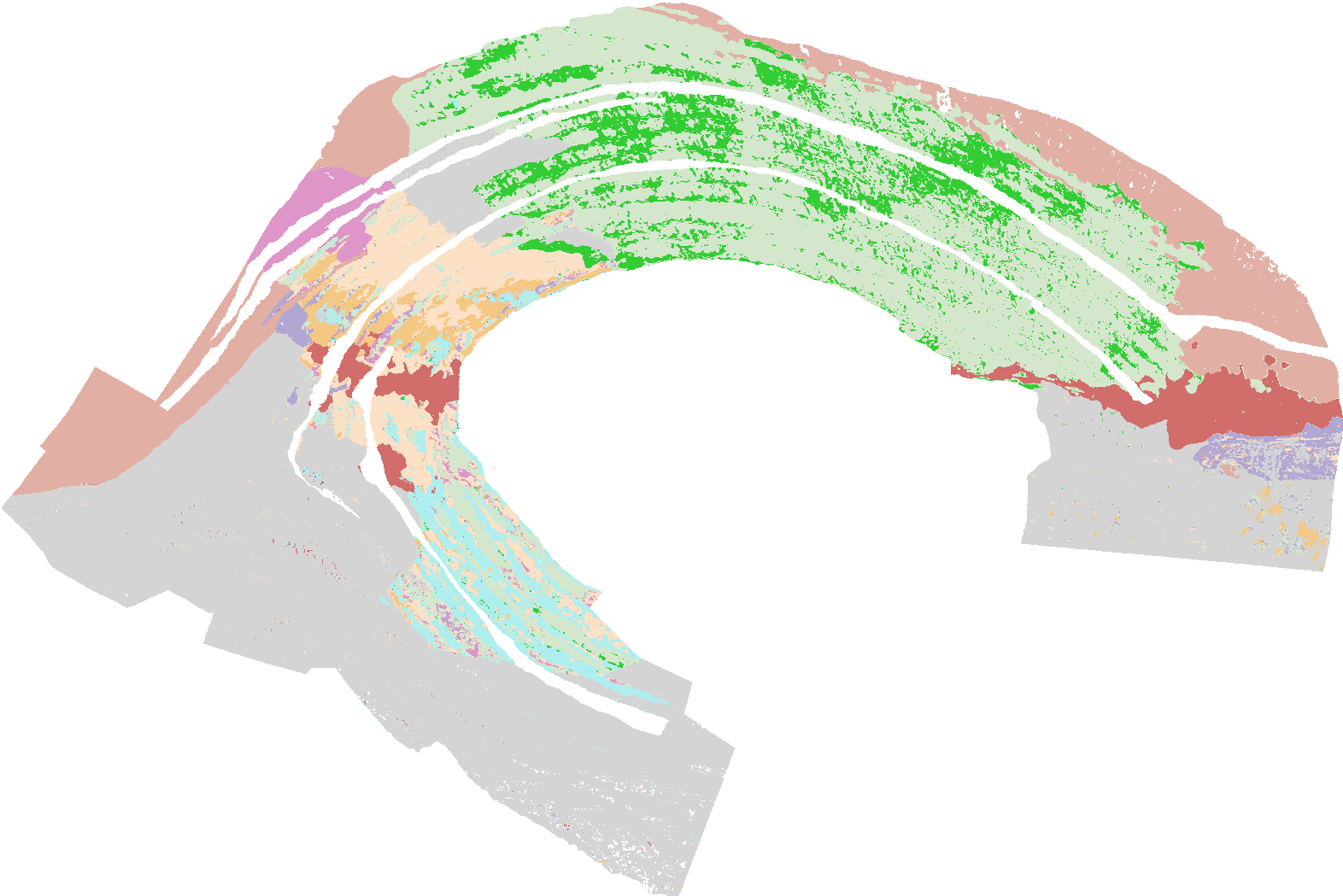} %
    \caption{VNIR ConvPoint clean}
\end{subfigure}%
\begin{subfigure}{0.24\textwidth}
    \centering
    \includegraphics[width=0.98\linewidth]{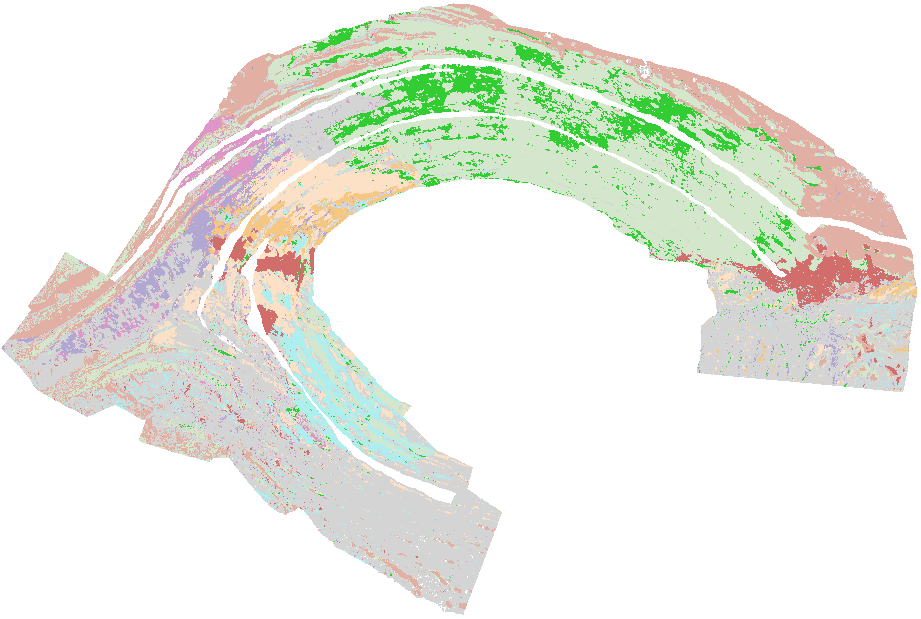} %
    \caption{VNIR ConvPoint noisy}
\end{subfigure}%
\begin{subfigure}{0.24\textwidth}
    \centering
    \includegraphics[width=0.98\linewidth]{figures/VNIR/vnir_real_dgcnn.png} %
    \caption{VNIR DGCNN real}
\end{subfigure}%

\begin{subfigure}{0.24\textwidth}
    \centering
    \includegraphics[width=0.98\linewidth]{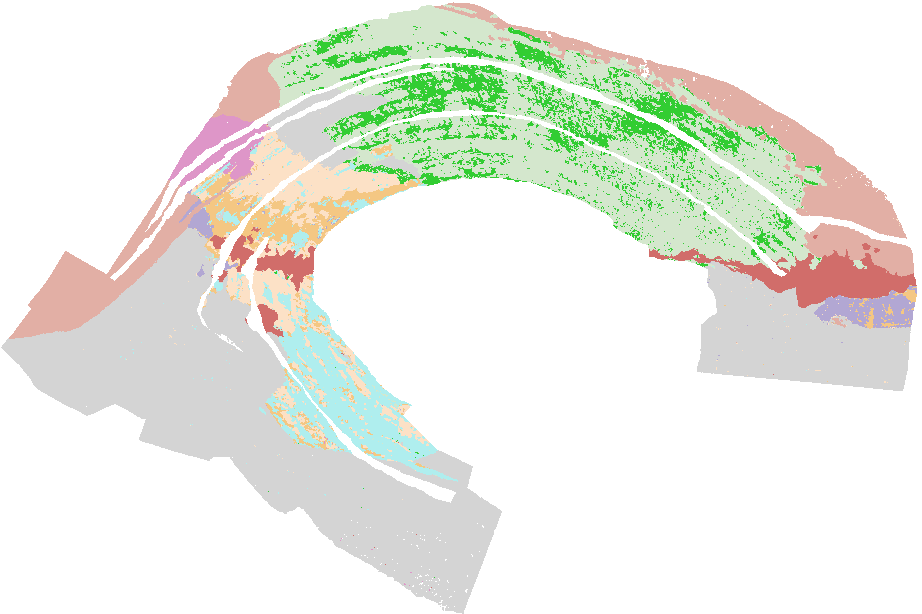} %
    \caption{VNIR DGCNN clean}
\end{subfigure}%
\begin{subfigure}{0.24\textwidth}
    \centering
    \includegraphics[width=0.98\linewidth]{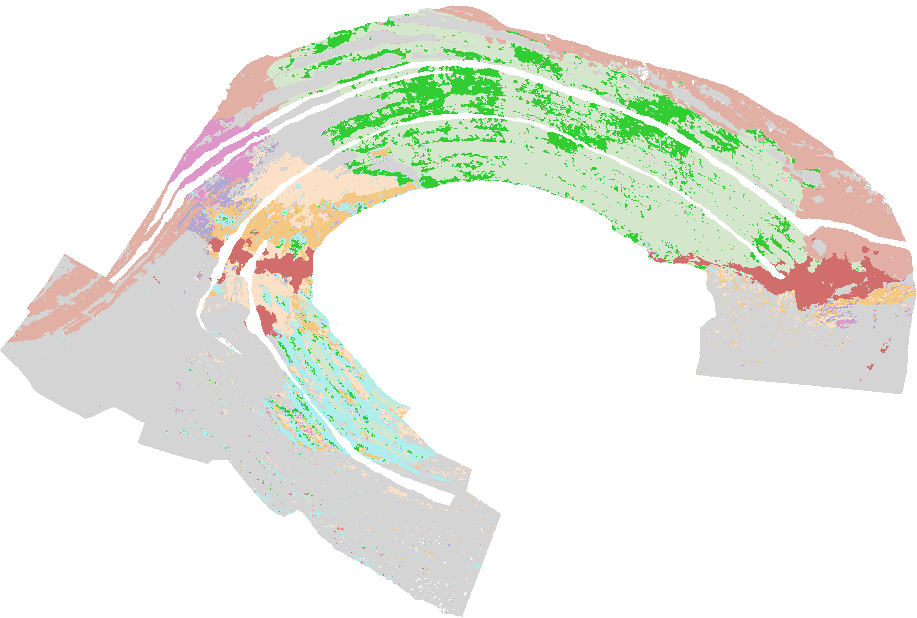} %
    \caption{VNIR DGCNN noisy}
\end{subfigure}%
\begin{subfigure}{0.24\textwidth}
    \centering
    \includegraphics[width=0.98\linewidth]{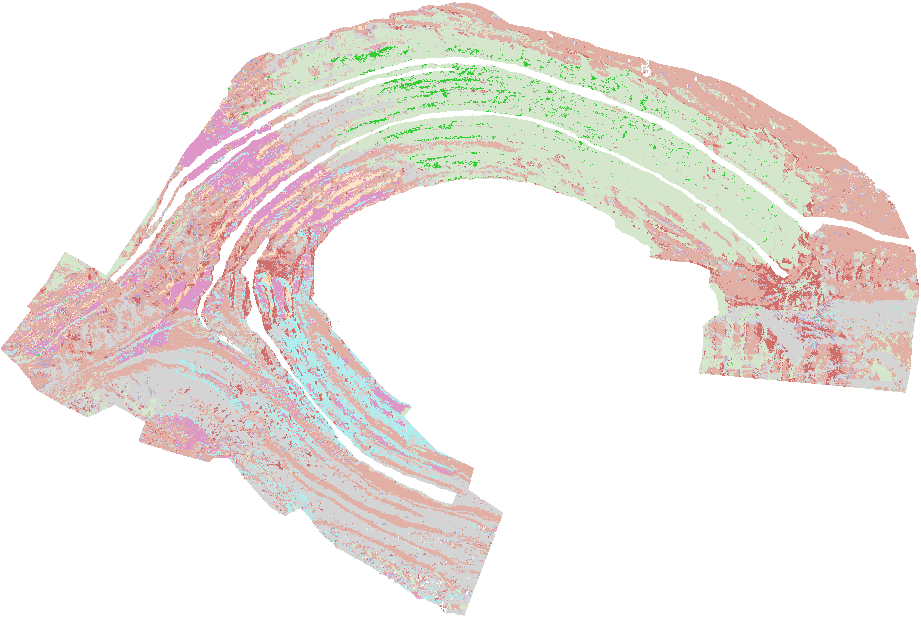} %
    \caption{VNIR PT real}
\end{subfigure}%
\begin{subfigure}{0.24\textwidth}
    \centering
    \includegraphics[width=0.98\linewidth]{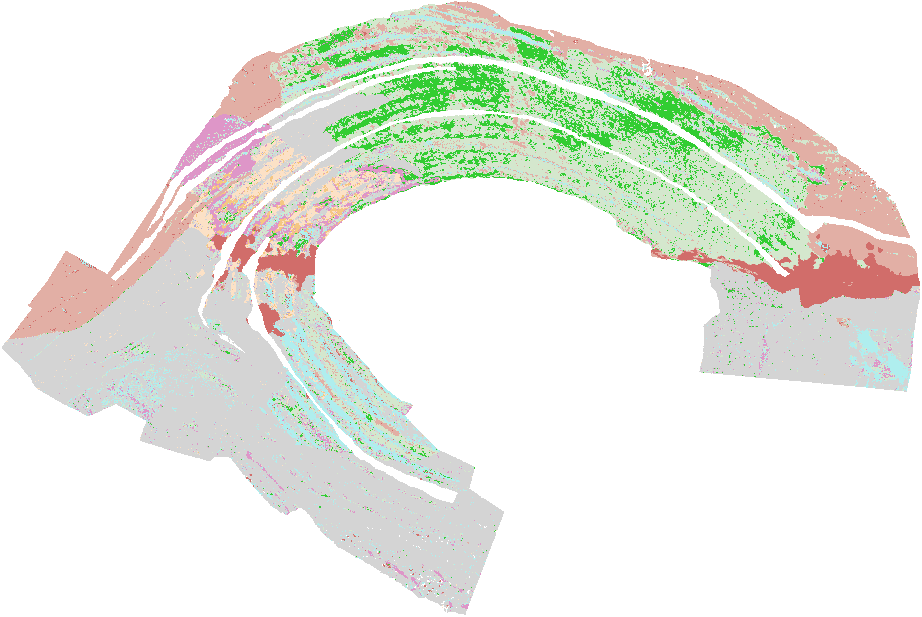} %
    \caption{VNIR PT clean}
\end{subfigure}%

\begin{subfigure}{0.24\textwidth}
    \centering
    \includegraphics[width=0.98\linewidth]{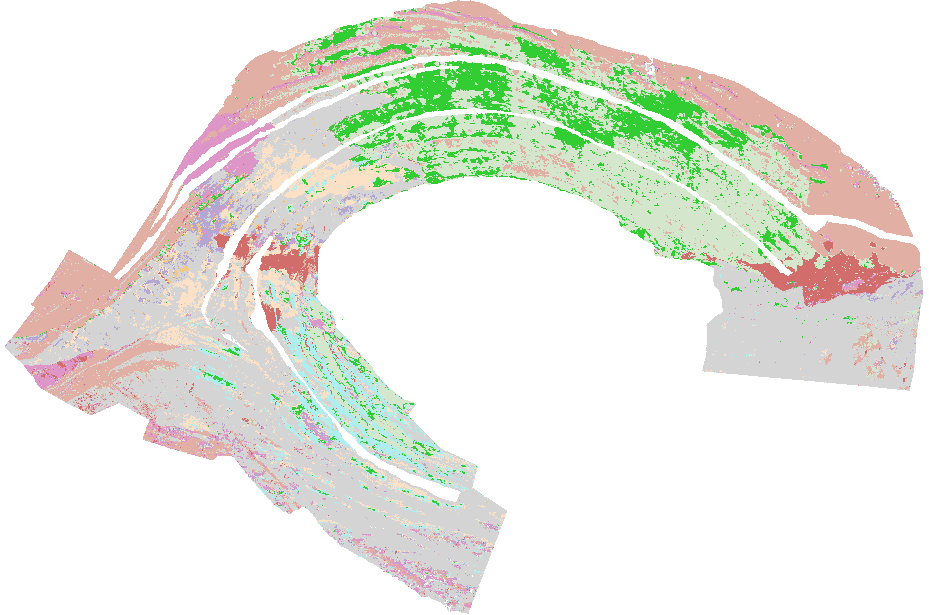} %
    \caption{VNIR PT noisy}
\end{subfigure}%
\begin{subfigure}{0.24\textwidth}
    \centering
    \includegraphics[width=0.98\linewidth]{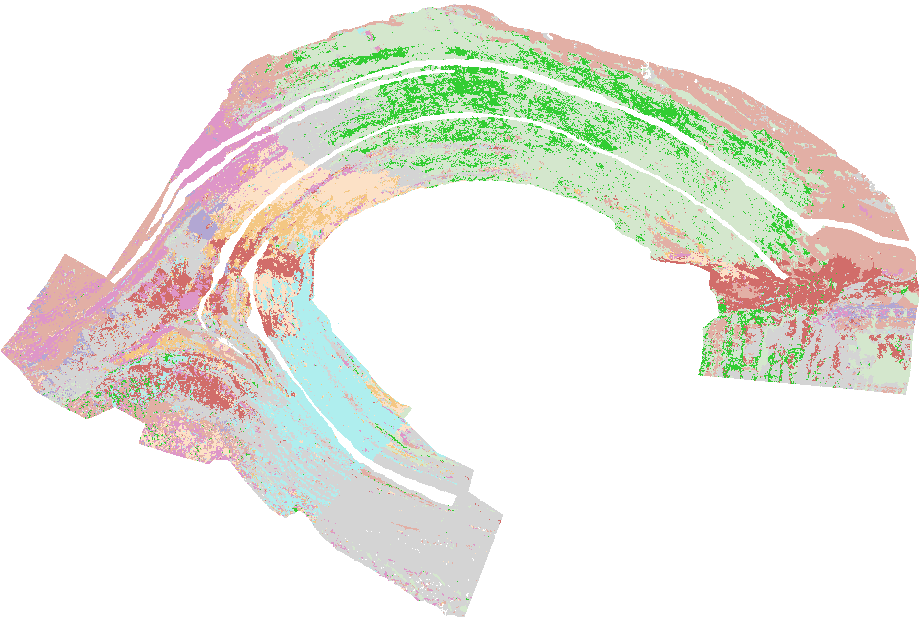} %
    \caption{VNIR PCT real}
\end{subfigure}%
\begin{subfigure}{0.24\textwidth}
    \centering
    \includegraphics[width=0.98\linewidth]{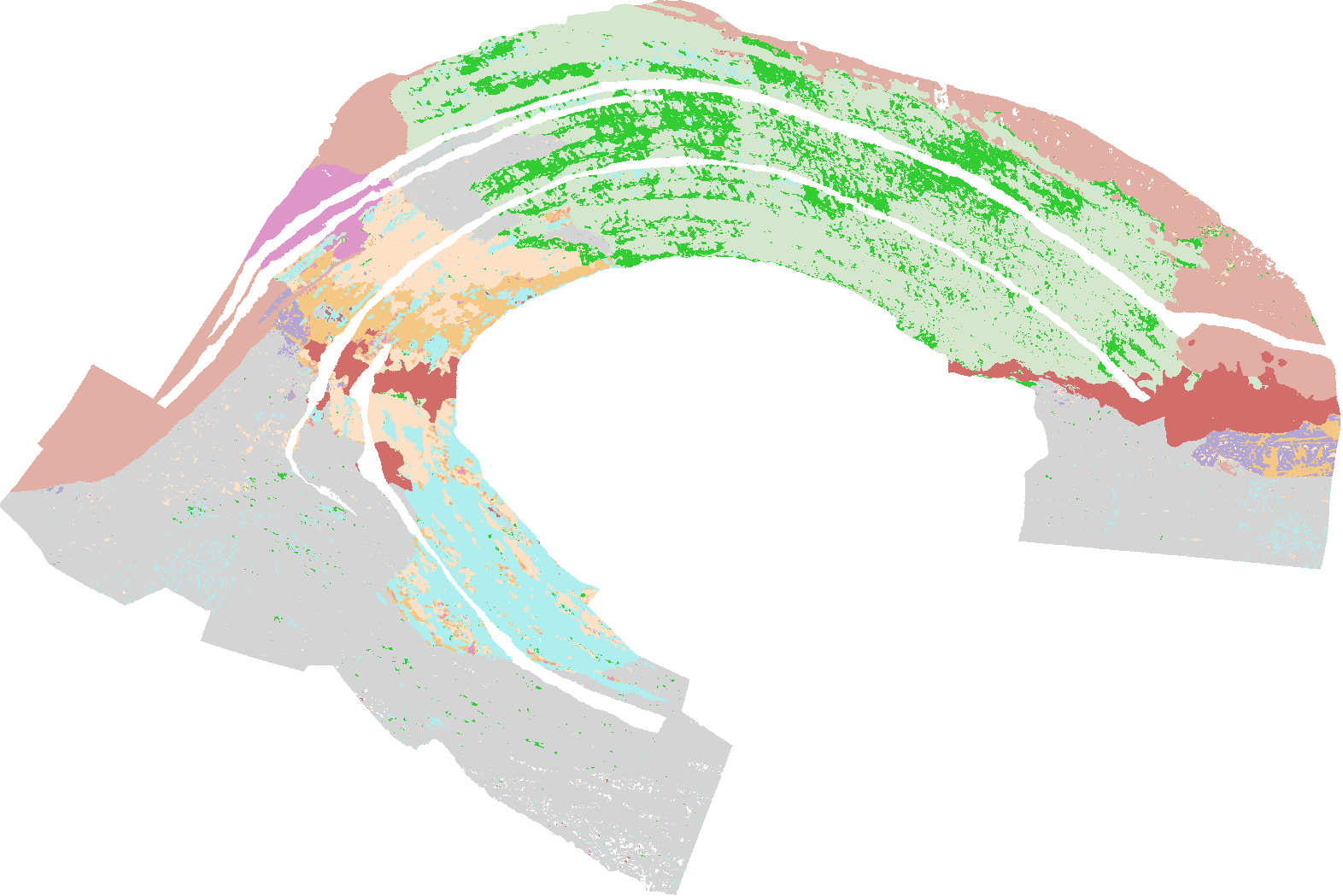} %
    \caption{VNIR PCT clean}
\end{subfigure}%
\begin{subfigure}{0.24\textwidth}
    \centering
    \includegraphics[width=0.98\linewidth]{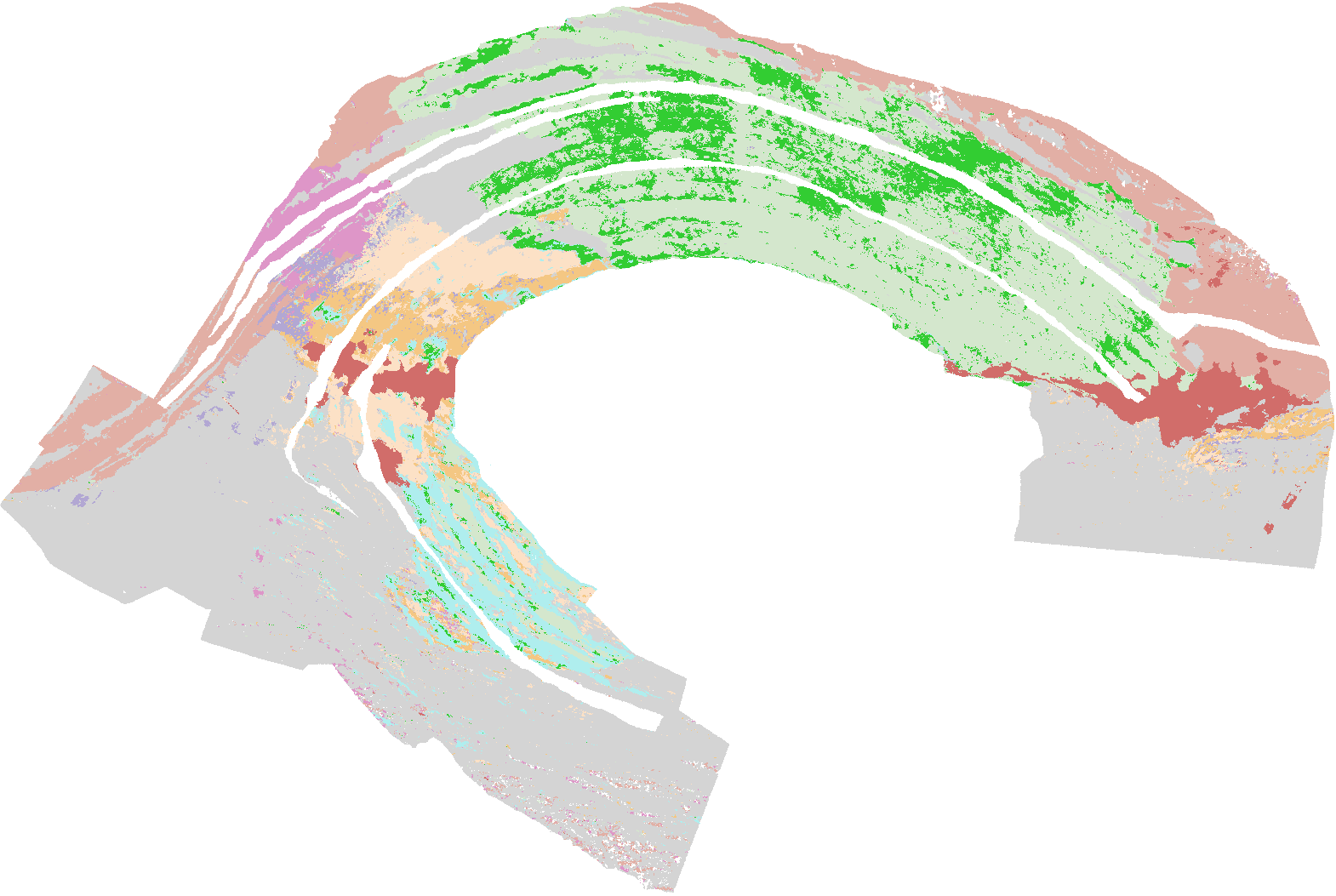} %
    \caption{VNIR PCT noisy}
\end{subfigure}%

\caption{Segmentation results of the baseline models on the VNIR test in various scenarios.}
\label{VNIR_results}
\end{figure*}

Figure \ref{LWIR_results}, Figure \ref{SWIR_results}, and Figure \ref{VNIR_results} showcase the qualitative outcomes of segmented point clouds generated by the trained baseline models using the testing split of the dataset on the LWIR, SWIR, and VNIR data, respectively in different scenarios (real data, clean synthetic data, noisy synthetic data). These illustrations highlight the baseline models' ability to produce segmented point clouds with a reasonably high degree of accuracy. When applied to clean synthetic data, most models excel in accurately segmenting the point cloud. However, it's crucial to acknowledge that real-world data typically contains noise, and our models exhibit decreased performance when noise is introduced into the synthetic data. Furthermore, the presence of imbalanced data can lead to misclassifications, particularly in instances where certain classes have a disproportionately smaller number of training samples compared to others. This issue becomes particularly noticeable in the case of the purple shale class, both in real data and noisy synthetic data scenarios.

%%%%%%%%%%%%%%%%%%%%%%%%%%%%%%%%%%%%%%%%%%%%%%%%%%%%%%%%%%%%%%%%

%%%%%%%%%%%%%%%%%%%%%%%%%%%%%%%%%%%%%%%%%%%%%%%%%%%%%%%%%%%%%%%%

%%%%%%%%%%%%%%%%%%%%%%%%%%%%%%%%%%%%%%%%%%%%%%%%%%%%%%%%%%%%%%%%

\section{Conclusion} \label{conclusion}

This paper introduces the first-ever fully-labeled multi-sensor hyperspectral benchmark dataset and we expect it to be a valuable resource for researchers working on point cloud segmentation in geosciences. The dataset is comprehensive and diverse, covering real and synthetic data with different levels of labeling for the classes. The Tinto dataset is a suitable benchmark for developing and evaluating point cloud segmentation algorithms for geological applications. We believe that the Tinto dataset will serve as a benchmark for future studies and contribute to the development of innovative solutions for point cloud segmentation in Earth sciences. Interestingly, deep models that leverage information from neighboring points demonstrate superior performance, thanks to their ability to extract both local and global features for point cloud segmentation. The strength of this dataset is that it is versatile and allows testing architecture robustness under different conditions (e.g. noise, data quality). Overall, the Tinto benchmark dataset represents a significant contribution to the field and holds promise for a broad range of applications in Earth sciences such as mineral exploration, geological mapping, and natural resources management.

\section*{Acknowledgment}

The authors would like to acknowledge extensive support from Atalaya Mining during fieldwork conducted for this publication, and subsequent validation of the results. This research received funding from the Initiative and Networking Fund (INF) of the Hermann von Helmholtz Association of German Research Centres in the framework of the Helmholtz Imaging Platform (HIP) under grant agreement No ZT-I-PF-4-021 and from the European Union’s Horizon 2020 research and innovation programme under grant agreement No 776487. The experiments were conducted on the high specification Nvidia A100 GPUs server which was provided by the European Regional Development Fund and the Land of Saxony.

\ifCLASSOPTIONcaptionsoff
  \newpage
\fi

\bibliographystyle{IEEEtran}
\bibliography{references}

\begin{IEEEbiography}[{\includegraphics[width=1in,height=1.25in,clip,keepaspectratio]{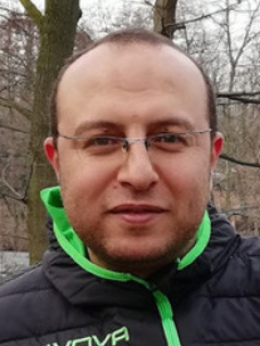}}]{Ahmed J. Afifi} received the bachelor’s and M.Sc. degrees in computer engineering from the Islamic University of Gaza (IUG), in 2008 and 2011, respectively. He completed his Ph.D. degree (Dr.-Ing.) at Technische Universität Berlin, Germany in 2021. Currently, he is a postdoctoral researcher with Helmholtz-Institute Freiberg for Resource Technology (HIF) and Karlsruhe Institute of Technology (KIT) where he is working on 3D point cloud classification and segmentation. His research interests include computer vision, deep learning, 3D object reconstruction from a single image, and medical image analysis.
\end{IEEEbiography}


\begin{IEEEbiography}[{\includegraphics[width=1in,height=1.25in,clip,keepaspectratio]{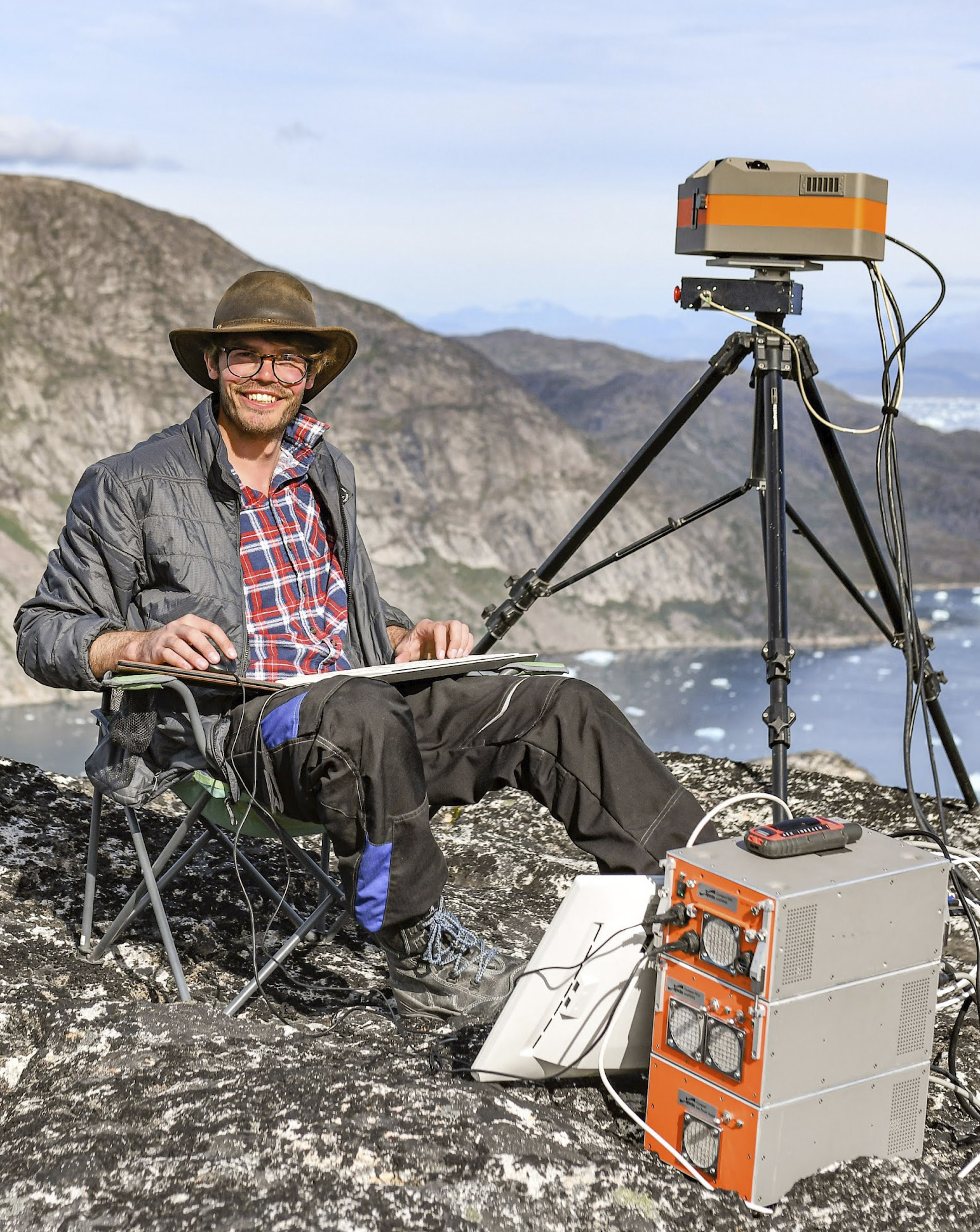}}]{Samuel T. Thiele}
received his M.Sc in Geology from the University of Western Australia in Perth. Following this, he was awarded a Westpac Future Leader scholarship to pursue a PhD project on the application of digital mapping technologies to volcanic landslides and magma plumbing systems. Since 2019 he has worked at the Helmholtz-Institute Freiberg for Resource Technology, with a focus on outcrop and drillcore hyperspectral remote sensing techniques and their application to minerals exploration and digital outcrop mapping.  
\end{IEEEbiography}

\begin{IEEEbiography}[{\includegraphics[width=1in,height=1.25in,clip,keepaspectratio]{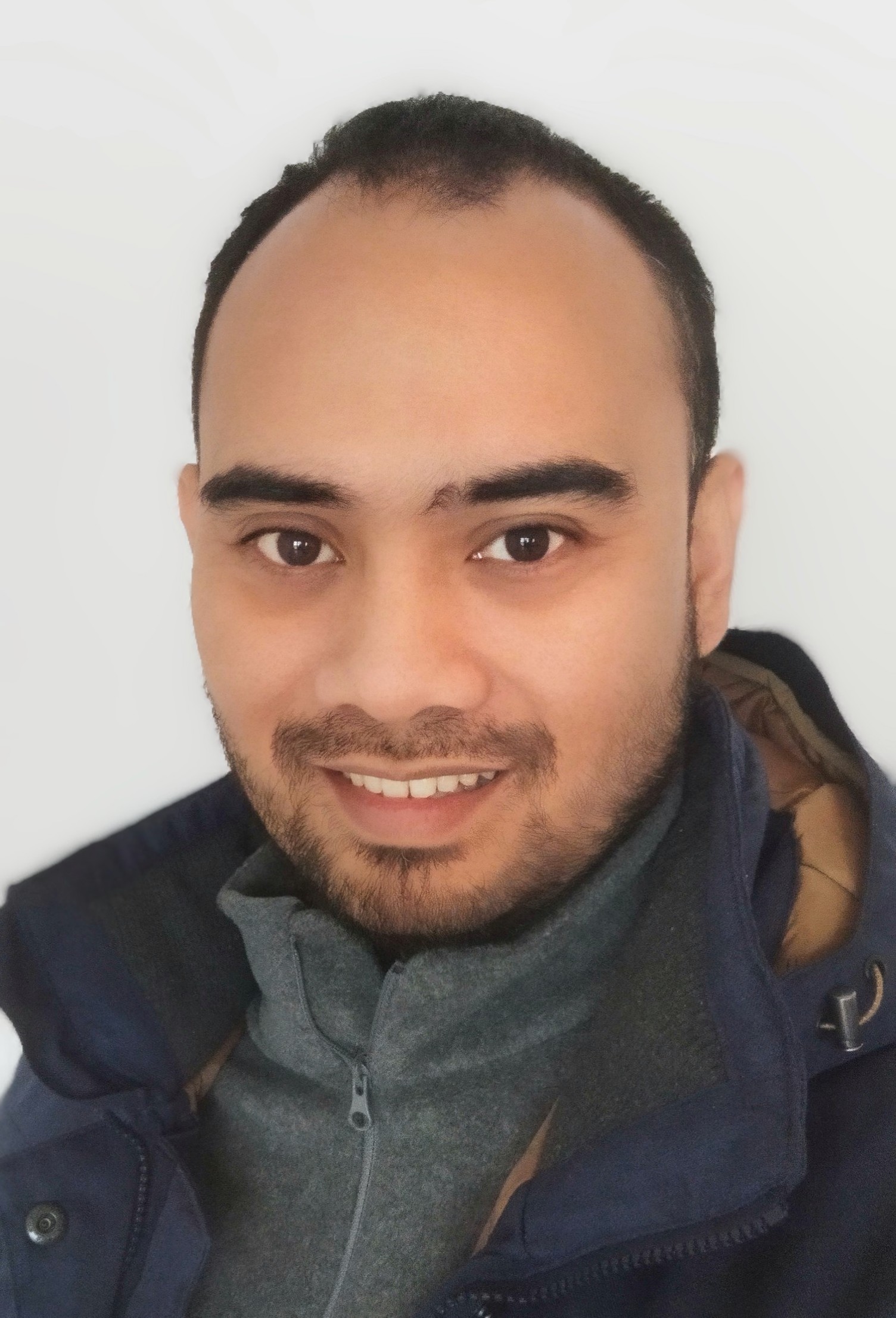}}]{Aldino Rizaldy} received his M.Sc. degree in Geoinformation from Faculty ITC, University of Twente, The Netherlands. Currently, he is a Ph.D. student at the Helmholtz-Institute Freiberg for Resource Technology. His main research interests are deep learning, computer vision, 3D point cloud, LiDAR and photogrammetry.
\end{IEEEbiography}

\begin{IEEEbiography}[{\includegraphics[width=1in,height=1.25in,clip,keepaspectratio]{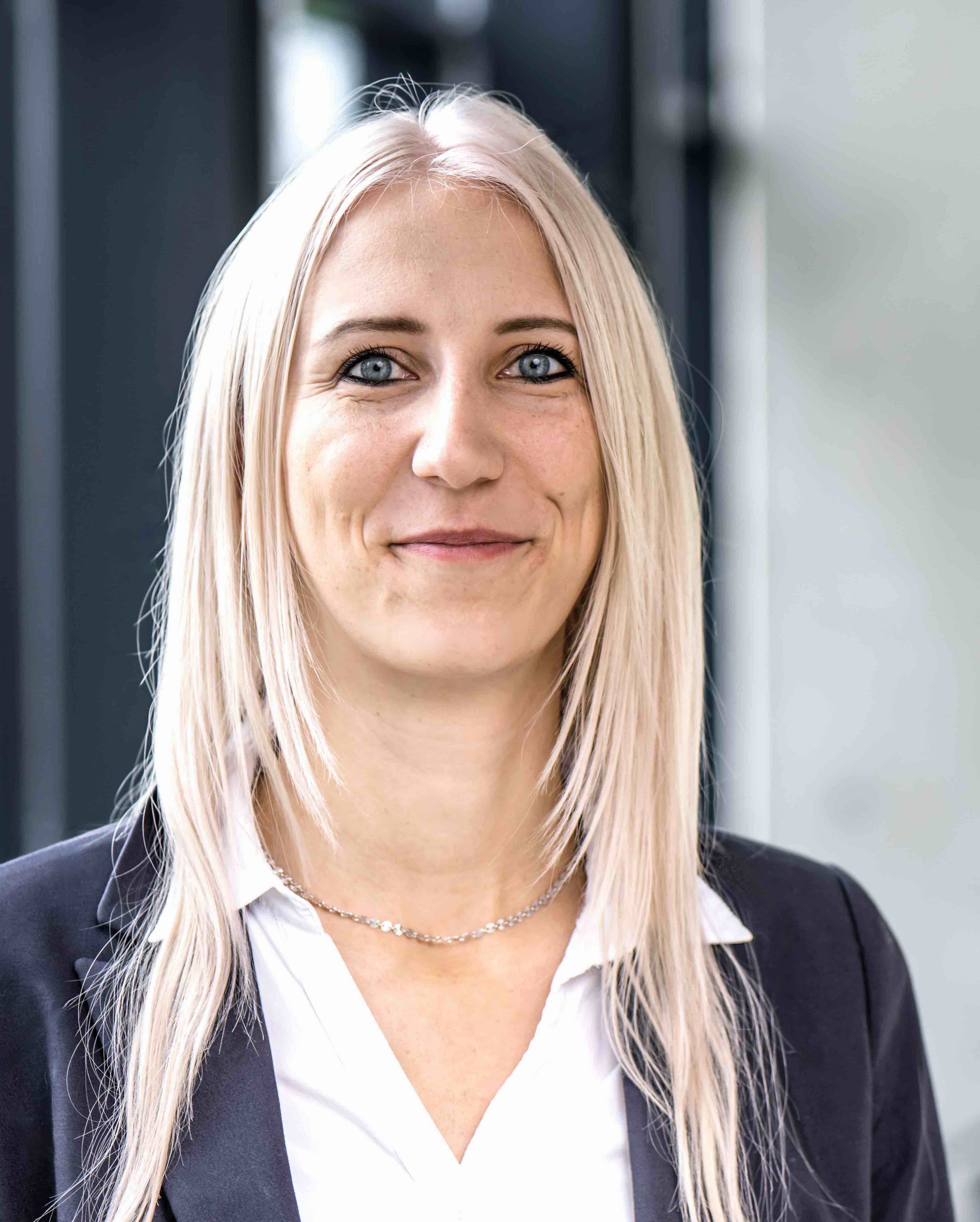}}]{Sandra Lorenz}
is research associate at the Helmholtz-Institute Freiberg for Resource Technology, where she currently leads the research group for Digital Processing within the Department of Exploration. She received the Ph.D. degree “Doctor rerum naturalis” in Geosciences at the Technical University Bergakademie Freiberg, Germany in 2019. Her main research interest lies in the development of spectral imaging as non-invasive material characterisation method at different scales, with a particular focus on spectral image and point-cloud processing and the implementation of autonomous platforms for hyperspectral surveys. 
\end{IEEEbiography}

\begin{IEEEbiography}[{\includegraphics[width=1in,height=1.25in,clip,keepaspectratio]{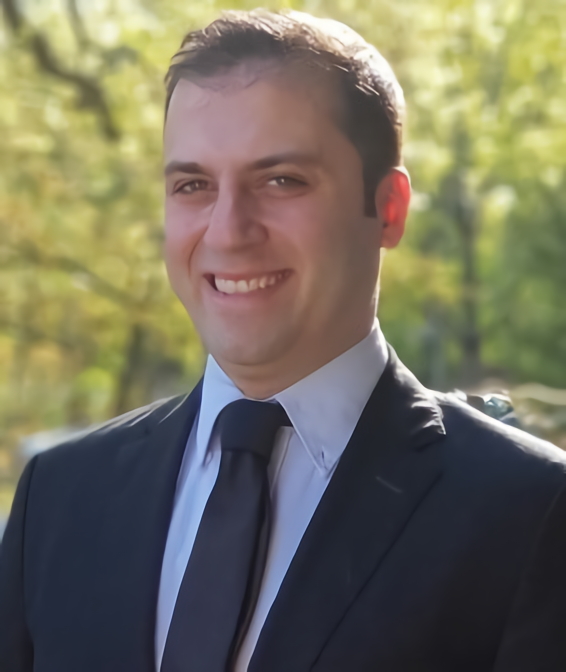}}]{Pedram Ghamisi}
(Senior Member, IEEE) earned his Ph.D. in Electrical and Computer Engineering from the University of Iceland. Currently, he serves as the Head of the Machine Learning Group at the Helmholtz-Institute Freiberg for Resource Technology in Germany. He also holds the position of Research Professor and Senior Principal Investigator, leading the AI for Remote Sensing (AI4RS) team, at the Institute of Advanced Research in Artificial Intelligence (IARAI) in Austria. In addition to his roles in Germany and Austria, he holds a visiting professorship at Lancaster University, UK. He specializes in deep learning for remote sensing applications.
\end{IEEEbiography}

\begin{IEEEbiography}[{\includegraphics[width=1in,height=1.25in,clip,keepaspectratio]{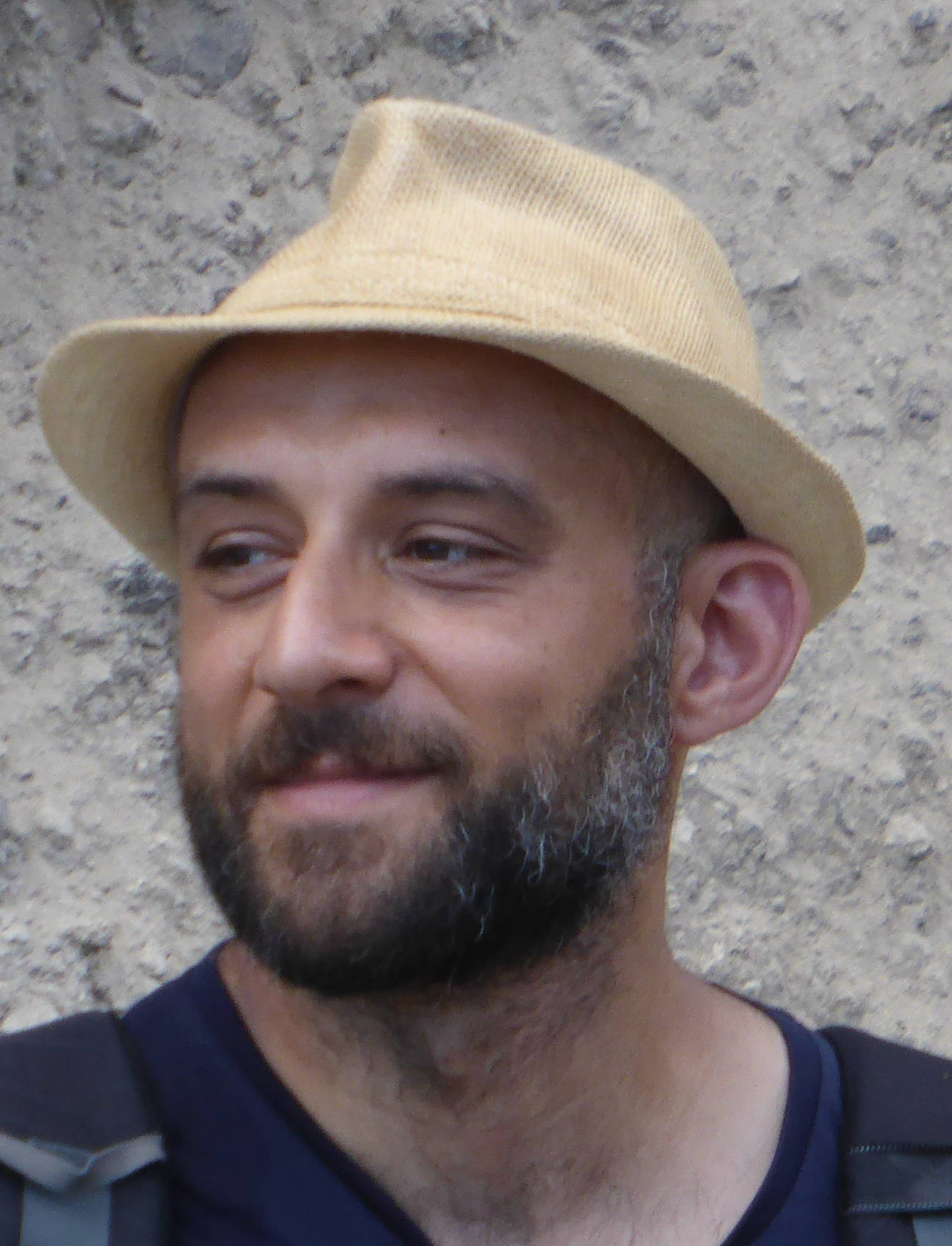}}]{Raimon Tolosana-Delgado}
is engineering geologist by the University of Barcelona and Technical University of Catalunya, Barcelona (Spain), doctor in environmental physics and technology by the University of Girona (Spain) and got  a professoral qualification in mathematical geosciences by the Technische Universit\"at Bergakademie Freiberg (Germany). He is specialist in geostatistics and compositional data analysis, and works at the Helmholtz-Institute Freiberg for Resource Technology applying stochastic modelling and data analysis to ore body modelling, geometallurgy and minerals processing modelling.
\end{IEEEbiography}

\begin{IEEEbiography}[{\includegraphics[width=1in,height=1.25in,clip,keepaspectratio]{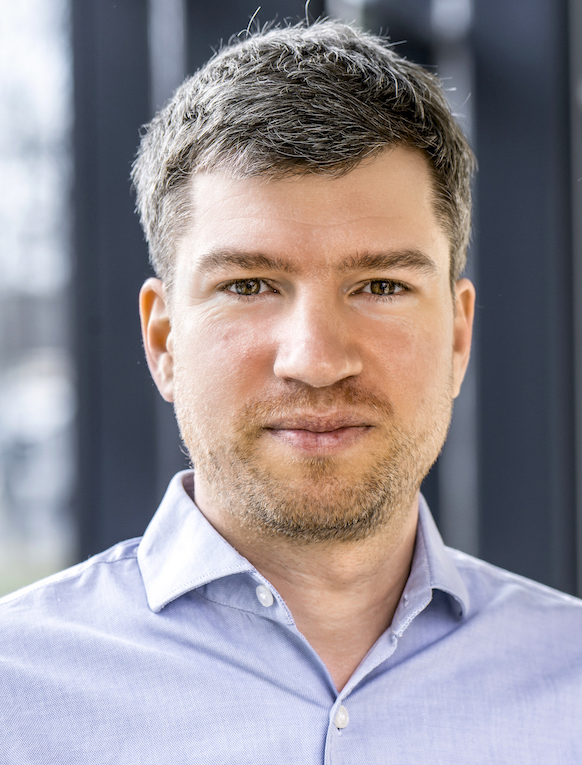}}]{Moritz Kirsch}
received his M.Sc. in Geology from the Technische Universität Bergakademie Freiberg, Germany, in 2008 and his Ph.D. from the Universidad Nacional Autónoma de México in 2012. Since 2016, he has been working as a research associate at the Helmholtz-Institute Freiberg for Resource Technology (HIF), Helmholtz-Zentrum Dresden-Rossendorf, where he leads a group on Innovative Applications in the Department of Exploration. His research interests lie at the intersection of geology and technology, with a focus on leveraging the latest digital tools, such as hyperspectral imaging and 3D modelling, to characterise mineral assemblages and understand geological processes.
\end{IEEEbiography}

\begin{IEEEbiography}[{\includegraphics[width=1in,height=1.25in,clip,keepaspectratio]{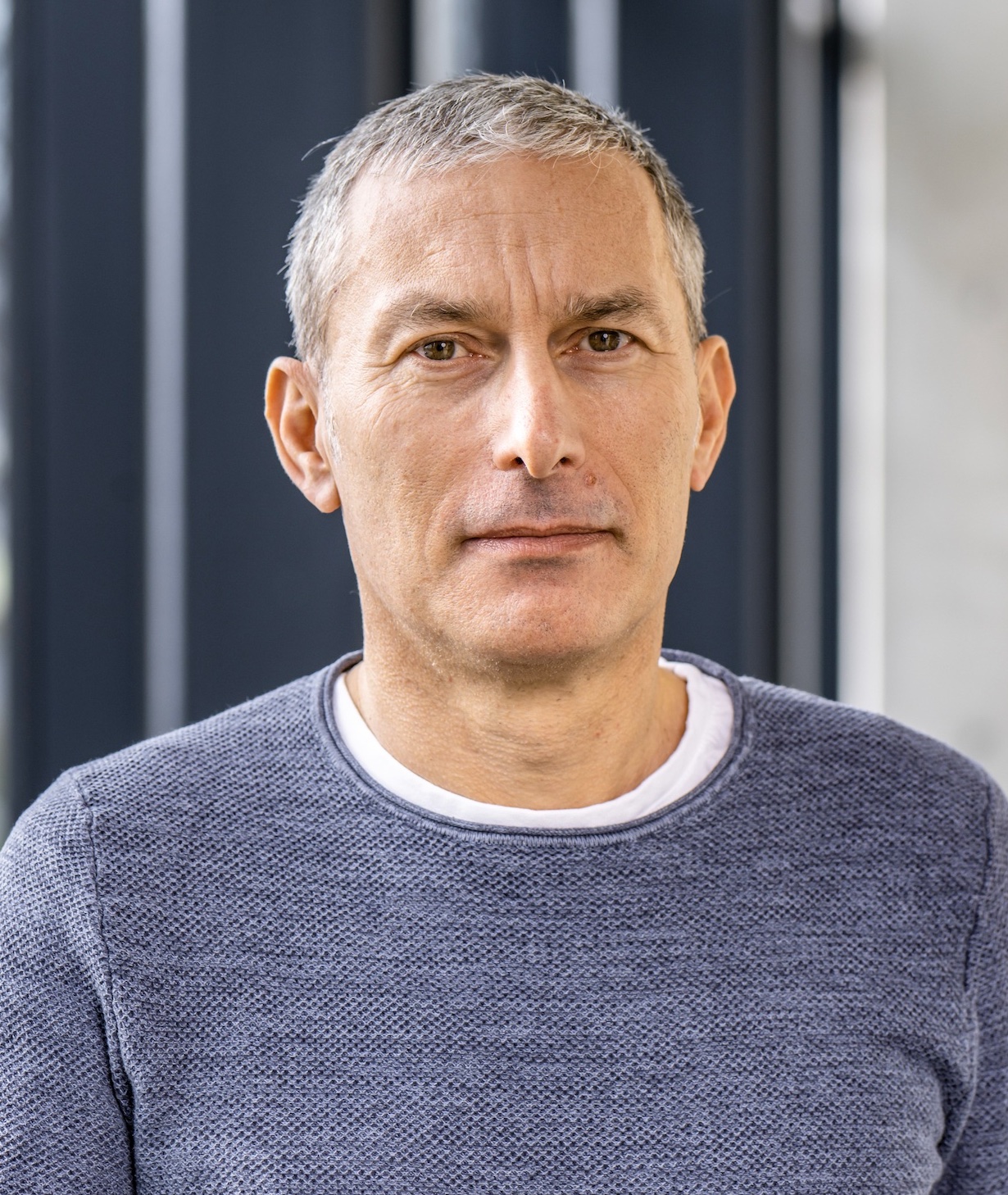}}]{Richard Gloaguen}
is the head of the Exploration Department at the Helmholtz-Institute Freiberg for Resource Technology. Richard Gloaguen received the Ph.D. degree “Doctor Communitatis Europae” in marine geosciences from the University of Western Brittany, Brest, France, in collaboration with the Royal Holloway University of London, U.K., and Göttingen University, Germany, in 2000. He was a Marie Curie Post-Doctoral Research Associate at the Royal Holloway University of London from 2000 to 2003. He led the Remote Sensing Group at University Bergakademie Freiberg, Freiberg, Germany, from 2003 to 2013. Since 2013, he has been leading the division “Exploration Technology” at the Helmholtz-Institute Freiberg for Resource Technology, Freiberg. He is currently involved in UAV-based multisource imaging, laser-induced fluorescence, and non-invasive exploration. His research interests focus on multisource and multiscale remote sensing integration using computer vision and machine learning.
\end{IEEEbiography}%

\begin{IEEEbiography}[{\includegraphics[width=1in,height=1.25in,clip,keepaspectratio]{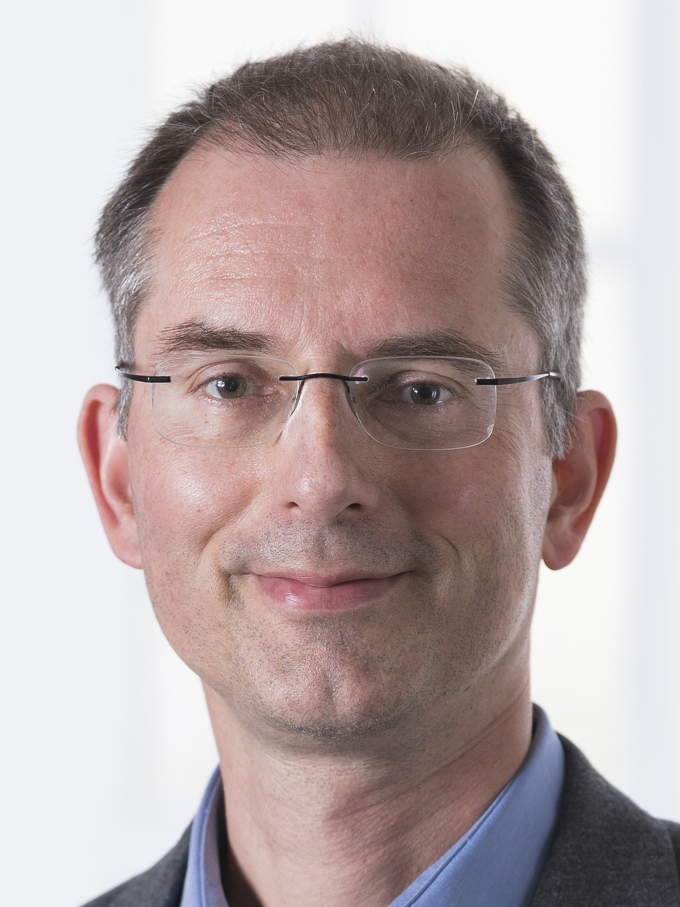}}]{Michael Heizmann}
received the M.Sc. degree in mechanical engineering and the Ph.D. degree in automated visual inspection from the University of Karlsruhe, Germany, in 1998 and 2004, respectively. Since 2016, he has been a Full Professor in mechatronic measurement systems and the Director of the Institute of Industrial Information Technology (IIIT), Karlsruhe Institute of Technology (KIT). His research interests include machine vision, image processing, image and information fusion, measurement technology, machine learning, artificial intelligence and their applications.
\end{IEEEbiography}

\end{document}